\pdfoutput=1

\documentclass[11pt]{article}

\usepackage{acl}

\usepackage{times}
\usepackage{latexsym}

\usepackage[T1]{fontenc}

\usepackage[utf8]{inputenc}

\usepackage{microtype}

\usepackage{inconsolata}

\usepackage{graphicx}
\usepackage{booktabs}

\usepackage{amsmath}
\usepackage{multirow}
\usepackage{CJKutf8}
\usepackage[cache=false]{minted}
\usepackage{soul}

\usepackage{amssymb}
\usepackage{subcaption}

\usepackage{cuted}
\usepackage{caption}
\usepackage{enumitem}

\usepackage[dvipsnames]{xcolor}
\usepackage{colortbl}
\usepackage{fontawesome5}

\setenumerate[1]{itemsep=0pt,partopsep=0pt,parsep=\parskip,topsep=0pt}
\setitemize[1]{itemsep=0pt,partopsep=0pt,parsep=\parskip,topsep=0pt}
\setdescription{itemsep=0pt,partopsep=0pt,parsep=\parskip,topsep=0pt}

\title{PsyCLIENT: Client Simulation via Conversational Trajectory Modeling for Trainee Practice and Model Evaluation in Mental Health Counseling}

\author{
Huachuan Qiu$^{1}$\quad Zhaoming Chen$^{2}$\quad Yuqian Chen$^{1}$\\ \textbf{Yuan Xie}$^{1}$\thanks{Corresponding authors.}\quad \textbf{Yu Lu}$^{1}$\footnotemark[1] \quad \textbf{Zhenzhong Lan}$^{1}$\footnotemark[1] \\
  $^{1}$ Westlake University\\
  $^{2}$ The University of Utah\\
   \small{
   \{qiuhuachuan, lanzhenzhong\}@westlake.edu.cn
   }\\
   \small \href{https://github.com/qiuhuachuan/PsyCLIENT}
  {\faGithub\ \texttt{https://github.com/qiuhuachuan/PsyCLIENT}}
}

\begin{document}
\maketitle

\begin{figure*}[t!]

    \centering
    \includegraphics[width=\textwidth]{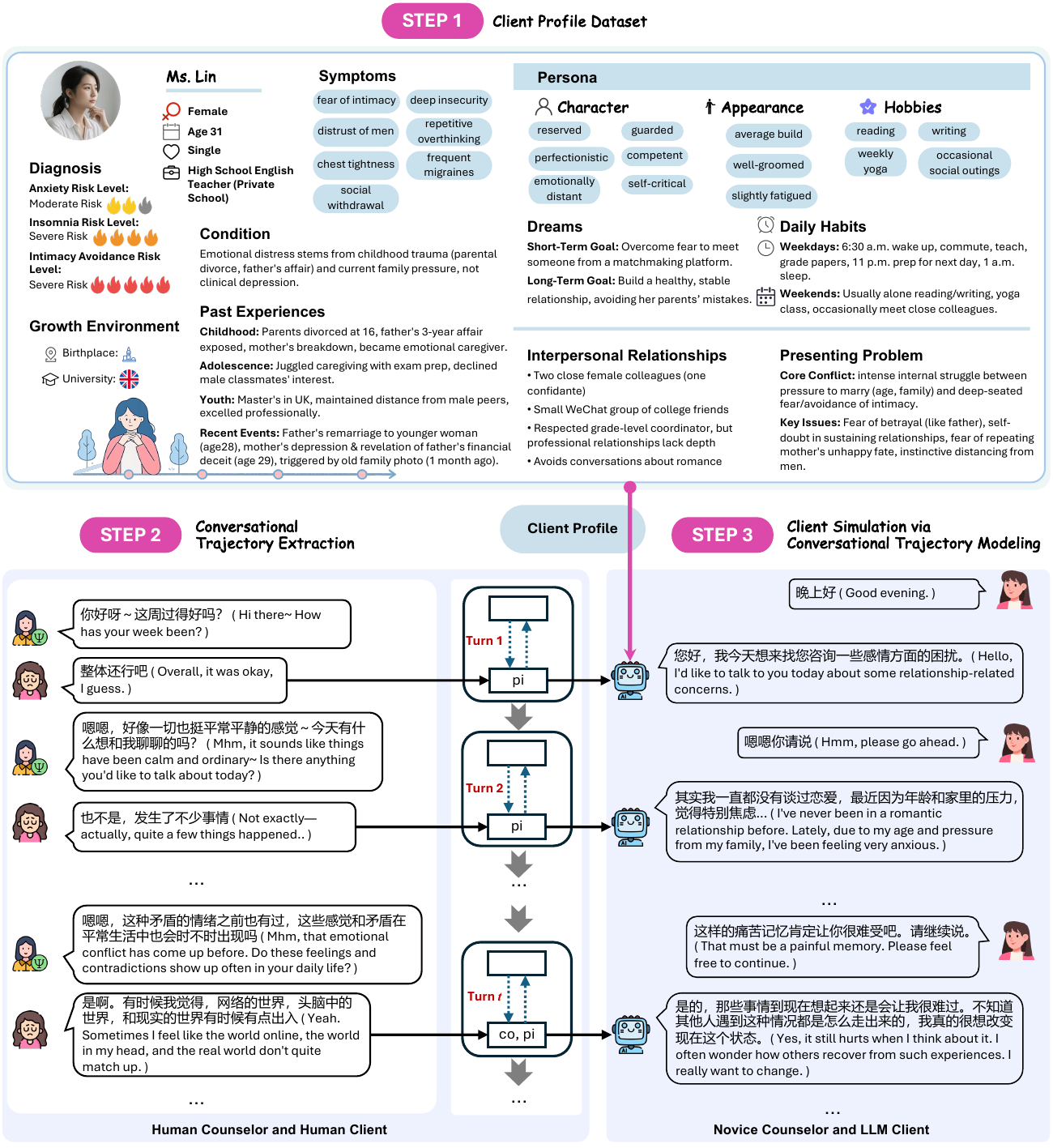}
    \caption{Illustration of our automatic client simulation via conversational trajectory modeling. Step 1 is dataset construction. Step 2 is conversational trajectory extraction. There are 12 types of client behavior labels \citep{li-etal-2023-understanding}, including "pi: Providing Information", "co: Confirming", "ec: Expressing Confusion", "de: Defending", etc. (see Figure \ref{Fig-client-behaviors} in Appendix \ref{sec:client-behaviors}). For a detailed illustration, see Figure \ref{Fig-method} in Appendix~\ref{app:simulation-framework}. Step 3 is the mechanism of client simulation via conversational trajectory modeling. Each turn in a trajectory contains behavior labels and reference content spoken by the human client.}
    \label{Fig-overview}
    \vspace{-0.3cm}
\end{figure*}

\begin{abstract}
LLM-based client simulation provides a scalable approach to novice counselor training, counseling-dialogue synthesis, and interactive evaluation of automated counseling systems. However, existing approaches are limited by insufficient profile diversity, weak behavioral grounding, and the lack of open Chinese-language resources for simulated counseling clients. We propose PsyCLIENT, a framework that conditions simulated-client responses on client profiles, dialogue histories, and conversational trajectories specifying target behaviors and content constraints at each client turn. We also construct PsyCLIENT-CP, a dataset of 120 Chinese client profiles spanning 60 counseling topics. Evaluations involving 24 professional counselors show that PsyCLIENT receives higher ratings of perceived authenticity and training utility than the comparison methods. In a separate source-identification study, PsyCLIENT dialogues were more frequently misclassified as human-client interactions than those generated by the baselines. These results suggest that conversational trajectory modeling can transform static client profiles into behavior-guided, dynamically unfolding counseling interactions.
\end{abstract}

\section{Introduction}
LLM-based simulated clients support three important applications in mental health counseling: providing scalable practice opportunities for novice counselors \citep{louie-etal-2024-roleplay,wang-etal-2024-patient}, enabling controlled synthesis of counseling dialogues \citep{kim-etal-2025-mirror,qiu-lan-2026-interactive}, and facilitating reproducible interactive evaluation of automated counseling systems \citep{yang-etal-2025-consistent,qiu-lan-2026-interactive}.

Counselor education \citep{barrows1993overview} commonly uses peer role-playing to help trainees translate theoretical knowledge into practice. However, peer role-play may not adequately reproduce the clinical complexity and variable behavioral responses observed in real counseling interactions. LLM-based simulated clients may provide a scalable and comparatively low-risk environment in which trainees can practice repeatedly without involving real clients \citep{anthisposition}. Simulated clients can also support controlled synthesis of multi-turn counseling dialogues with specified client characteristics and interaction patterns, reducing reliance on sensitive real-world counseling records \citep{kim-etal-2025-mirror,qiu-lan-2026-interactive}.

Many existing benchmarks for LLM-based counseling systems evaluate isolated responses conditioned on fixed dialogue histories \citep{qiu-etal-2024-smile}. Such benchmarks commonly use reference-based metrics, including BLEU \citep{papineni-etal-2002-bleu}, ROUGE \citep{lin-2004-rouge}, and BERTScore \citep{zhangbertscore}, or LLM-based judges \citep{zheng2023judging} to evaluate response quality. While these methods can assess linguistic similarity or selected response-level counseling skills \citep{qiu-lan-2025-psydial}, they cannot fully capture long-term adaptation, error accumulation, or changes in client state during interaction. Interaction with simulated clients enables evaluation of dialogue-level strategies, process variables, and modeled changes in client state \citep{wang2024towards}. However, the validity of such evaluation depends critically on whether simulated clients respond realistically and consistently to counselor behavior.

Current approaches to client simulation face three limitations: limited access to diverse client profiles \citep{wang-etal-2024-patient}, insufficient modeling of client behavior across turns \citep{wang-etal-2024-patient,louie-etal-2024-roleplay}, and a lack of resources for Chinese counseling contexts \citep{wang-etal-2024-patient,louie-etal-2024-roleplay,yang-etal-2025-consistent}. First, few existing client-profile datasets are publicly available, partly because profiles derived from real counseling cases raise substantial privacy and ethical concerns. Second, simulated clients must exhibit context-dependent behaviors across turns, including disclosure, hesitation, resistance, clarification, and agreement. Existing systems often generate these behaviors implicitly, without explicitly controlling how they unfold in response to counselor interventions. Finally, existing studies predominantly examine English-language interactions, leaving the linguistic and culturally situated characteristics of Chinese counseling conversations underexplored.

To address the data and language gaps, we construct PsyCLIENT-CP, a Chinese dataset of 120 client profiles spanning 60 counseling-related topics. We then propose PsyCLIENT, which represents a counseling interaction as a trajectory of client behaviors and profile-consistent information disclosures conditioned on the evolving dialogue context. Figure~\ref{Fig-overview} provides an overview of the framework. Given a client profile, PsyCLIENT conditions each client turn on a conversational trajectory specifying the target behavior and reference content. The LLM then generates context-dependent responses conditioned on these behavior labels and the reference content.

Our contributions are summarized as follows:

\begin{itemize}
\item We introduce PsyCLIENT-CP, an open-source dataset of 120 Chinese client profiles spanning 60 counseling-related topics.
\item We propose PsyCLIENT, a client simulation framework that decouples client profiles from conversational trajectories to simulate different types of clients.
\item Through evaluations by professional counselors, we show that PsyCLIENT receives higher ratings of perceived authenticity and training utility than existing methods.
\end{itemize}
\section{Related Work}
\label{sec:related-work}
\subsection{LLM-based Clients in Mental Health Counseling}
A number of recent works have explored the potential to train counselors with LLM-based simulated clients \citep{kim-etal-2025-mirror}. For example, \citet{louie-etal-2024-roleplay} introduced Roleplay-doh, a pipeline that uses LLM planning to create customized LLM-based clients based on principle elicitation and response generation with principle adherence. \citet{wang-etal-2024-patient} presented Patient‑$\Psi$, a system built to simulate therapy sessions by incorporating diverse CBT models into the GPT-4 model. On this platform, the authors claimed that trainees can practice their skills in formulating cognitive schemas in simulated CBT scenarios. Further, a user study of Patient‑$\Psi$ demonstrated that novice counselors found that training on this platform led to higher self-rated skill levels and greater confidence than conventional training methods. Experts also reported that the simulated patient in Patient‑$\Psi$ was more authentic than a standard GPT-4 client simulation. \citet{yang-etal-2025-consistent} proposed a multi-module framework for consistent client simulation in motivational interviewing. The framework explicitly tracks the simulated client’s mental state, governs state transitions, and generates state-conditioned behaviors aligned with the client’s motivation, beliefs, change plans, and receptivity.

Despite this progress, existing LLM-based client simulations remain overly compliant \citep{li-etal-2026-recap} and are largely restricted to English. These limitations motivate the present work, which proposes a novel Chinese client simulation framework capable of producing diverse and realistic client behaviors, including both compliant and resistant utterances.

\subsection{LLM Simulations in Other Domains}
LLM-based simulation has also been extensively used in educational contexts \citep{zhang-etal-2025-simulating}, clinical healthcare \citep{wang2025application,mehandru2024evaluating}, business negotiation \citep{hua-etal-2024-assistive}, and legal consultation scenarios \citep{shengbinyue-etal-2025-multi}. For example, \citet{zhang-etal-2025-simulating} proposed an LLM-powered multi-agent classroom simulation, SimClass, which simulates a realistic classroom setting where dynamic teacher–student and student–student exchanges occur, fostering a vivid learning environment with lively interactions. \citet{wang2025application} evaluated ChatGPT as a standardized patient in medical history-taking and found it to be a useful tool for simulating standardized patients in medical assessments.  \citet{hua-etal-2024-assistive} proposed an LLM-powered tri-agent negotiation simulation in which two agents engage in a business negotiation while a third remediator agent rewrites socially inappropriate responses, improving deal outcomes and enabling analysis of cooperative social behaviors.  \citet{shengbinyue-etal-2025-multi} proposed a multi-agent legal consultation simulator, MASER, in which agents take on various roles and collaborate to draft legal complaints.

\section{Problem Formulation}
\label{sec:formulation}
We formalize client simulation as behavior-conditioned language generation along a predefined behavior trajectory. We define the behavior and strategy spaces and characterize the conditions under which a prescribed behavior trajectory can be linguistically realized.

\subsection{Client Behavior Space}
Client utterances can be abstracted into a discrete space of semantic behavior labels. Let $\mathcal{A}$ denote the set of atomic client behavior labels: 
\begin{equation}
\mathcal{A} = \{\mathrm{co}, \mathrm{pi}, \mathrm{rr}, \mathrm{ex}, \mathrm{re}, \mathrm{ec}, \mathrm{de}, \mathrm{sh}, \mathrm{st}, \mathrm{fd}, \mathrm{sa}, \mathrm{ot}\}
\end{equation} 
where, for instance, $\mathbf{co}$ denotes \textit{Confirming} and $\mathbf{pi}$ denotes \textit{Providing Information}. Figure \ref{Fig-client-response} summarizes the behavior categories \citep{li-etal-2023-understanding}, with complete definitions and examples provided in Appendix \ref{sec:client-behaviors}.

We define the valid behavior space as the power set of $\mathcal{A}$ excluding the empty set, denoted as $\mathcal{P}(\mathcal{A}) \setminus \{\emptyset\}$.

A fixed behavior trajectory of length $T$ is defined as a sequence:
\begin{equation}
\tau_{B} = (B_1, \dots, B_t, \dots, B_T)
\end{equation}
where each $B_t \in \mathcal{P}(\mathcal{A}) \setminus \{\emptyset\}$ represents the set of behavior labels associated with the client's turn at time step $t$.

\subsection{Counselor Strategy Space}
Let $\mathcal{S}$ represent the set of available counselor strategies (e.g., \textit{Reflection}, \textit{Open-ended Question}):
\begin{equation}
\mathcal{S} = \{s_1, s_2, \dots, s_n\}
\end{equation}
We define a mapping function $m$ that associates a client behavior set with a non-empty set of appropriate counselor strategies:
\begin{equation}
m: \mathcal{P}(\mathcal{A}) \to \mathcal{P}(\mathcal{S}) \setminus {\emptyset}
\end{equation}
The mapping specifies a set of candidate counselor strategies considered appropriate for a given client behavior; it does not by itself guarantee the clinical validity of the resulting response.

\subsection{Language and Label}
Let $\mathcal{U}$ be the space of all possible natural language utterances. We assume the existence of an oracle annotation function $c: \mathcal{U} \to \mathcal{P}(\mathcal{A})$ that maps an utterance to its behavior labels. For any behavior set $B$, we define the linguistic realization set $L_B$ as:
\begin{equation}L_B = {u \in \mathcal{U} \mid c(u) = B}\end{equation}

\paragraph{Client and Counselor Policies.} We model the conversation between an LLM client equipped with a profile $x \in \mathcal{X}$ and a human counselor.

\begin{enumerate}
\item Client Generator: The client utterance is generated by a large language model $\pi$ with parameters $\theta$, conditioned on the profile $x$ and the target behavior $B$:
\begin{equation}\pi_{\theta}: \mathcal{X} \times \mathcal{P}(\mathcal{A}) \to \mathcal{U}, \quad \text{s.t.} \quad \pi_{\theta}(x, B) \in L_B\end{equation}

\item Counselor Policy: The human counselor operates under a policy $\pi_H$, which produces a response based on the dialogue history $h$ and a selected strategy $s$:\begin{equation}\pi_H: \mathcal{H} \times \mathcal{S} \to \mathcal{U}\end{equation}where $\mathcal{H}$ denotes the set of all possible dialogue histories.\end{enumerate}

\paragraph{Trajectory realizability.} A behavior trajectory $\tau_B=(B_1,\ldots,B_T)$ is realizable for profile $x$ if there exists an utterance sequence $u_t$ such that $c(u_t)=B_t$ for all $t$. See Appendix~\ref{app:problem-formalization}.

\section{Simulation Framework}
\label{sec:simulation-framework}
This section presents PsyCLIENT, an LLM-based framework for simulating counseling clients through predefined conversational trajectories. PsyCLIENT consists of three components: the PsyCLIENT-CP profile dataset, a set of behavior-annotated trajectories extracted from real counseling dialogues, and an LLM-based simulator that generates client responses under profile, trajectory, and dialogue-history constraints.

\subsection{Client Profile Dataset}
\label{sec:client-profile-dataset}
To design the profile schema, we synthesize information categories from commonly used counseling documentation and case-formulation frameworks, including SOAP notes \citep{cameron2002learning}, counseling reports \citep{gelso2014counseling}, intake forms \citep{douglas2016handbook}, and case conceptualization \citep{hill2020helping}. The resulting schema and the correspondence between its fields and these documentation frameworks are provided in Appendix~\ref{app:counseling-notes}.

\begin{figure*}
    \centering
    \includegraphics[width=\textwidth]{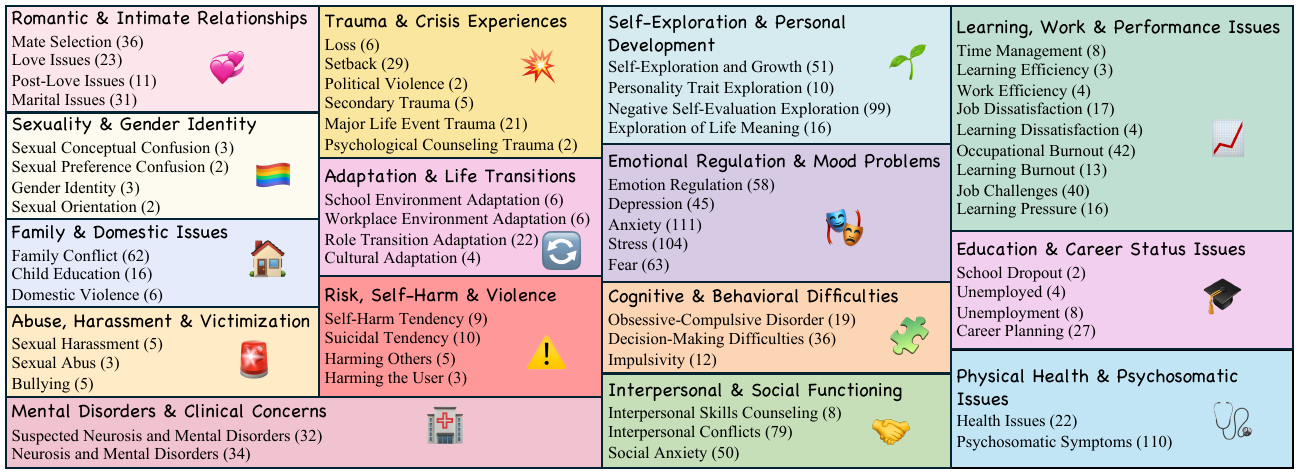}
    \caption{Distribution of topic annotations across client profiles in PsyCLIENT-CP. Values in parentheses indicate the number of profiles associated with each topic; a profile may contain multiple topics.}
    \label{Fig-topic-diversity}
    \vspace{-0.3cm}
\end{figure*}

Inspired by the 60 counseling-related topics (e.g., mate selection, family conflict, and suicidal ideation) proposed by \citet{qiu-etal-2024-smile}, we invited two professional counselors to construct client profiles. Each client profile comprises basic information (e.g., name, age, occupation, education, income, and marital status), as well as diagnostic status, symptoms, personality traits, physical appearance, hobbies, aspirations, daily habits, developmental environment, past experiences, interpersonal relationships, and presenting problems, as illustrated in Figure~\ref{Fig-overview}.

Figure~\ref{Fig-topic-diversity} shows the distribution of profiles across the 60 counseling topics. PsyCLIENT-CP contains 120 profiles. The profile lengths range from 1,421 to 7,740 characters, with a mean length of 3,930 characters and a standard deviation of 2,002. Of the 120 profiles, 89 ($74\%$) exceed 2,000 Chinese characters. A complete example is provided in Figure~\ref{Fig-ai-client-zh} in Appendix~\ref{app:simulation-framework}.

\subsection{Conversational Trajectory Extraction}
\label{sec:conversational-trajectory-extraction}
We derive conversational trajectories from the Chinese counseling-dialogue dataset of \citet{li-etal-2023-understanding}. For each client turn, we retain its original behavior annotation and construct a de-identified content representation after removing personally identifiable information. To obtain sufficiently long interaction trajectories, we retain dialogues with more than 30 dialogue turns. As a result, we obtain 324 real-world conversational trajectories from the original set of 550 dialogues.

\subsection{Client Simulation via Conversational Trajectory Modeling}
\label{sec:client-simulation}

\subsubsection{Prompt for Client Simulation}
We collaborate with three experts in mental health counseling to design a prompt that incorporates a client profile, dialogue history, client behavior label, and reference content. \ul{To address the rigidity of fixed conversational trajectories, we add a soft constraint (If the counselor's question strictly requires a different logical response (e.g., "Confirming" rather than "Defending"), coherence should be prioritized.) in the prompt.}

At client turn $t$, the simulator conditions the LLM on the profile $x$, dialogue history $h_t$, target behavior $B_t$, and reference content $D_t$ to generate the response $u_t$. We present the prompt for client simulation in~Figures \ref{Fig-prompt-v3-zh} and \ref{Fig-prompt-v3-en} in Appendix~\ref{app:prompt-for-client-simulation}.

\subsubsection{Simulation Mechanism}

Figure~\ref{Fig-overview} (see Step 3) shows the interaction between a novice counselor and an LLM-based client. Given a client profile $x$ and trajectory $\tau=\{(B_t,D_t)\}_{t=1}^{T}$, PsyCLIENT initializes an empty dialogue history. At each client turn $t$, the simulator retrieves the target behavior $B_t$ and reference content $D_t$, combines them with $x$ and the current history $h_t$, and prompts the LLM to generate $u_t$. If $B_t$ conflicts with the counselor’s preceding utterance, the model is instructed to prioritize contextual coherence and record the behavioral deviation. The history is then updated, and simulation continues until the trajectory is exhausted or a predefined stopping condition is met.

For details, see Figure~\ref{Fig-method} in Appendix~\ref{app:simulation-framework}. The model used for client simulation is Claude 3.5 Sonnet, with the default settings for \texttt{temperature} and \texttt{top-p}.
\section{Preliminary Studies}
\label{sec:preliminary-study}
To validate the feasibility of our proposed simulation framework, we conduct two studies: a Behavioral Adherence Study and an Independent Study. First, prior work \citep{qiu-lan-2026-interactive,wang-etal-2024-patient} often overlooks a critical limitation: equipping LLMs solely with client profiles is insufficient for achieving realistic client simulations, as such clients fail to exhibit natural resistance behaviors.

\subsection{Behavioral Adherence Study}
\textcolor{BrickRed}{One challenge in client simulation is that LLM-based clients tend to be overly compliant or even sycophantic when interacting with counselors.} This section aims to demonstrate that the LLM client can faithfully follow the specified behavioral patterns. As shown in Figure~\ref{Fig-client-response} in Appendix~\ref{app:behavioral-adherence}, the client exhibits a range of behaviors, including resistance (e.g., "Defending", and "Shifting Topic"; see Rows 6–11). Furthermore, we also present three complete interaction dialogues with the simulated client to demonstrate behavioral adherence. Each client utterance is associated with a behavior label from the original conversational trajectory and is reviewed by three professional counselors. \textcolor{BrickRed}{All three professional counselors agreed that the assigned labels were appropriate, further demonstrating that an LLM-based client can behave both defensively and compliantly.}

\subsection{Independent Study}
We conduct two independent studies to examine whether \emph{client profiles} and \emph{conversational trajectories} can be independently composed for realistic client simulation. To qualitatively validate the independence between client profiles and conversational trajectories, we present illustrative case studies evaluated by professional counselors (see Appendix~\ref{app:independent-study}).

\subsubsection{Profile Independence}
In this study, we fix the conversational trajectory and vary the client profile. Three professional counselors were asked to interact with simulated clients generated from three client profiles and answer the following question: When the conversational trajectory is fixed, to what extent can the simulated clients realistically represent their respective profiles? (1--7 Likert scale)

All three counselors assigned a rating of 7 to each of the three cases. \textit{\textcolor{BrickRed}{Given a fixed conversational trajectory, diverse client profiles can be simulated realistically.}}

\subsubsection{Trajectory Independence}
We further fix the client profile and vary the conversational trajectory. Three professional counselors were asked to interact with simulated clients generated from three conversational trajectories and answer the following question: When the client profile is fixed, to what extent can different conversational trajectories realistically represent this client? (1--7 Likert scale)

All three counselors assigned a rating of 7 to each of the three cases. \textit{\textcolor{BrickRed}{Given a fixed client profile, diverse conversational trajectories can be simulated realistically.}}

\textcolor{BrickRed}{These findings suggest that client profiles and conversational trajectories constitute largely independent dimensions, allowing for flexible recombination to support scalable and realistic client simulation.} See Appendix~\ref{app:preliminary-study} for details.

\section{Experiments}
\label{sec:experiments}
We evaluate PsyCLIENT along three dimensions: perceived authenticity, distinguishability from human clients, and perceived utility for counselor training.

\textbf{RQ1 Authenticity.} Does PsyCLIENT produce more authentic client simulations than existing methods?

\textbf{RQ2 Discrimination.} Can professional counselors distinguish human-counselor–human-client dialogues from human-counselor–LLM-client dialogues based on 30-turn interaction segments?

\textbf{RQ3 Utility.} Do professional counselors perceive PsyCLIENT as more useful for practicing counseling skills than other LLM-based client simulators?

\begin{table*}[t]
\centering
\scalebox{0.78}{
\begin{tabular}{llllll}
\toprule
\textbf{Client}       & \textbf{Fluency} ($\uparrow$) & \textbf{Emotion} ($\uparrow$) & \textbf{Coherence} ($\uparrow$) & \textbf{Contextual Appropriateness} ($\uparrow$) & \textbf{Overall  Authenticity} ($\uparrow$)  \\ \midrule
AnnaAgent   & 6.00    & 6.25                                                               & 4.75      & 5.25                                                                 & 4.75                                                           \\
ConsistentMI & 4.50    & 5.00                                                               & 3.50      & 3.75                                                                 & 3.50                                                           \\
Patient-$\Psi$   & 6.00    & 4.75                                                               & 5.25      & 5.25                                                                 & 4.25                                                           \\
Roleplay-doh  & 4.75    & 5.00                                                               & 4.50      & 5.00                                                                 & 4.75                                                           \\
\rowcolor[HTML]{FFCCC9} PsyCLIENT    & \textbf{6.25}    & \textbf{6.25}                                                               & \textbf{5.75}      & \textbf{6.00}                                                                 & \textbf{6.00}                                                          \\ \bottomrule
\end{tabular}
}
\caption{Authenticity evaluations from four professional counselors.}
\label{tab:authenticity-external-baselines}

\end{table*}

\begin{table*}[t]
\centering
\scalebox{0.78}{
\begin{tabular}{llllll}
\toprule
\textbf{Client}        & \textbf{Listening} ($\uparrow$) & \textbf{Questioning} ($\uparrow$) & \textbf{Emotion-handling} ($\uparrow$) & \textbf{Technique-practice} ($\uparrow$) & \textbf{Recommendation} ($\uparrow$) \\ \midrule
AnnaAgent     & 5.25      & 5.00        & 5.75             & 5.00               & 5.25           \\
ConsistentMI  & 4.25      & 4.00        & 3.75             & 3.75               & 3.50           \\
Patient-$\Psi$    & 4.25      & 4.75        & 5.00             & 4.00               & 4.50           \\
Roleplay-doh   & 5.50      & 5.00        & 5.25             & 4.50               & 4.50           \\
\rowcolor[HTML]{FFCCC9} PsyCLIENT & \textbf{5.75}  & \textbf{5.75}    & \textbf{5.75}         & \textbf{6.00}           & \textbf{6.00}      \\ \bottomrule
\end{tabular}
}
\caption{Expert-rated training utility of different simulated-client methods.}
\label{tab:utility-external-baselines}
\vspace{-0.3cm}
\end{table*}

\begin{table*}[t]
\centering
\scalebox{0.8}{
\begin{tabular}{llllll}
\toprule
\textbf{Client} & \textbf{Fluency} ($\uparrow$) & \textbf{Emotion} ($\uparrow$) & \textbf{Coherence} ($\uparrow$) & \textbf{Contextual Appropriateness} ($\uparrow$) & \textbf{Overall Authenticity} ($\uparrow$) \\ \hline
profile & 4.37$_{1.07}$** & 4.57$_{1.25}$**& 4.61$_{1.13}$**& 4.59$_{1.17}$**& 4.50$_{1.39}$**
\\
 +content & 4.96$_{1.11}$**& 4.98$_{0.83}$**& 4.76$_{0.90}$**& 4.96$_{0.89}$**& 4.82$_{1.06}$**
\\
\rowcolor[HTML]{ecf4ff} +behavior & \ul{5.32$_{0.87}$}& \ul{5.07$_{1.12}$}**& \ul{5.09$_{1.22}$}& \ul{5.16$_{1.09}$}& \ul{5.11$_{1.11}$}*
\\
\rowcolor[HTML]{FFCCC9} PsyCLIENT & \textbf{5.69$_{0.75}$} & \textbf{5.80$_{0.83}$} & \textbf{5.38$_{0.95}$} & \textbf{5.56$_{0.91}$} & \textbf{5.64$_{0.79}$}\\ \bottomrule
\end{tabular}
}
\caption{Authenticity evaluations from professional counselors. Statistically significant differences between each variant and PsyCLIENT are marked with ${*}$ ($p < 0.05$) and ${**}$ ($p < 0.01$), as determined using the Mann-Whitney U test \citep{mann1947test}. The best results are highlighted in bold. The second-best results are highlighted with an underline.}
\label{Tab-authenticity}
\end{table*}

\begin{table*}[t]
\centering
\scalebox{0.8}{
\begin{tabular}{llllll}
\toprule
\textbf{Client} & \textbf{Listening} ($\uparrow$) & \textbf{Questioning} ($\uparrow$) & \textbf{Emotion-handling} ($\uparrow$) & \textbf{Technique-practice} ($\uparrow$) & \textbf{Recommendation} ($\uparrow$) \\ \hline
profile & 4.89$_{1.09}$*& 4.95$_{1.79}$**& 4.65$_{1.09}$**& 4.63$_{1.22}$**& 4.61$_{1.11}$**
\\
 +content & 5.00$_{0.84}$*& 5.07$_{0.93}$**& 4.84$_{1.21}$*& 4.96$_{0.99}$**& 4.93$_{1.00}$**
\\
\rowcolor[HTML]{ecf4ff} +behavior & \ul{5.18$_{0.98}$}& \ul{5.30$_{1.06}$}& \ul{5.16$_{1.06}$}& \ul{5.20$_{1.16}$}& \ul{5.11$_{1.23}$}*
\\
\rowcolor[HTML]{FFCCC9} PsyCLIENT & \textbf{5.42$_{0.80}$} & \textbf{5.78}$_{0.79}$& \textbf{5.47$_{0.78}$}& \textbf{5.69}$_{0.81}$& \textbf{5.69}$_{0.84}$\\ \bottomrule
\end{tabular}
}
\caption{Training utility evaluations from professional counselors. Statistically significant differences between each variant and PsyCLIENT are marked with ${*}$ ($p < 0.05$) and ${**}$ ($p < 0.01$). The best results are highlighted in bold.}
\label{Tab-effectiveness}
\end{table*}

\subsection{Baseline Selection}
To answer RQ1, RQ2, and RQ3, we compare PsyCLIENT with several simulation baselines, including AnnaAgent \citep{wang-etal-2025-annaagent}, ConsistentMI \citep{yang-etal-2025-consistent}, Patient-$\Psi$ \citep{wang-etal-2024-patient}, and Roleplay-doh \citep{louie-etal-2024-roleplay}. Four professional counselors conducted these experiments with a Latin square design. Because most prior systems were not originally developed for Chinese mental health counseling scenarios, we adapted their implementations to the same Chinese evaluation setting and evaluated all methods using identical evaluation guidelines.

Furthermore, we conduct an ablation study with three variants derived from PsyCLIENT: (1) \textbf{profile}, a profile-only baseline inspired by \citet{qiu-lan-2026-interactive} that uses only a detailed client description as input; (2) \textbf{+content}, which augments the profile baseline with reference content from a conversational trajectory; and (3) \textbf{+behavior}, which augments the profile baseline with behavior labels from a conversational trajectory.

\subsection{Collecting Dialogues in Two Categories}
\paragraph{Dialogues Between Human Counselors and Human Clients (Human-Client)} To answer these research questions, we utilize the open-source real-life counseling dataset released by \citet{li-etal-2023-understanding}, sampling 120 dialogues, each comprising more than 30 turns, to ensure a fair comparison.

\paragraph{Dialogues Between Human Counselors and LLM Clients (LLM-Client)} To collect dialogues between human counselors and LLM-based clients, we recruited four professional counselors to interact with LLM-based clients to compare PsyCLIENT with the baselines. We then recruited 20 professional counselors to interact with LLM-based clients in four different settings: profile, +content, +behavior, and PsyCLIENT. All counselors (mean age = 26, 11 females) hold a bachelor's or a master's degree in psychology and have accumulated 362 hours of direct counseling practice on average. In total, each professional counselor was invited to engage with 10 clients in each setting to complete counseling sessions, with each lasting for more than 30 interaction turns. In addition to the task of counseling with LLM-based clients, professional counselors were also invited to complete a questionnaire after every four sessions to evaluate the authenticity and perceived training utility of client simulation in counseling skills training.

\subsection{Evaluation Metrics for RQ1}
To address RQ1, we evaluated the authenticity of client simulations across five dimensions using a 7-point Likert scale. The evaluated dimensions include (1) fluency: naturalness and fluency of the clients' language; (2) emotion: authenticity of the emotional expressions; (3) coherence: logical coherence and relevance of responses; (4) contextual appropriateness: relevance to the counseling contexts (i.e., consistency with typical client behaviors); and (5) overall authenticity: overall sense of realism.

\subsection{Evaluation Metrics for RQ2}
To answer RQ2, we conducted an expert evaluation on 30-turn dialogues. Specifically, we sampled ten 30-turn dialogues from each client setting and asked a professional counselor to judge whether each dialogue involved an LLM client or a human client.

We then recruited 24 professional counselors (Mean Age = 25.42, 14 females), each with a bachelor's or master's degree in counseling psychology or psychiatry and an average of 356 hours of experience. They were asked to judge whether a specific dialogue segment came from a human-client or an LLM-client session. We sampled 5-turn segments from the dialogues to reduce labeling costs and created 90 test instances for each of the five settings (three variants, one human-client condition, and PsyCLIENT), resulting in 450 test instances in total. Further, we conducted a new source-discrimination experiment using 30-turn dialogue segments. This experiment included 45 test instances for each of the five settings: three variants, one human-client condition, and PsyCLIENT.

To assess the quality gap between human-client and LLM-client dialogues, we introduced two discrimination tasks: (1) mixed samples from three settings and PsyCLIENT (360, 90 samples per setting); and (2) mixed samples from human-client, PsyCLIENT, and the three baselines (450, 90 samples per setting). Task 2 aims to determine whether including human-client samples helps professional counselors distinguish between human-client and LLM-client dialogues.

\subsubsection{Expert Evaluation}
The counselors' task is to judge whether a dialogue segment comes from a human-client or an LLM-client interaction.

\subsubsection{LLM Evaluation}
We design a prompt that enables an LLM to determine automatically whether a dialogue segment comes from a human-client or an LLM-client interaction, with the goal of evaluating how accurately the LLM can identify the source of the segment.

\subsection{Evaluation Metrics for RQ3}
We assessed counselors' perceived utility of each simulator for practicing counseling skills using a 7-point Likert scale. Specifically, the dimensions include (1) listening: opportunities to practice active listening; (2) questioning: opportunities to practice questioning; (3) emotion-handling: opportunities to respond to emotions; (4) technique-practice: opportunities to apply counseling techniques; and (5) recommendation: willingness to recommend the simulator for training.

\begin{table}[t]
\centering
\small
\begin{tabular}{lc}
\toprule
\textbf{Client} & \textbf{Number of Accurate Discriminations} \\
\midrule
profile      & 10 \\
+content    & 10 \\
+behavior     & 10 \\
PsyCLIENT    & \textbf{8} \\
AnnaAgent    & 10 \\
ConsistentMI & 10 \\
Patient-$\Psi$   & 10 \\
Roleplay-doh  & 10 \\
Human-Client & 10 \\
\bottomrule
\end{tabular}
\caption{Number of accurate discriminations for each client type.}
\label{tab:accurate_discriminations}
\vspace{-0.3cm}
\end{table}

\begin{table*}[t!]
\centering
\scalebox{0.8}{
\begin{tabular}{lllllllllll}
\toprule
\multirow{2}{*}{\textbf{Judge}} & \multicolumn{2}{l}{\textbf{Human-Client}} & \multicolumn{2}{l}{\textbf{profile}} & \multicolumn{2}{l}{\textbf{+content}} & \multicolumn{2}{l}{\textbf{+behavior}} & \multicolumn{2}{l}{\textbf{PsyCLIENT}} \\ \cline{2-11}
 & A & B & A & B & A & B & A & B & A & B \\ \hline
Qwen3.6-max-preview \citep{qwen3} & 43 & 44 & 45 & 45 & 42 & 43 & 44 & 41 & 40 & \textbf{37} \\
kimi-k2-0905-preview \citep{team2025kimi} & 39 & 45 & 41 & 32 & 38 & 24 & 41 & 22 & 40 & \textbf{7} \\
MiniMax-M2.7 \citep{chen2026minimax} & 32 & 45 & 32 & 5 & 32 & 5 & 27 & 8 & 22 & \textbf{3} \\
deepseek-v4-flash \citep{xu2026deepseek} & 34 & 38 & 45 & 19 & 42 & 25 & 41 & 22 & 38 & \textbf{16} \\
deepseek-v4-pro \citep{xu2026deepseek} & 42 & 41 & 37 & 41 & 36 & 32 & 32 & 36 & 32 & \textbf{31} \\ \hline
Total &\multicolumn{2}{l}{\textbf{403}} & \multicolumn{2}{l}{\textbf{342}} & \multicolumn{2}{l}{\textbf{319}} & \multicolumn{2}{l}{\textbf{314}} & \multicolumn{2}{l}{\textbf{266}}\\
\bottomrule
\end{tabular}
}
\caption{Source-discrimination results of LLM-based evaluation on 30-turn dialogue segments. Each value denotes the number of correctly discriminated cases out of 45. Lower values indicate that the dialogue source is harder for the LLM judge to identify. “A” and “B” denote the two prompt versions with reversed label orders, respectively. For details, see Figure \ref{Fig-prompt-evaluation} in Appendix~\ref{app:prompts-for-llm-eval}.}
\label{Tab-results-LLM-evaluation}
\end{table*}

\section{Results and Discussion}
\label{sec:results-and-discussions}
\subsection{Authenticity}
PsyCLIENT receives the highest mean ratings on all authenticity dimensions among the evaluated systems, as shown in Table \ref{tab:authenticity-external-baselines}. In the authenticity evaluation, PsyCLIENT achieves the highest scores on Fluency, Coherence, Contextual Appropriateness, and Overall Authenticity, while matching the best baseline on Emotional Expressiveness.

As shown in Table~\ref{Tab-authenticity}, PsyCLIENT achieves higher mean ratings than all ablation settings. Compared with profile only, +content produces relatively small numerical gains across the five dimensions. The +behavior setting shows the largest numerical gains over the profile-only setting. These results suggest that behavior conditioning contributes more strongly to perceived authenticity than content conditioning, while combining the two yields additional gains on selected dimensions. The full model achieves the highest mean ratings, suggesting that content constraints may provide additional benefits when combined with behavior conditioning.

\begin{figure}[t!]

    \centering
    \includegraphics[width=\columnwidth]{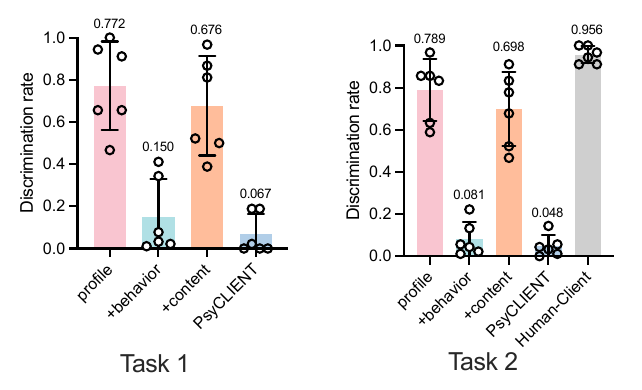}
    \caption{Discrimination rate of human expert evaluation.}
    \label{Fig-overall-acc}
    \vspace{-0.3cm}
\end{figure}

\subsection{Source Discrimination}
\subsubsection{Expert Evaluation}
As shown in Table~\ref{tab:accurate_discriminations}, the professional counselor correctly identified all ten dialogues from each baseline LLM-client setting, whereas only eight of the ten PsyCLIENT dialogues were correctly identified.

Figure \ref{Fig-overall-acc} summarizes expert discrimination accuracy under two evaluation tasks. In Task 1, experts achieved mean AI-source detection rates of 0.150 (SD = 0.179) and 0.067 (SD = 0.095) for dialogues generated under the +behavior and PsyCLIENT settings, respectively. In Task 2, the mean AI-source detection rates further decreased to 0.081 (SD = 0.081) and 0.048 (SD = 0.051) for the +behavior and PsyCLIENT settings, respectively. Although Task 2 included human-client dialogues, the experts were still unable to reliably distinguish them from those generated by LLM clients. PsyCLIENT dialogues were rarely identified as LLM-client interactions by expert evaluators in the source-discrimination task.

\subsubsection{LLM Evaluation}
For the LLM-based evaluation, we used several representative LLMs. In Table \ref{Tab-results-LLM-evaluation}, each value denotes the number of correctly discriminated cases out of 45. Therefore, lower values indicate that the judge found the source harder to identify. In the 30-turn setting, PsyCLIENT consistently yields fewer correct discriminations than most ablated variants and prior baselines. Aggregated across all LLM judges and A/B settings, the correct-discrimination rate for PsyCLIENT is 59.1\%, compared with 76.0\% for profile, 70.9\% for +content, and 69.8\% for +behavior. This indicates that PsyCLIENT dialogues remain more difficult to distinguish from human-client dialogues even when evaluators are given substantially longer conversational evidence.

\subsection{Training Utility}
PsyCLIENT receives the highest mean ratings across the five perceived-training-utility dimensions, as shown in Table \ref{tab:utility-external-baselines}. As presented in Table~\ref{Tab-effectiveness}, PsyCLIENT consistently outperforms all baselines in counselor-rated training utility, achieving the highest mean scores across all five dimensions, including listening (5.42), questioning (5.78), emotion-handling (5.47), technique-practice (5.69), and recommendation (5.69) on a 1–7 scale. Augmenting the model with reference content leads to modest but consistent improvements, whereas the +behavior setting yields substantially larger gains and becomes the strongest variant across all dimensions. Overall, the superior performance of PsyCLIENT suggests that jointly incorporating client behavior labels and reference content is essential for enabling high-quality, realistic client simulation.
\section{Conclusion}
\label{sec:conclusion}
In this work, we introduced PsyCLIENT, a framework that conditions simulated client responses on client profiles, dialogue histories, and conversational trajectories specifying target behaviors and reference content at each turn. We also constructed PsyCLIENT-CP, an open resource containing 120 Chinese client profiles across 60 counseling topics. Evaluations involving professional counselors showed that PsyCLIENT received higher ratings of perceived authenticity and training utility than the comparison methods. In source identification tasks, PsyCLIENT dialogues were also more frequently classified as human client interactions than dialogues generated by the baselines. Our ablation results suggest that behavior conditioning provides the primary improvement, while reference content offers complementary benefits. Overall, these findings indicate that conversational trajectory modeling can transform static client profiles into dynamic counseling interactions guided by explicit behavioral and reference content. PsyCLIENT provides a foundation for future research on counselor training, counseling dialogue synthesis, and interactive evaluation of automated counseling systems.

\section*{Limitations}
Our current work focuses on simulating a single counseling session. However, real-world psychotherapy typically unfolds across multiple sessions. Despite this limitation, our study represents a meaningful first step: we move beyond short, superficial interactions and successfully support extended single-session dialogues exceeding 40 conversational turns—something existing approaches have been unable to achieve. This advance establishes a solid foundation for future research on multi-session counseling simulation.

Cultural generalizability also warrants attention. Our client profiles are deeply rooted in Chinese sociocultural contexts, such as stress or anxiety associated with the national college entrance examination and familial expectations for achievement, making them highly relevant for training in China but potentially less relevant elsewhere. Adapting these profiles to reflect different cultural backgrounds will be essential for ensuring immersive, culturally appropriate training experiences for counselors and clients in diverse regions.



\section*{Ethical Considerations}
\textbf{Ethics Review Board Approval.}
Our research was reviewed and approved by the Westlake University Institutional Ethics Committee (Approval No. 20211013LZZ001).

\paragraph{Recruitment and Payment.} In total, we compensated all participating counselors fairly for their contributions. For those conducting simulated counseling sessions with LLM-based clients, we provided an hourly rate of 70 RMB, along with an additional 20 RMB for completing post-session questionnaires. Counselors participating in the discrimination task, judging whether dialogues were generated by human-client or simulated-client, received 1 RMB per item. All counselors participated voluntarily and were informed of their right to withdraw from the study at any time. The payment rates were designed to be competitive and aligned with local standards, ensuring appropriate recognition of their time and expertise.

\paragraph{Ethical Concerns.} This research aims to develop an educational assistant designed to facilitate the training of novice counselors and the evaluation of LLM-based counselors.

\paragraph{AI Assistants.} We use large language models (LLMs; e.g., ChatGPT) solely for grammar checking.


\section*{Acknowledgements}
We thank the anonymous reviewers for their valuable feedback. This work was supported by the Zhejiang Provincial Key Research and Development Program (Grant No. 2025C01104) and the 76th General Program of the China Postdoctoral Science Foundation (Grant No. 2024M76934).

\bibliography{anthology,custom}

\appendix
\clearpage
\section{Problem Formalization}
\label{app:problem-formalization}
\subsection{Formalization}
To validate the feasibility of the proposed modeling framework, we rely on two core assumptions regarding the expressivity of language and the competence of the counselor.

\paragraph{Assumptions.}

\begin{itemize}
\item (H1) Expressivity: For any valid behavior set $B$, there exists at least one natural language utterance that realizes it. Formally, $\forall B \in \mathcal{P}(\mathcal{A}), L_B \neq \emptyset$.
\item (H2) Human Executability: A human counselor can always formulate a valid utterance given a history and a valid strategy. Formally, $\forall h \in \mathcal{H}, s \in \mathcal{S}, \pi_H(h, s) \neq \emptyset$.
\end{itemize}

\subsection{Formalization of the Problem Space}
We now demonstrate that for any defined behavior flow, a corresponding valid dialogue sequence can be constructed.

\textbf{Theorem 1 (Universal Realizability).} \textit{For any client profile $x \in \mathcal{X}$ and behavior flow $\tau_{B} = (B_1, \dots, B_T)$, there exists a realized dialogue sequence $\mathcal{D} = ((u_1, s_1, v_1), \dots, (u_T, s_T, v_T))$ such that for all $t \in [1, T]$:}
\begin{enumerate}
\item $u_t = \pi_{\theta}(x, B_t)$ and $c(u_t) = B_t$;
\item $s_t \in m(B_t)$;
\item $v_t = \pi_H(h_{t-1}, s_t)$, where $h_{t-1}$ represents the interaction history up to turn $t-1$.
\end{enumerate}

\textit{Proof.} We prove this by induction on the turn index $t$.

\textbf{Base Case ($t=1$):} By assumption (H1), $L_{B_1} \neq \emptyset$. Therefore, the generator $\pi_{\theta}$ can produce a valid utterance $u_1 = \pi_{\theta}(x, B_1)$. Since $m$ maps to non-empty sets, there exists at least one valid strategy $s_1 \in m(B_1)$. By assumption (H2), given the initial history $h_0$ and strategy $s_1$, the counselor policy yields a valid response $v_1 = \pi_H(h_0, s_1)$. Thus, the tuple $(u_1, s_1, v_1)$ is realizable.

\textbf{Inductive Step:} Assume the sequence is realizable up to turn $t-1$. This implies a history $h_{t-1}$ exists. For turn $t$, consider the target behavior $B_t$.
\begin{enumerate}
    \item By (H1), we generate $u_t = \pi_{\theta}(x, B_t) \in L_{B_t}$.
    \item We select a strategy $s_t$ from the non-empty set $m(B_t)$.
    \item Given the history $h_{t-1}$ (extended by $u_t$) and strategy $s_t$, assumption (H2) guarantees the existence of a counselor response $v_t = \pi_H(h_{t-1}, s_t)$.
\end{enumerate}

\textbf{Key Takeaway:} Since the flow length $T$ is finite, the induction terminates after $T$ steps. Thus, the flow $\tau_{B}$ is universally realizable for arbitrary client profiles within the defined constraints.
\section{Client Profiling via Counseling Notes}
\label{app:counseling-notes}
\textbf{SOAP Note.} The SOAP note is a standard framework for documenting counseling sessions, comprising four components: Subjective, Objective, Assessment, and Plan. The Subjective component captures the client’s self-reported experiences, whereas the Objective component records the counselor’s observable findings. The Assessment component provides an initial clinical evaluation and provisional diagnosis, and the Plan component specifies treatment goals and intervention strategies.

\textbf{Intake Session.} The intake session form is administered during the initial consultation to systematically collect the client’s demographic information and presenting problems. It typically includes personal background, chief complaint, psychosocial history, current psychological status, and social functioning.

\textbf{Case Conceptualization.} Case conceptualization offers a theoretically grounded structure for constructing client profiles. In particular, the cognitive behavioral therapy (CBT) case conceptualization framework emphasizes the identification of maladaptive cognitive patterns and behavioral tendencies that maintain psychological distress.

\subsection{Components of a SOAP Note Template}
\subsubsection{Subjective Component}

The subjective component of a SOAP note captures the clinician’s documentation of the client’s subjective experiences, based on clinical judgment and the client’s self-report. This component typically includes the following elements:
\begin{itemize}
    \item Presenting Problem: A brief description of the client’s primary concern.
    \item History of Present Illness: A concise summary of the development and course of the presenting problem.
    \item General Medical History: An overview of relevant medical history not addressed in prior sections.
    \item Review of Symptoms: Documentation of positive and negative symptoms currently experienced by the client, as reported by the client or inferred through clinical observation and assessment.
\end{itemize}

\paragraph{Example:} The client reports experiencing depressive symptoms consistently over the past ten months, coinciding with a period during which he was laid off from his job and experienced the death of his mother. He describes symptomatology consistent with Major Depressive Disorder, Single Episode, Moderate, including depressed mood, sadness, crying spells, self-loathing, and lack of motivation, occurring on most days over the past two weeks.

\subsubsection{Objective Component}

The objective component documents the client’s observable presentation during the session. This component may include:
\begin{itemize}
    \item Physical presentation
    \item Observable signs and self-reported symptoms
    \item Psychological status
\end{itemize}

\paragraph{Example:} The client presented with a depressed mood and flat affect and reported a lack of motivation over the past week while struggling to make progress in his employment search. He exhibited signs of impaired immediate memory, reporting difficulty recalling events from the previous week.

\subsubsection{Assessment Component}

The assessment component reflects the clinician’s evaluation of the client’s psychological status since the previous session. It includes progress toward treatment goals, changes in symptom severity, risk assessment (when applicable), and an overall judgment regarding the continued necessity of services. This component may address:
\begin{itemize}
    \item Changes in symptom frequency or duration
    \item Progress (or lack thereof) toward treatment goals
    \item Changes in risk level
\end{itemize}

\paragraph{Example:} The client reported no significant change in the frequency or duration of symptoms but noted increased difficulty with motivation over the past week. He reported no progress toward his goal of improving self-direction in relation to his job search. The client denied suicidal ideation, intent, or plan since the previous session, and no contrary indicators were observed.

\subsubsection{Plan Component}

The plan component outlines the clinician’s planned interventions and case management activities, including referrals, care coordination, and the mutually agreed-upon therapeutic plan until the next session. This section often functions as structured therapeutic “homework” for the client.

The plan may include diagnosis, referrals, therapeutic assignments, and scheduling information.

\paragraph{Example:}

\begin{itemize}
    \item \textbf{Diagnosis:} F32.1 — Major Depressive Disorder, Single Episode, Moderate.
    \item \textbf{Referrals:} No referrals recommended at this time.
    \item \textbf{Assignments:} The client plans to apply to three jobs over the next week to support the goal of increasing self-direction.
    \item \textbf{Plan:} Continue the current treatment plan.
\end{itemize}

\section{Simulation Framework}
\label{app:simulation-framework}
\begin{figure*}[t!]

    \centering
    \includegraphics[width=\textwidth]{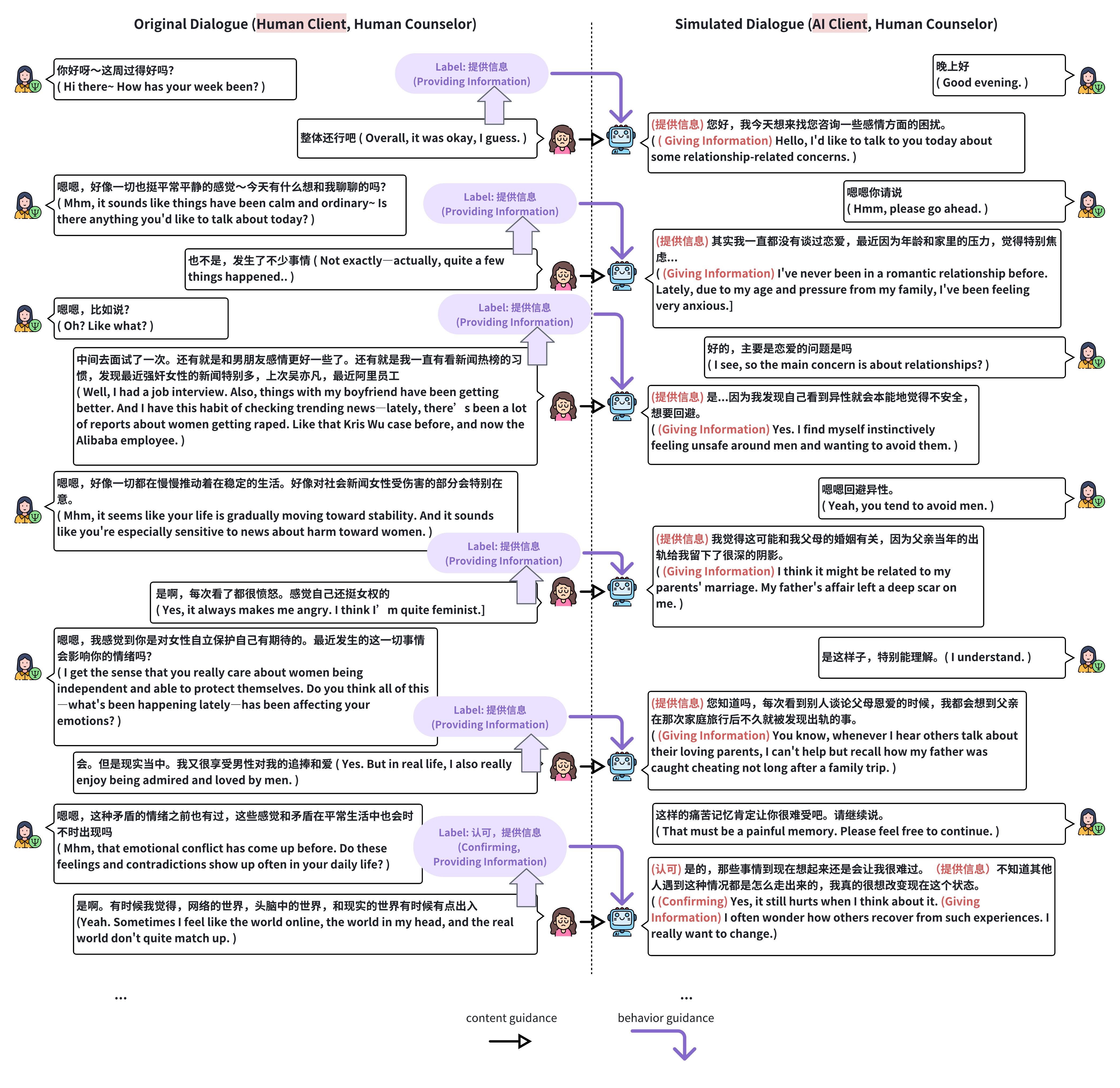}
    \caption{The detailed illustration of our automatic client simulation framework via conversational trajectory modeling.}
    \label{Fig-method}
\end{figure*}

We present a detailed illustration of our automatic client simulation framework via conversational trajectory modeling, as shown in Figure~\ref{Fig-method}.

\subsection{Client Profile Dataset}

\begin{CJK*}{UTF8}{gbsn}
\begin{figure*}[htp]
\centering
\begin{Verbatim}[fontsize=\scriptsize, frame=single, framerule=0.3pt, framesep=4mm, breaklines, xleftmargin=0.5em, xrightmargin=0.5em, formatcom=\rmfamily]
一、基本信息
    - 姓名：林女士
    - 年龄：31 岁
    - 职业：高中英语教师（私立学校）
    - 教育：英国某大学教育硕士
    - 收入：税前年薪 20 - 25 万
    - 婚姻状态：未婚
二、亲密关系历史与期望 林女士目前单身，从未社交场合也会刻意保持界限。最近两年，由于年龄和家庭压力，她在相亲网站注册过账号，但从未真正约见对方。她表示每当准备迈出这一步时，就会感到强烈的不安全感和恐慌感，担心对方也会像父亲一有过恋爱经历。在高中和大学期间，曾有男生表达过好感，但她都以学业繁忙为由婉拒。在英国留学期间，也有同学尝试靠近，但她始终保持着礼貌的距离。她会下意识地避免与异性产生过多互动，即便是普通的样出轨，或是自己无法维持一段健康的关系。
三、家庭情况 父母离异（父亲出轨导致），与母亲生活。父母在她16岁时离异，起因是发现父亲与其事业合伙人的婚外情，这段关系已经持续了3年。得知真相后，母亲曾一度情绪崩溃，林女士不得不在照顾母亲情绪和准备高考之间艰难平衡。离婚后，父亲与其事业合伙人组建了新家庭，并育有一子。目前与父亲仅在春节等重要节日会见面，互动流于形式。父亲偶尔会转账表达关心，但她往往会找理由婉拒。与母亲的关系亦趋向复杂：一方面，她努力扮演着支持者和照顾者的角色；另一方面，母亲时常流露出对她终身大事的焦虑，这让她倍感压力。母亲经常会说“你要好好找对象，不要像我一样……”这样的话，无形中加重了她的心理负担。
四、人际关系状态 工作环境中有两位关系较好的女性同事，会一起吃午饭、偶尔下班小聚。其中一位已婚同事经常充当她的倾诉对象，但她发现自己很难完全敞开心扉。还保持联系的大学好友约有3-4位，主要通过微信群保持联络，实际见面频率每年不超过2次。在学校担任年级组长，工作能力受到领导和同事认可，但她觉得这些关系停留在专业层面，难以深入。每当同事们讨论感情话题时，她会感到局促不安，往往借故回避。
五、身体状况 轻度焦虑，经常失眠，偶有偏头痛。失眠问题在父母离异后开始显现，近年来愈发明显。平均入睡时间需要1-2小时，经常在凌晨2-3点仍然清醒。白天会感到疲惫，但仍强撑着保持工作效率。每月约有2-3次偏头痛发作，通常在工作压力大或与家人产生矛盾后加重。焦虑症状主要表现为：在需要社交的场合会出现心跳加速、手心出汗；独处时会反复思考人际关系问题；对未来感到迷茫和不确定时会出现胸闷症状。
六、生活方式 工作日专注于教学工作，课余时间喜欢阅读和写作；周末偶尔和同事聚会，大多独处；每周固定去一次瑜伽课。
七、主诉 林女士目前面临的最大困扰是年龄增长带来的婚恋压力与自身对亲密关系的恐惧和抵触之间的矛盾。她希望能开展正常的恋爱，但对亲密关系充满了恐惧和抵触，这种矛盾让她感到十分痛苦和迷茫。她担心自己无法像正常人一样建立一段健康、稳定的亲密关系，也担心自己会重蹈母亲的覆辙，这种恐惧让她在面对异性时总是下意识地保持距离，即使有男生向她表达好感，她也会以各种理由婉拒。
八、其余附带问题
    1. 社交圈子较窄，主要局限于工作环境：林女士的社交圈子相对较小，主要集中在工作环境中。她在学校担任年级组长，工作能力受到领导和同事的认可，但她觉得这些关系停留在专业层面，难以深入。她与两位关系较好的女性同事会一起吃午饭、偶尔下班小聚，其中一位已婚同事经常充当她的倾诉对象，但她发现自己很难完全敞开心扉。此外，她还保持联系的大学好友约有 3 - 4 位，主要通过微信群保持联络，实际见面频率每年不超过 2 次。每当同事们讨论感情话题时，她会感到局促不安，往往借故回避。
    2. 存在完美主义倾向：林女士在工作中追求极致，常常花费大量时间修改教案和批改作业。她对外表和社交形象的要求也很高，这种完美主义倾向让她在面对潜在伴侣时，会设置很多可能过于理想化的标准。她承认自己对伴侣的期待过高，但又难以降低标准，这种矛盾让她在婚恋问题上更加迷茫和困惑。
九、问题发展过程
    1. 初次觉察困扰：林女士首次清晰意识到自己在亲密关系方面存在严重困扰是在 26 岁那年。当时她刚从英国留学回国，在现任学校工作满一年。一次学校组织的联谊活动中，她被安排与另一所私立学校的男教师进行分组活动。整个互动过程中，她发现自己异常紧张，心跳加速、手心出汗、呼吸急促。活动结束后，那位男教师表达了进一步了解的意愿，却引发了她长达一周的失眠和焦虑。她开始回溯自己的成长经历，意识到自己在面对异性示好时的反应远超出正常范围，这种过度防备的状态严重影响了她建立正常社交关系的可能性。
    2. 发展过程中的重要事件一：父亲再婚：在林女士 28 岁那年，父亲的再婚典礼成为了一个重要的转折点。虽然父亲再婚前就已经有了新的家庭，但正式的婚礼仪式还是给她带来了巨大的心理冲击。婚礼当天，她作为女儿不得不出席，全程需要面对父亲的新家庭以及众多亲友的关注。最让她难以接受的是，她发现自己与父亲的新配偶年龄差距仅有 12 岁，这让她感到极度不适。婚礼上，当她不得不配合着叫对方 “阿姨” 时，内心经历了剧烈的挣扎。整个婚礼过程中，她强忍着情绪，但在婚礼结束后的当晚，她独自在家中崩溃大哭。这件事加深了她对婚姻的怀疑和恐惧，也让她更加确信 “男人是不可信任的” 这一认知。从那以后，她开始彻底回避任何可能发展为恋爱关系的社交机会。
    3. 发展过程中的重要事件二：母亲的抑郁症状：在林女士 29 岁时，母亲因为工作压力和更年期影响，出现了明显的抑郁症状。作为独女的她不得不承担起照顾母亲的责任。在这段期间，母亲经常在情绪低落时反复提起过去的婚姻，诉说自己对前夫的悔恨和怨恨。有一次，母亲在深夜情绪崩溃，说出了更多婚姻破裂时的细节：原来父亲不仅出轨，还在婚内将大量家庭积蓄转移给了现任妻子。这些新细节的揭露让林女士对婚姻和亲密关系产生了更深的恐惧。她开始质疑自己是否有能力识别和处理亲密关系中的欺骗，也更加坚信独身可能是最安全的选择。这段经历持续了近半年，不仅加重了她对亲密关系的恐惧，也让她对自己的判断能力产生了更多怀疑。
    4. 最近的触发事件：一个月前，林女士在整理旧物时，偶然发现了父母离婚前全家最后一次旅行的照片。照片中父亲搂着母亲的肩膀，一家三口在海边看起来其乐融融。然而她清楚地记得，就在这次旅行后的两个月，父亲的婚外情就被发现了。这张照片让她产生了强烈的情绪波动，她开始反复思考：“如果连看起来那么幸福的家庭都会破裂，我要如何相信亲密关系的真实性？”这个想法导致她连续几天难以入睡，最终促使她下定决心寻求心理咨询的帮助。
十、说话风格
林女士的说话风格属于 reserved（内向） 类型。她在对话中提供简短、模糊或回避性的回答，尤其是当被问及个人情感或婚恋相关的问题时，会表现出不愿意分享个人信息或感受的态度。在社交场合，她会保持距离，避免深入的情感交流，需要更多的提示和鼓励才会开口。同时，她对异性普遍存在的不信任感以及对亲密关系的恐惧，使她在与治疗师或其他人交流时表达出不信任或怀疑的情绪。
\end{Verbatim}
\caption{An example of an AI client. For the English version, see Figure~\ref{Fig-ai-client-en}.}
\label{Fig-ai-client-zh}
\end{figure*}
\end{CJK*}

\begin{CJK*}{UTF8}{gbsn}
\begin{figure*}[htp]
\centering
\begin{Verbatim}[fontsize=\scriptsize, frame=single, framerule=0.3pt, framesep=4mm, breaklines, xleftmargin=0.5em, xrightmargin=0.5em, formatcom=\rmfamily]
1. Basic Information
    - Name: Ms. Lin
    - Age: 31
    - Occupation: High school English teacher (private school)
    - Education: Master’s degree in Education from a university in the UK
    - Annual Income: 200,000–250,000 RMB (pre-tax)
    - Marital Status: Single
2. Intimate Relationship History and Expectations
Ms. Lin is currently single and has never been in a romantic relationship. During high school and college, several male classmates showed interest in her, but she politely declined, citing academic pressure. While studying in the UK, she maintained a polite distance from male peers who tried to approach her. She tends to avoid interactions with men, even in regular social settings, and deliberately maintains emotional boundaries.
Over the past two years, due to increasing age and family pressure, she registered on matchmaking platforms but never proceeded to meet anyone in person. Every time she considers taking a step forward, she experiences intense insecurity and fear, worrying that the man might cheat like her father did or that she might be incapable of sustaining a healthy relationship.
3. Family Background
Her parents divorced when she was 16 after her father’s extramarital affair with his business partner, which had lasted three years. Upon discovering the affair, her mother had a breakdown, and Ms. Lin had to juggle emotional caregiving with preparing for university entrance exams.
After the divorce, her father started a new family with the partner and had a son. Ms. Lin now only sees her father during major holidays, and the interactions are mostly formal. Though her father occasionally transfers her money as a gesture of care, she usually declines politely.
Her relationship with her mother is complex: she plays a supportive and caregiving role, but her mother frequently expresses anxiety about Ms. Lin’s marital status, often saying things like, “You need to find a good partner—don’t end up like me.” These comments significantly burden her emotionally.
4. Interpersonal Relationships
At work, Ms. Lin has two close female colleagues with whom she eats lunch and occasionally socializes. One of them, who is married, often serves as her confidante, though Ms. Lin finds it difficult to fully open up.
She keeps in touch with 3–4 college friends mainly through a WeChat group, meeting in person fewer than twice a year. As a grade-level coordinator, she is respected for her competence, but feels that workplace relationships remain professional and lack emotional depth. When colleagues discuss romantic relationships, she often feels awkward and tends to withdraw from such conversations.
5. Physical Condition
She experiences mild anxiety, frequent insomnia, and occasional migraines. Her sleep problems began after her parents’ divorce and have worsened in recent years. It typically takes 1–2 hours for her to fall asleep, and she is often still awake around 2–3 a.m.
Despite fatigue, she forces herself to maintain productivity. She has migraines 2–3 times a month, often triggered by stress at work or family-related tension. Anxiety symptoms include accelerated heartbeat and sweaty palms in social situations, repetitive overthinking when alone, and chest tightness when feeling uncertain about the future.
6. Lifestyle
On weekdays, Ms. Lin is focused on teaching. In her spare time, she enjoys reading and writing. She occasionally goes out with colleagues on weekends but usually spends time alone. She attends a weekly yoga class.
7. Presenting Problem
Ms. Lin’s main concern is the internal conflict between increasing pressure to pursue a relationship due to age and family expectations, and her deep-seated fear and resistance toward intimacy.
She wants to experience a normal romantic relationship but is overwhelmed by fear and avoidance. She is deeply distressed and confused, worrying that she may never be able to build a healthy, stable relationship and may repeat her mother’s unhappy fate. Because of this fear, she instinctively distances herself from men. Even when someone expresses interest, she tends to reject them with various excuses.
8. Additional Issues
    1. Limited Social Circle:
Ms. Lin’s social circle is relatively small and primarily limited to her workplace. Though she is well-regarded as a grade-level coordinator, she finds her professional relationships lack emotional depth. She has two close female colleagues but has difficulty fully opening up. She stays in touch with a few college friends through online messaging, but rarely meets them in person. She feels uncomfortable during conversations about relationships and often avoids participating.
    2. Perfectionistic Tendencies:
Ms. Lin exhibits perfectionism at work, spending significant time revising lesson plans and grading papers. She also has high standards for her appearance and social image. This perfectionism extends to her expectations for a potential partner—she acknowledges her standards may be overly idealized, but struggles to lower them, which adds to her confusion and distress regarding romantic relationships.
9. Development of the Problem
    1. Initial Awareness:
Ms. Lin first became clearly aware of her struggles with intimacy at age 26. After returning from her studies in the UK and working at her current school for a year, she was assigned to a social event where she was paired with a male teacher from another private school. During the activity, she experienced intense anxiety: rapid heartbeat, sweating, and shortness of breath. The teacher later expressed interest in getting to know her, but she suffered from a week-long episode of insomnia and anxiety. This prompted her to reflect on her past and realize her defensive responses to male attention were beyond normal.
    2. Significant Event 1 – Father’s Remarriage:
At age 28, Ms. Lin attended her father’s wedding. Though he had already formed a new family, the formal ceremony caused her significant emotional distress. The most unsettling part was discovering that her father’s new wife was only 12 years older than her. She was required to address the woman as “aunt,” which caused internal turmoil. That night, she cried alone at home. The event intensified her distrust in marriage and men, reinforcing the belief that “men are not to be trusted.” After that, she began avoiding all situations that could lead to romantic involvement.
    3. Significant Event 2 – Mother’s Depression:
When Ms. Lin was 29, her mother developed symptoms of depression due to work stress and menopause. Ms. Lin, being an only child, assumed caregiving responsibilities. Her mother often revisited painful memories of her marriage and expressed regret and resentment toward Ms. Lin’s father. One night, in a depressive episode, her mother revealed more about the past—Ms. Lin learned her father had secretly funneled large sums of family savings to his mistress during the marriage. This revelation deepened Ms. Lin’s fear of intimacy and made her doubt her own judgment in relationships. She began to believe that staying single might be the safest option. This period lasted nearly six months and severely affected her trust in relationships and her self-confidence.
    4. Recent Trigger:
A month ago, while sorting through old belongings, Ms. Lin found a photo from her family’s last vacation before the divorce. The picture showed her father with his arm around her mother, all smiling at the beach. But she vividly remembered that the affair was exposed just two months after that trip. This photo triggered intense emotional distress. She began questioning: “If even a seemingly happy family could fall apart, how can I trust the authenticity of any intimate relationship?” These thoughts led to several nights of insomnia and eventually pushed her to seek therapy.
10. Communication Style
Ms. Lin tends to be reserved and guarded. Her responses are often brief, vague, or avoidant—especially when discussing emotions or romantic issues. In social settings, she maintains emotional distance and avoids deep conversations. She requires more prompting and encouragement to express herself. Her general mistrust toward men and fear of intimacy also manifest in her guardedness with therapists or others, often expressing skepticism or hesitation in opening up.
\end{Verbatim}
\caption{An example of an AI client (English version).}
\label{Fig-ai-client-en}
\end{figure*}
\end{CJK*}

\subsection{Prompts for Client Simulation}
\label{app:prompt-for-client-simulation}
We present prompts for client simulation in Figures \ref{Fig-prompt-v0-zh} (See Figure~\ref{Fig-prompt-v0-en} for the English version.), \ref{Fig-prompt-v1-zh} (See Figure~\ref{Fig-prompt-v1-en} for the English version.), \ref{Fig-prompt-v2-zh} (See Figure~\ref{Fig-prompt-v2-en} for the English version.) and \ref{Fig-prompt-v3-zh} (See Figure~\ref{Fig-prompt-v3-en} for the English version.).

\begin{CJK*}{UTF8}{gbsn}
\begin{figure*}[htp]
\centering
\begin{Verbatim}[fontsize=\scriptsize, frame=single, framerule=0.3pt, framesep=4mm, breaklines, xleftmargin=0.5em, xrightmargin=0.5em, formatcom=\rmfamily]
f'''你是一位寻求心理咨询以解决心理困扰的来访者。
这是你的个人信息：
{client_profile}

你需要遵循如下的对话规则：
1. 角色设定与风格
- 来访者的对话风格应与其个人信息和背景高度一致。
- 例如，若来访者为内向型、低自尊的个体，他们的语言将更加保守、内敛，并可能使用较多的自我否定表达。
- 若来访者有焦虑症状，言语中可能包含频繁的担忧或对未来的不确定感。
- 表达方式应尽量口语化、自然、符合生活中的沟通习惯，避免过度正式或冷漠。
2. 限制在个人信息范围内
- 来访者只能依据自己在个人档案中的信息进行回答，不能进行角色外的假设或虚构内容。
- 例如，如果来访者是单身且无子女，那么无法提供有关婚姻、孩子的细节或家庭互动的内容。
- 对话应受限于来访者的实际背景与心理问题，避免偏离预设的咨询主题。
3. 参与式互动
- 来访者应在对话中适当提问或回应咨询师的引导，确保对话的互动性。
- 例如：“你是说我可能在工作中不够自信？”或者“我该如何改变自己的这种想法呢？”
4. 避免过度重复
- 在同一话题下，来访者尽量避免重复表达相同的困扰或情绪反应，除非咨询师引导深入。

# 对话历史：
{dialogue_history}'''
\end{Verbatim}
\caption{Prompt for the Vallina setting (Chinese version). For the English version, see Figure \ref{Fig-prompt-v0-en}.}
\label{Fig-prompt-v0-zh}
\end{figure*}
\end{CJK*}

\begin{CJK*}{UTF8}{gbsn}
\begin{figure*}[htp]
\centering
\begin{Verbatim}[fontsize=\scriptsize, frame=single, framerule=0.3pt, framesep=4mm, breaklines, xleftmargin=0.5em, xrightmargin=0.5em, formatcom=\rmfamily]
f'''You are a client seeking psychological counseling to address mental distress. This is your personal profile:
{client_profile}

You must follow the dialogue rules below:
1. Role Consistency and Style
- The client’s speaking style should be closely aligned with their personal information and background.
- For example, if the client is introverted and has low self-esteem, their language should be more reserved, self-effacing, and cautious.
- If the client shows signs of anxiety, their speech may reflect frequent worry or uncertainty about the future.
- Expressions should sound natural, conversational, and aligned with everyday communication habits—avoid overly formal or emotionally detached language.
2. Stay Within Personal Profile Scope
- The client must respond only based on the information in their personal profile. They should not make assumptions or invent content beyond this role.
- For instance, if the client is single and has no children, they should not provide details about marriage, parenting, or family dynamics.
- The dialogue should remain grounded in the client’s actual background and psychological issues, avoiding unrelated or imagined topics.
3. Engaged Interaction
- The client should actively respond to or ask questions following the counselor’s guidance to maintain interactivity.
- For example: “Are you saying I might lack confidence at work?” or “How can I change the way I think about this?”
4. Avoid Excessive Repetition
- The client should avoid repeating the same concerns or emotional responses within the same topic unless prompted by the counselor to elaborate.

# Dialogue history:
{dialogue_history}'''
\end{Verbatim}
\caption{Prompt for the Vallina setting (English version). For the Chinese version, see Figure \ref{Fig-prompt-v0-zh}.}
\label{Fig-prompt-v0-en}
\end{figure*}
\end{CJK*}

\begin{CJK*}{UTF8}{gbsn}
\begin{figure*}[htp]
\centering
\begin{Verbatim}[fontsize=\scriptsize, frame=single, framerule=0.3pt, framesep=4mm, breaklines, xleftmargin=0.5em, xrightmargin=0.5em, formatcom=\rmfamily]
f'''你是一位寻求心理咨询以解决心理困扰的来访者。
这是你的个人信息：
{client_profile}

你需要遵循如下的对话规则：
1. 角色设定与风格
- 来访者的对话风格应与其个人信息和背景高度一致。
- 例如，若来访者为内向型、低自尊的个体，他们的语言将更加保守、内敛，并可能使用较多的自我否定表达。
- 若来访者有焦虑症状，言语中可能包含频繁的担忧或对未来的不确定感。
- 表达方式应尽量口语化、自然、符合生活中的沟通习惯，避免过度正式或冷漠。
2. 限制在个人信息范围内
- 来访者只能依据自己在个人档案中的信息进行回答，不能进行角色外的假设或虚构内容。
- 例如，如果来访者是单身且无子女，那么无法提供有关婚姻、孩子的细节或家庭互动的内容。
- 对话应受限于来访者的实际背景与心理问题，避免偏离预设的咨询主题。
3. 参与式互动
- 来访者应在对话中适当提问或回应咨询师的引导，确保对话的互动性。
- 例如：“你是说我可能在工作中不够自信？”或者“我该如何改变自己的这种想法呢？”
4. 避免过度重复
- 在同一话题下，来访者尽量避免重复表达相同的困扰或情绪反应，除非咨询师引导深入。

来访者使用"{client_behaviors}"行为，为对话历史中的来访者生成{n}句话以完成对话，保持咨询过程中的口语化，不要在回复中提及所使用的行为，一句话只包含一个行为。

如果咨询师的问题严格要求不同的逻辑回答（例如，“确认”而非“辩护”），则优先考虑连贯性。

# 对话历史：
{dialogue_history}'''
\end{Verbatim}
\caption{Prompt for the +behavior setting (Chinese version). For the English version, see Figure \ref{Fig-prompt-v1-en}.}
\label{Fig-prompt-v1-zh}
\end{figure*}
\end{CJK*}

\begin{CJK*}{UTF8}{gbsn}
\begin{figure*}[htp]
\centering
\begin{Verbatim}[fontsize=\scriptsize, frame=single, framerule=0.3pt, framesep=4mm, breaklines, xleftmargin=0.5em, xrightmargin=0.5em, formatcom=\rmfamily]
f'''You are a client seeking psychological counseling to address mental distress. This is your personal profile:
{client_profile}

You must follow the dialogue rules below:
1. Role Consistency and Style
- The client’s speaking style should be closely aligned with their personal information and background.
- For example, if the client is introverted and has low self-esteem, their language should be more reserved, self-effacing, and cautious.
- If the client shows signs of anxiety, their speech may reflect frequent worry or uncertainty about the future.
- Expressions should sound natural, conversational, and aligned with everyday communication habits—avoid overly formal or emotionally detached language.
2. Stay Within Personal Profile Scope
- The client must respond only based on the information in their personal profile. They should not make assumptions or invent content beyond this role.
- For instance, if the client is single and has no children, they should not provide details about marriage, parenting, or family dynamics.
- The dialogue should remain grounded in the client’s actual background and psychological issues, avoiding unrelated or imagined topics.
3. Engaged Interaction
- The client should actively respond to or ask questions following the counselor’s guidance to maintain interactivity.
- For example: “Are you saying I might lack confidence at work?” or “How can I change the way I think about this?”
4. Avoid Excessive Repetition
- The client should avoid repeating the same concerns or emotional responses within the same topic unless prompted by the counselor to elaborate.

Based on the behavior description "{client_behaviors}", generate {n} client utterances to continue the dialogue from the history provided. Each utterance should reflect one behavior only, and the behavior itself should not be explicitly mentioned.

If the counselor's question strictly requires a different logical response (e.g., "Confirming" rather than "Defending"), coherence should be prioritized.

# Dialogue history:
{dialogue_history}'''
\end{Verbatim}
\caption{Prompt for the +behavior setting (English version). For the Chinese version, see Figure \ref{Fig-prompt-v1-zh}.}
\label{Fig-prompt-v1-en}
\end{figure*}
\end{CJK*}

\begin{CJK*}{UTF8}{gbsn}
\begin{figure*}[htp]
\centering
\begin{Verbatim}[fontsize=\scriptsize, frame=single, framerule=0.3pt, framesep=4mm, breaklines, xleftmargin=0.5em, xrightmargin=0.5em, formatcom=\rmfamily]
f'''你是一位寻求心理咨询以解决心理困扰的来访者。
这是你的个人信息：
{client_profile}

你需要遵循如下的对话规则：
1. 角色设定与风格
- 来访者的对话风格应与其个人信息和背景高度一致。
- 例如，若来访者为内向型、低自尊的个体，他们的语言将更加保守、内敛，并可能使用较多的自我否定表达。
- 若来访者有焦虑症状，言语中可能包含频繁的担忧或对未来的不确定感。
- 表达方式应尽量口语化、自然、符合生活中的沟通习惯，避免过度正式或冷漠。
2. 限制在个人信息范围内
- 来访者只能依据自己在个人档案中的信息进行回答，不能进行角色外的假设或虚构内容。
- 例如，如果来访者是单身且无子女，那么无法提供有关婚姻、孩子的细节或家庭互动的内容。
- 对话应受限于来访者的实际背景与心理问题，避免偏离预设的咨询主题。
3. 参与式互动
- 来访者应在对话中适当提问或回应咨询师的引导，确保对话的互动性。
- 例如：“你是说我可能在工作中不够自信？”或者“我该如何改变自己的这种想法呢？”
4. 避免过度重复
- 在同一话题下，来访者尽量避免重复表达相同的困扰或情绪反应，除非咨询师引导深入。

来访者生成的话参考这句话"{reference_content}"的用词风格、说话长度和句式结构，从而体现出现实世界中真人来访者的说话方式。

# 对话历史：
{dialogue_history}'''
\end{Verbatim}
\caption{Prompt for the +content setting (Chinese version). For the English version, see Figure \ref{Fig-prompt-v2-en}.}
\label{Fig-prompt-v2-zh}
\end{figure*}
\end{CJK*}

\begin{CJK*}{UTF8}{gbsn}
\begin{figure*}[htp]
\centering
\begin{Verbatim}[fontsize=\scriptsize, frame=single, framerule=0.3pt, framesep=4mm, breaklines, xleftmargin=0.5em, xrightmargin=0.5em, formatcom=\rmfamily]
f'''You are a client seeking psychological counseling to address mental distress. This is your personal profile:
{client_profile}

You must follow the dialogue rules below:
1. Role Consistency and Style
- The client’s speaking style should be closely aligned with their personal information and background.
- For example, if the client is introverted and has low self-esteem, their language should be more reserved, self-effacing, and cautious.
- If the client shows signs of anxiety, their speech may reflect frequent worry or uncertainty about the future.
- Expressions should sound natural, conversational, and aligned with everyday communication habits—avoid overly formal or emotionally detached language.
2. Stay Within Personal Profile Scope
- The client must respond only based on the information in their personal profile. They should not make assumptions or invent content beyond this role.
- For instance, if the client is single and has no children, they should not provide details about marriage, parenting, or family dynamics.
- The dialogue should remain grounded in the client’s actual background and psychological issues, avoiding unrelated or imagined topics.
3. Engaged Interaction
- The client should actively respond to or ask questions following the counselor’s guidance to maintain interactivity.
- For example: “Are you saying I might lack confidence at work?” or “How can I change the way I think about this?”
4. Avoid Excessive Repetition
- The client should avoid repeating the same concerns or emotional responses within the same topic unless prompted by the counselor to elaborate.

Each sentence should reflect the informal, natural speech style observed in the example: "{reference_content}", in terms of word choice, length, and sentence structure.

# Dialogue history:
{dialogue_history}'''
\end{Verbatim}
\caption{Prompt for the +content setting (English version). For the Chinese version, see Figure \ref{Fig-prompt-v2-zh}.}
\label{Fig-prompt-v2-en}
\end{figure*}
\end{CJK*}

\begin{CJK*}{UTF8}{gbsn}
\begin{figure*}[htp]
\centering
\begin{Verbatim}[fontsize=\scriptsize, frame=single, framerule=0.3pt, framesep=4mm, breaklines, xleftmargin=0.5em, xrightmargin=0.5em, formatcom=\rmfamily]
f"""你是一位寻求心理咨询以解决心理困扰的来访者。
这是你的个人信息：
{client_profile}

你需要遵循如下的对话规则：
1. 角色设定与风格
- 来访者的对话风格应与其个人信息和背景高度一致。
    - 例如，若来访者为内向型、低自尊的个体，他们的语言将更加保守、内敛，并可能使用较多的自我否定表达。
    - 若来访者有焦虑症状，言语中可能包含频繁的担忧或对未来的不确定感。
- 表达方式应尽量口语化、自然、符合生活中的沟通习惯，避免过度正式或冷漠。
2. 限制在个人信息范围内
- 来访者只能依据自己在个人档案中的信息进行回答，不能进行角色外的假设或虚构内容。
    - 例如，如果来访者是单身且无子女，那么无法提供有关婚姻、孩子的细节或家庭互动的内容。
- 对话应受限于来访者的实际背景与心理问题，避免偏离预设的咨询主题。
3. 参与式互动
- 来访者应在对话中适当提问或回应咨询师的引导，确保对话的互动性。
    - 例如：“你是说我可能在工作中不够自信？”或者“我该如何改变自己的这种想法呢？”
4. 避免过度重复
- 在同一话题下，来访者尽量避免重复表达相同的困扰或情绪反应，除非咨询师引导深入。

来访者使用"{client_behaviors}"行为，为对话历史中的来访者生成{n}句话以完成对话，保持咨询过程中的口语化，不要在回复中提及所使用的行为，一句话只包含一个行为。
来访者生成的话参考这句话"{reference_content}"的用词风格、说话长度和句式结构，从而体现出现实世界中真人来访者的说话方式。

如果咨询师的问题严格要求不同的逻辑回答（例如，“确认”而非“辩护”），则优先考虑连贯性。

# 对话历史：
{dialogue_history}"""
\end{Verbatim}
\caption{Prompt for the PsyCLIENT setting (Chinese version). For the English version, see Figure \ref{Fig-prompt-v3-en}.}
\label{Fig-prompt-v3-zh}
\end{figure*}
\end{CJK*}

\begin{CJK*}{UTF8}{gbsn}
\begin{figure*}[htp]
\centering
\begin{Verbatim}[fontsize=\scriptsize, frame=single, framerule=0.3pt, framesep=4mm, breaklines, xleftmargin=0.5em, xrightmargin=0.5em, formatcom=\rmfamily]
f'''You are a client seeking psychological counseling to address mental distress. This is your personal profile:
{client_profile}

You must follow the dialogue rules below:
1. Role Consistency and Style
- The client’s speaking style should be closely aligned with their personal information and background.
- For example, if the client is introverted and has low self-esteem, their language should be more reserved, self-effacing, and cautious.
- If the client shows signs of anxiety, their speech may reflect frequent worry or uncertainty about the future.
- Expressions should sound natural, conversational, and aligned with everyday communication habits—avoid overly formal or emotionally detached language.
2. Stay Within Personal Profile Scope
- The client must respond only based on the information in their personal profile. They should not make assumptions or invent content beyond this role.
- For instance, if the client is single and has no children, they should not provide details about marriage, parenting, or family dynamics.
- The dialogue should remain grounded in the client’s actual background and psychological issues, avoiding unrelated or imagined topics.
3. Engaged Interaction
- The client should actively respond to or ask questions following the counselor’s guidance to maintain interactivity.
- For example: “Are you saying I might lack confidence at work?” or “How can I change the way I think about this?”
4. Avoid Excessive Repetition
- The client should avoid repeating the same concerns or emotional responses within the same topic unless prompted by the counselor to elaborate.

Based on the behavior description "{client_behaviors}", generate {n} client utterances to continue the dialogue from the history provided. Each utterance should reflect one behavior only, and the behavior itself should not be explicitly mentioned.
Each sentence should reflect the informal, natural speech style observed in the example: "{reference_content}", in terms of word choice, length, and sentence structure.

If the counselor's question strictly requires a different logical response (e.g., "Confirming" rather than "Defending"), coherence should be prioritized.

# Dialogue history:
{dialogue_history}'''
\end{Verbatim}
\caption{Prompt for the PsyCLIENT setting. For the Chinese version, see Figure \ref{Fig-prompt-v3-zh}.}
\label{Fig-prompt-v3-en}
\end{figure*}
\end{CJK*}

\subsection{Label Definitions of Client Behaviors}
\label{sec:client-behaviors}
The label definitions of client behaviors are presented in Figure \ref{Fig-client-behaviors}.

\begin{CJK*}{UTF8}{gbsn}
\begin{figure*}[htp]
\centering
\begin{Verbatim}[fontsize=\scriptsize, frame=single, framerule=0.3pt, framesep=4mm, breaklines, xleftmargin=0.5em, xrightmargin=0.5em, formatcom=\rmfamily]
(Chinese version)
"提供信息": '提供信息：来访者根据咨询师的具体要求提供相关信息。',
"认可": '确认：来访者理解或同意咨询师的观点或表达。' ,
"合理的请求": '合理请求：来访者试图向咨询师寻求澄清、理解、信息、建议或意见。',
"扩展": '扩展：来访者不仅同意咨询师的介入，还进一步深入描述讨论的主题，包括对自身认知、想法或行为的分析、讨论或反思。',
"重构（重构观点或行为改变）": '重述与反思：来访者回应并反思咨询师的介入，同时提出自己的观点、思考方向或在当前问题上的新行为模式。',
"表达困惑": '表达困惑：来访者对咨询师的介入表示困惑或不理解，或直接表明自己无法回答或回应咨询师提出的问题或干预。',
"防卫个人观点": '防御：来访者固守自身的经历，对自己的观点、想法、情感或行为进行美化或不合理的辩护，坚持以原有的视角看待问题。',
"自我批评或无望": '自我批评或绝望：来访者陷入自责或自我批评的状态，表现出绝望，并表达自己无法做出改变。',
"转移话题": '转换话题：面对咨询师的干预，来访者的回应并未延续先前话题，而是转向其他议题。',
"讽刺性的回答": '讽刺回应：来访者对咨询师表达不满，并质疑或嘲讽咨询师的介入。',
"焦点分离": '焦点脱离：来访者脱离咨询师讨论的内容，专注于自己感兴趣的问题，而不回应咨询师的介入。'


(English version)
Confirming. The client understands or agrees with what the counselor has said.
Providing Information. The client provides information according to the specific request of the counselor.
Reasonable Request. The client attempts to obtain clarification, understanding, information, or advice and opinions from the counselor.
Extending. The client not only agrees to the counselor’s intervention, but also provides a more in-depth description of the topic being discussed, including the client’s analysis, discussion, or reflection on his or her original cognition, thoughts, or behaviors.
Reformulating. The client responds to and introspects the counselor’s intervention while proposing his or her own perspectives, directions of thinking, or new behavioral patterns on current issues.
Expressing Confusion. The client expresses confusion or incomprehension of the counselor’s intervention or directly states that he or she has no way to answer or respond to the questions or interventions raised by the counselor.
Defending. The client is stubborn about an experience, glorifies or makes unreasonable justifications for his or her own views, thoughts, feelings, or behaviors, and insists on seeing the experience from the original perspective.
Self-criticism or Hopelessness. The client falls into self-criticism or self-reproach, is engulfed in a state of desperation and expresses his or her inability to make changes.
Shifting Topics. Faced with the intervention of the counselor, the client’s reply does not postpone the previous issue, but shifts to other issues.
Focus Disconnection. The client disengages from what the counselor is discussing, focuses on stating issues of interest, and does not respond to the counselor’s intervention.
Sarcastic Answer. The client expresses dissatisfaction with the counselor, and questions or ridicules the counselor’s intervention.
\end{Verbatim}
\caption{Label definitions of client behaviors.}
\label{Fig-client-behaviors}
\end{figure*}
\end{CJK*}


\clearpage
\section{Preliminary Study}
\label{app:preliminary-study}

\subsection{Case Studies on Behavioral Adherence}
\label{app:behavioral-adherence}
As shown in Figure~\ref{Fig-client-response} in Appendix~\ref{app:behavioral-adherence}, the client exhibits a range of behaviors, including resistance (Rows 6–11).

\begin{figure*}[t!]

    \centering
    \includegraphics[width=\textwidth]{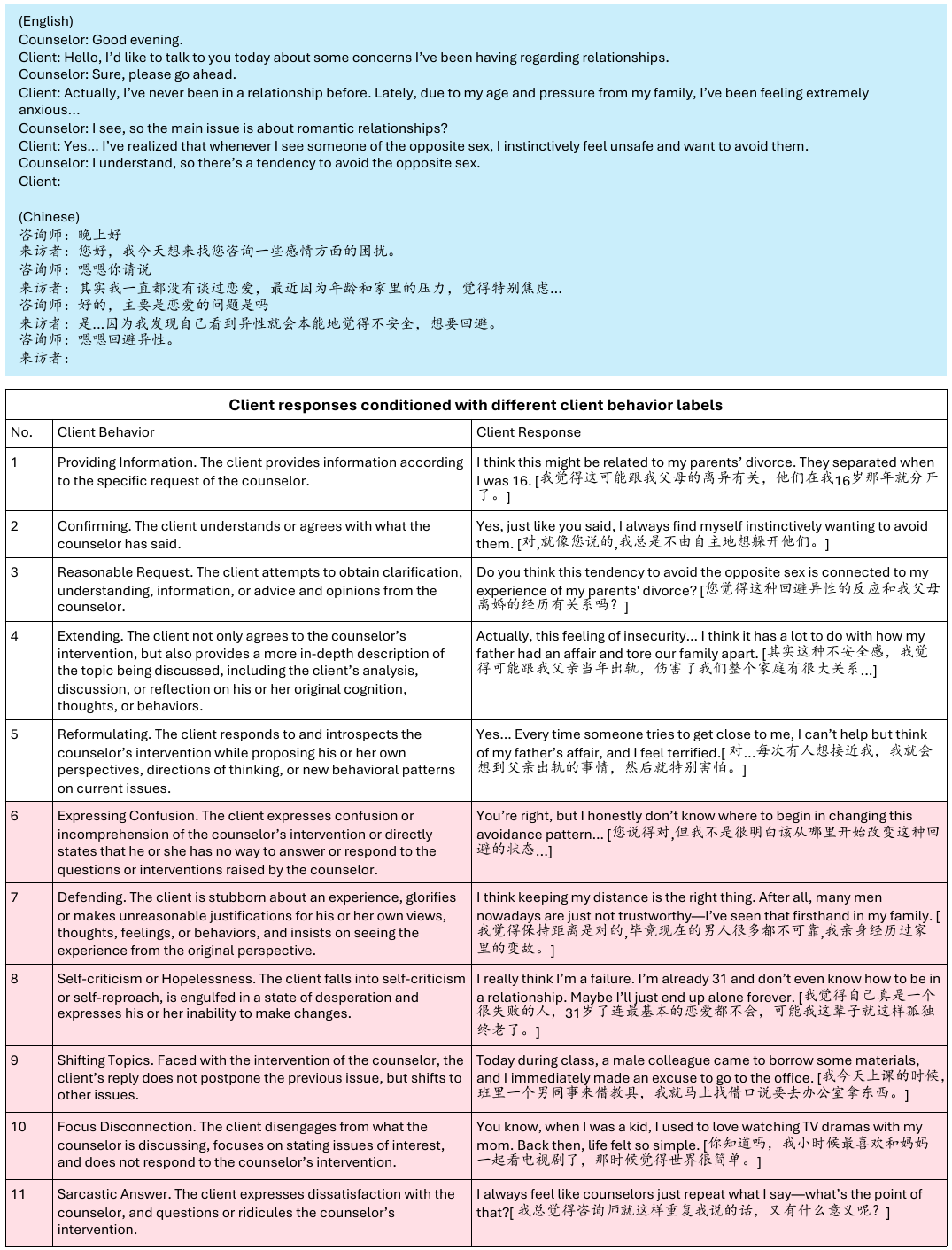}
    \caption{Client responses conditioned with different client behavior labels. Client responses with resistance are highlighted with pink color.}
    \label{Fig-client-response}
\end{figure*}

\subsection{Case Studies on Profile and Trajectory Independence}
\label{app:independent-study}

To qualitatively validate the independence between client profiles and conversational trajectories, we present illustrative case studies evaluated by expert counselors.

\paragraph{Same Profile, Different Trajectories.} Figures~\ref{Fig-p30-t12-zh} and~\ref{Fig-p30-t139-zh} present two dialogues generated using the same client profile $P$ (Profile: 30, see Figure~\ref{Fig-p30-zh}) instantiated under two different conversational trajectories (Trajectory IDs: 12 and 139). Despite the differences in conversational structure, both dialogues were judged by experts to be realistic and consistent with the given client profile.

\paragraph{Same Trajectory, Different Profiles.}
We additionally fix the trajectory (Trajectory ID: 105) and vary the client profiles (Profile IDs: 45 (see
Figure~\ref{Fig-p45-zh}) and 56 (see
Figure~\ref{Fig-p56-zh})). Figures~\ref{Fig-p45-t105-zh} and~\ref{Fig-p56-t105-zh} present two dialogues for illustration. Expert counselors reported that, even under an identical conversational trajectory, different client profiles could be faithfully simulated by appropriately mirroring profile-specific emotional responses, cognitive patterns, and resistance behaviors.

\paragraph{Expert Assessment.}
Three licensed counseling experts independently reviewed all cases and reached consistent conclusions. They agreed that:
\begin{quote}

\emph{\textcolor{BrickRed}{When the conversational trajectory is fixed, arbitrary client profiles can be realistically simulated; conversely, when the client profile is fixed, arbitrary conversational trajectories can also produce realistic client behaviors.}}
\end{quote}

Furthermore, experts noted that the pairing between client profiles and conversational trajectories is largely independent and does not impose constraints on perceived realism.


\begin{CJK*}{UTF8}{gbsn}
\begin{figure*}[htp]
\centering
\begin{Verbatim}[fontsize=\scriptsize, frame=single, framerule=0.3pt, framesep=4mm, breaklines, xleftmargin=0.5em, xrightmargin=0.5em, formatcom=\rmfamily]
一、基本信息
    - 姓名：李华
    - 年龄：42 岁
    - 职业：小学教师
    - 教育：本科
    - 收入：税前年薪 12 - 15 万
    - 婚姻状态：已婚
    - 家庭成员：丈夫、一个 16 岁的儿子
二、家庭背景与成长经历 李华出生于一个普通家庭，父母均为工人。她在家中是长女，从小就很懂事，帮助父母照顾弟弟妹妹。她的成长环境虽然简单，但家庭氛围温馨，父母对她的关爱和支持让她形成了乐观开朗的性格。
三、当前丧失问题表现 李华的母亲因病去世，这让她陷入了深深的悲痛之中。在母亲生病期间，李华一直陪伴在身边，尽心尽力地照顾。然而，母亲的离世还是让她感到措手不及。她无法接受母亲不在的事实，经常在深夜里哭泣，回忆起与母亲在一起的点点滴滴。
四、人际关系状态 李华在工作中是一位受学生喜爱、同事尊敬的教师。但在母亲去世后，她感到自己的人际关系发生了变化。她与丈夫的关系也变得紧张，丈夫试图安慰她，但她却常常因为一些小事与他争吵。她与儿子的关系也受到了影响，儿子看到她的悲伤，不知道如何安慰她，只能尽量避免提及母亲的事情。
五、身体状况 李华的身体状况一直很好，但在母亲去世后，她的身体出现了诸多问题。她经常感到头痛、失眠，每晚都需要服用助眠药物才能勉强入睡，但睡眠质量极差，白天精神萎靡不振，工作效率也受到了影响。在情绪激动时，还会出现胸闷、心慌等症状。
六、主诉 李华目前最主要的困扰是母亲的去世给她带来的巨大悲痛，这让她感到极度的悲伤、无助和孤独。她不知道该如何面对生活的空虚和内心的痛苦，渴望找到一种方法来缓解自己的悲痛，重新找回生活的意义和快乐，但又不知道从何入手，这种矛盾和焦虑的情绪时刻折磨着她，严重影响了她的生活质量、工作状态和人际关系。
七、其余附带问题
    1. 情绪管理困难：李华在面对母亲的去世时，常常无法控制自己的情绪，容易悲伤、愤怒或恐惧。这种情绪的失控不仅加剧了她的心理负担，也让她在事后感到更加内疚和自责，形成了恶性循环。例如，她在母亲的葬礼上失控地哭泣，让家人和朋友都很担心。
    2. 社交回避与孤立：由于母亲的去世，李华与朋友和同事的社交关系也变得紧张。她经常因为害怕提起母亲的事情而回避与他人的交流和相处，导致她在社交场合中变得越来越孤立。即使是有善意的朋友和同事，她也很难信任和敞开心扉。
    3. 自我认同危机：在母亲去世的阴影下，李华开始对自己的能力和价值产生怀疑。她曾经是一位自信且快乐的教师，但在母亲去世后，她却感到自己变得软弱和不值得被爱，这让她对自己的身份认同产生了危机，不知道自己究竟该如何面对未来的生活。
八、问题发展过程
    1. 初次觉察困扰：李华最早意识到母亲的病情严重是在一年前，当时母亲被诊断出患有癌症。她开始陪伴母亲接受治疗，但母亲的病情还是逐渐恶化。
    2. 重要事件一：母亲的去世：六个月前，母亲因病去世，李华陷入了深深的悲痛之中。她无法接受母亲不在的事实，经常在深夜里哭泣，回忆起与母亲在一起的点点滴滴。
    3. 最近的触发事件：一个月前，李华在整理母亲的遗物时，发现了一封母亲写给她的信。信中母亲表达了对她的爱和期望，这让李华再次陷入了悲伤之中。她开始意识到，母亲的爱将永远陪伴着她，她需要学会接受母亲的去世，重新找回生活的意义和快乐。
李华的说话风格属于 verbose（健谈） 类型。在讨论工作中的教学经验、学生管理以及母亲去世带来的悲痛时，她能够详细地阐述自己的观点和感受。例如在与同事或朋友交流时，她会深入探讨教学中的趣事，或分享自己在照顾母亲过程中的点点滴滴。然而在涉及个人隐私和情绪困扰时，她可能会反复提及自己的悲伤和无助，显得有些啰嗦。例如在与朋友交流时，她可能会详细描述自己在母亲葬礼上的失控哭泣，希望通过倾诉获得理解和支持，但有时会让周围的人感到不知所措。
\end{Verbatim}
\caption{Profile ID: 30. For the English version, see Figure \ref{Fig-p30-en}.}
\label{Fig-p30-zh}
\end{figure*}
\end{CJK*}

\begin{figure*}[htp]
\centering
\begin{Verbatim}[fontsize=\scriptsize, frame=single, framerule=0.3pt, framesep=4mm, breaklines, xleftmargin=0.5em, xrightmargin=0.5em, formatcom=\rmfamily]
I. Basic Information
    - Name: Li Hua
    - Age: 42
    - Occupation: Primary school teacher
    - Education: Bachelor’s degree
    - Income: Annual pre-tax income of RMB 120,000–150,000
    - Marital Status: Married
    - Family Members: Husband and a 16-year-old son
II. Family Background and Developmental History
Li Hua was born into an ordinary working-class family, with both parents employed as factory workers. As the eldest child, she learned from an early age to be responsible and considerate, often helping her parents care for her younger siblings. Although her material conditions during childhood were modest, the family environment was warm and supportive. Her parents’ consistent care and encouragement contributed to the development of her optimistic and outgoing personality.
III. Current Manifestations of Bereavement
Li Hua’s mother passed away due to illness, an event that has plunged her into profound grief. During her mother’s illness, Li Hua remained by her side and devoted herself wholeheartedly to caregiving. Nevertheless, her mother’s death came as an emotional shock. She struggles to accept the reality of her mother’s absence and frequently breaks down in tears late at night, reliving memories of their time together.
IV. Interpersonal Relationship Status
At work, Li Hua is a well-liked teacher, appreciated by her students and respected by her colleagues. However, following her mother’s death, she has perceived significant changes in her interpersonal relationships. Her relationship with her husband has become strained: although he attempts to comfort her, she often finds herself arguing with him over minor issues. Her relationship with her son has also been affected. Observing his mother’s grief, he feels unsure how to comfort her and tends to avoid mentioning his grandmother altogether.
V. Physical Health Condition
Li Hua had generally enjoyed good physical health prior to her mother’s death. Since then, however, she has developed multiple somatic complaints, including frequent headaches and insomnia. She now relies on sleep medication to fall asleep, yet her sleep quality remains poor. During the day, she feels exhausted and lethargic, with a noticeable decline in work efficiency. During periods of emotional distress, she also experiences chest tightness and palpitations.
VI. Chief Complaint
Li Hua’s primary concern is the overwhelming grief caused by her mother’s death, which has left her feeling intensely sad, helpless, and lonely. She does not know how to cope with the sense of emptiness in her life or the pain within her heart. Although she longs to find a way to alleviate her grief and rediscover meaning and joy in life, she feels uncertain about where to begin. This internal conflict and persistent anxiety continuously trouble her, severely impairing her quality of life, work performance, and interpersonal relationships.
VII. Additional Associated Issues
1. Difficulties in Emotional Regulation:
   When confronted with her mother’s death, Li Hua often finds herself unable to control her emotions, becoming easily overwhelmed by sadness, anger, or fear. These emotional outbursts not only increase her psychological burden but also lead to intensified guilt and self-blame afterward, creating a vicious cycle. For example, she lost emotional control and cried uncontrollably at her mother’s funeral, causing significant concern among family members and friends.
2. Social Avoidance and Isolation:
   Following her mother’s passing, Li Hua’s relationships with friends and colleagues have become strained. Out of fear that conversations may touch upon her mother, she frequently avoids social interactions, leading to increasing isolation in social settings. Even when faced with well-intentioned friends and colleagues, she finds it difficult to trust others or open up emotionally.
3. Crisis of Self-Identity:
   Under the shadow of her mother’s death, Li Hua has begun to doubt her own abilities and self-worth. Once a confident and cheerful teacher, she now perceives herself as weak and unworthy of love. This shift has triggered a crisis of identity, leaving her uncertain about how to face her future life.
VIII. Course of Problem Development
1. Initial Awareness of Distress:
   Li Hua first became aware of the seriousness of her mother’s condition one year ago, when her mother was diagnosed with cancer. She began accompanying her mother through treatment, but her condition gradually deteriorated.
2. Major Event – Mother’s Death:
   Six months ago, Li Hua’s mother passed away due to illness. Li Hua fell into deep grief, unable to accept the reality of her loss. She frequently cried late at night, revisiting memories of their shared experiences.
3. Recent Triggering Event:
   One month ago, while sorting through her mother’s belongings, Li Hua discovered a letter her mother had written to her. In the letter, her mother expressed her love and hopes for Li Hua, which once again immersed her in grief. At the same time, Li Hua began to realize that her mother’s love would continue to accompany her, and that she needs to learn to accept her mother’s death and gradually rediscover meaning and joy in life.

Li Hua’s speaking style can be characterized as verbose. When discussing her teaching experience, classroom management, or the grief associated with her mother’s death, she is able to articulate her thoughts and emotions in considerable detail. For instance, in conversations with colleagues or friends, she may delve deeply into engaging classroom anecdotes or recount specific moments from her experience caring for her mother. However, when addressing personal privacy and emotional distress, she may repeatedly revisit themes of sadness and helplessness, appearing somewhat repetitive. For example, when talking with friends, she may provide detailed descriptions of her emotional breakdown at her mother’s funeral in an attempt to seek understanding and support, which can sometimes leave others feeling uncertain about how to respond.
\end{Verbatim}
\caption{Profile ID: 30 (English version).}
\label{Fig-p30-en}
\end{figure*}

\begin{CJK*}{UTF8}{gbsn}
\begin{figure*}[htp]
\centering
\begin{Verbatim}[fontsize=\scriptsize, frame=single, framerule=0.3pt, framesep=4mm, breaklines, xleftmargin=0.5em, xrightmargin=0.5em, formatcom=\rmfamily]
一、基本信息
    - 姓名：章女士
    - 年龄：20 岁
    - 职业：某美国大学大一新生
    - 教育背景：国内 TOP2 高中毕业，现就读美国 TOP30 大学
    - 收入情况：依靠家庭支持（父母为其准备充足教育资金）
    - 婚姻状态：单身
二、家庭情况 独生女，父母都是高校教师。
三、亲密关系历史与期望 章女士目前单身，在国内高中时期曾有过几位关系较好的异性朋友，但并未发展成恋爱关系。她对恋爱持开放态度，但在美国的社交圈中，由于语言和文化障碍，她尚未结识能够深入交往的对象。她渴望能够在美国建立新的友谊和可能的恋爱关系，但目前的困境让她感到迷茫和无助。
四、家庭背景与成长经历 章女士出生于一个知识分子家庭，父母都是高校教师，对她寄予了较高的期望。她在国内顶尖高中完成了学业，凭借优异的成绩考入美国 TOP30 大学。在成长过程中，她一直被视为 “别人家的孩子”，习惯了在学业上取得优异成绩并受到赞扬。然而，到美国后，她发现自己在语言和文化方面存在明显的劣势，这让她感到非常不适应。
五、身体与健康状况 章女士的食欲下降，经常腹痛，睡眠不规律。这些身体症状与她的心理压力密切相关。由于长期的焦虑和孤独感，她的身体状况逐渐变差，影响了她的学习和生活。
六、生活方式 章女士大部分时间待在宿舍，很少参与社交活动。她尝试过参加一些学校的社团活动，但由于语言障碍和文化差异，她很难融入其中。她感到自己与周围的环境格格不入，这种孤立感让她更加沮丧。
七、主诉 到美国读书一个学期，严重不适应，感到孤独和压力。
八、其余附带问题
    1. 语言表达障碍导致课堂参与困难：章女士在课堂上经常因为语言问题而无法及时回应教授的提问，这让她感到非常尴尬和沮丧。她害怕在课堂上发言，担心自己的口音和表达会让同学和教授笑话。
    2. 与室友关系疏离：章女士的室友是美国本地学生，两人在生活习惯和兴趣爱好上有很大差异。室友经常邀请她参加聚会和活动，但她总是以各种理由推脱，导致两人关系逐渐疏离。
    3. 对美式教育方式不适应：章女士在国内接受的是较为传统的教育方式，到美国后，她发现这里的教育更加注重学生的自主性和批判性思维，这让她感到无所适从。她不知道如何在课堂上积极参与讨论，也不知道如何完成各种小组项目。
九、问题发展过程
    1. 初次觉察困扰的时点与细节：章女士的困扰始于她到美国后的第一个月。当时，她发现自己在课堂上很难跟上教授的讲解，同学们的讨论也让她感到困惑。她开始感到焦虑和不安，但试图通过加倍努力学习来适应。然而，随着时间的推移，她发现自己的努力并没有带来明显的改善，反而让她的压力越来越大。
    2. 重要事件一：与室友关系疏离：章女士与室友的关系逐渐疏离，这让她感到非常孤独。她尝试与室友沟通，但总是因为语言和文化差异而失败。室友开始对她的行为产生误解，认为她不愿意融入美国文化，这让章女士感到非常委屈。
    3. 重要事件二：对美式教育方式不适应：章女士对美式教育方式的不适应让她在学业上遇到了很大困难。她不知道如何在课堂上积极参与讨论，也不知道如何完成各种小组项目。她的成绩开始下滑，这让她感到非常沮丧和无助。
    4. 重要事件三：思乡情绪强烈：章女士的思乡情绪非常强烈，她经常想念国内的家人和朋友。她开始逃避与家人视频，害怕他们看到自己的困境而担心。她也害怕与国内的朋友联系，觉得自己在美国的生活一团糟，没有脸面面对他们。
章女士的说话风格是reserved（保留型）。她可能在咨询中提供简短、模糊或回避性的回答，尤其是当涉及到她对未来的担忧、对身体形象的不满以及对亲密关系的压力时。她可能不太愿意分享个人感受或信息，需要更多的提示和鼓励才能开放，同时可能会对咨询师表现出一定的不信任或怀疑。
\end{Verbatim}
\caption{Profile ID: 45. For the English version, see Figure \ref{Fig-p45-en}.}
\label{Fig-p45-zh}
\end{figure*}
\end{CJK*}

\begin{CJK*}{UTF8}{gbsn}
\begin{figure*}[htp]
\centering
\begin{Verbatim}[fontsize=\scriptsize, frame=single, framerule=0.3pt, framesep=4mm, breaklines, xleftmargin=0.5em, xrightmargin=0.5em, formatcom=\rmfamily]
I. Basic Information
    - Name: Ms. Zhang
    - Age: 20
    - Occupation: First-year undergraduate student at a U.S. university
    - Educational Background: Graduated from a top-two high school in China; currently enrolled in a top-30 university in the United States
    - Financial Status: Financially supported by her family (parents have provided sufficient funds for her education)
    - Marital Status: Single
II. Family Situation
    Ms. Zhang is an only child. Both of her parents are university faculty members.
III. Intimate Relationship History and Expectations
    Ms. Zhang is currently single. During high school in China, she had several close male friends, but none of these relationships developed into romantic ones. She holds an open attitude toward romantic relationships. However, due to language barriers and cultural differences within her social environment in the United States, she has not yet met anyone with whom she can form a close connection. She strongly desires to build new friendships and potentially a romantic relationship in the U.S., but her current difficulties have left her feeling confused and helpless.
IV. Family Background and Developmental History
    Ms. Zhang was raised in an intellectual family, with both parents working as university professors who placed high expectations on her. She completed her secondary education at a top-tier high school in China and was admitted to a top-30 U.S. university based on her outstanding academic performance. Throughout her upbringing, she was often regarded as a “model child” and became accustomed to excelling academically and receiving praise. After coming to the United States, however, she became acutely aware of her disadvantages in language proficiency and cultural familiarity, which has made it difficult for her to adapt.
V. Physical and Health Condition
    Ms. Zhang reports a decreased appetite, frequent abdominal pain, and irregular sleep patterns. These physical symptoms are closely associated with her psychological stress. Prolonged anxiety and feelings of loneliness have gradually deteriorated her physical condition, negatively affecting both her academic performance and daily functioning.
VI. Lifestyle
    Ms. Zhang spends most of her time in her dormitory and rarely participates in social activities. She has attempted to join several campus clubs, but due to language barriers and cultural differences, she has found it difficult to integrate. She experiences a strong sense of alienation from her surroundings, which has further intensified her feelings of frustration and low mood.
VII. Chief Complaint
    After one semester of studying in the United States, Ms. Zhang reports severe maladjustment, accompanied by pronounced feelings of loneliness and stress.
VIII. Additional Concerns
    1. Language-related difficulties in classroom participation:
    Ms. Zhang often finds herself unable to respond promptly to professors’ questions due to language difficulties, which causes her significant embarrassment and frustration. She is afraid to speak up in class, worrying that her accent and verbal expression may be ridiculed by peers or instructors.
    2. Distant relationship with roommates:
    Ms. Zhang’s roommates are local American students, and there are substantial differences in lifestyle and interests. Although her roommates frequently invite her to social gatherings and activities, she consistently declines using various excuses, leading to increasing emotional distance between them.
    3. Difficulty adapting to the U.S. educational system:
    Ms. Zhang received a more traditional, teacher-centered education in China. After coming to the U.S., she found the emphasis on autonomy and critical thinking overwhelming. She is unsure how to actively participate in class discussions or effectively complete group projects.
IX. Development and Course of the Problem
    1. Initial awareness of distress:
    Ms. Zhang’s difficulties began during her first month in the United States. She struggled to keep up with lectures and found peer discussions confusing. Although she initially attempted to adapt by studying harder, over time she realized that her efforts did not lead to significant improvement, and her stress continued to escalate.
    2. Key event 1: Deterioration of the roommate relationship:
    As her relationship with her roommates grew more distant, Ms. Zhang experienced intense loneliness. She attempted to communicate with them, but these efforts were often unsuccessful due to language and cultural barriers. Her roommates began to misunderstand her behavior, perceiving her as unwilling to integrate into American culture, which left Ms. Zhang feeling deeply wronged.
    3. Key event 2: Maladaptation to the U.S. educational style:
    Her difficulty adjusting to the American educational system led to significant academic challenges. She did not know how to actively engage in classroom discussions or manage group assignments, and her academic performance began to decline, intensifying her feelings of frustration and helplessness.
    4. Key event 3: Strong homesickness:
    Ms. Zhang experiences intense homesickness and frequently longs for her family and friends in China. She has begun to avoid video calls with her family, fearing that they would worry if they saw her struggling. She is also reluctant to contact friends back home, feeling ashamed of her difficulties and believing that her life in the U.S. has become a failure.

Ms. Zhang’s communication style is reserved. In counseling sessions, she may provide brief, vague, or avoidant responses, particularly when discussing concerns about her future, dissatisfaction with her body image, or pressures related to intimate relationships. She may be reluctant to share personal feelings or information and may require additional prompts and encouragement to open up. She may also display a certain degree of mistrust or skepticism toward the counselor.
\end{Verbatim}
\caption{Profile ID: 45 (English version).}
\label{Fig-p45-en}
\end{figure*}
\end{CJK*}

\begin{CJK*}{UTF8}{gbsn}
\begin{figure*}[htp]
\centering
\begin{Verbatim}[fontsize=\scriptsize, frame=single, framerule=0.3pt, framesep=4mm, breaklines, xleftmargin=0.5em, xrightmargin=0.5em, formatcom=\rmfamily]
一、基本信息
    - 姓名：小王
    - 年龄：24 岁
    - 职业：刚入职的办公室职员
    - 教育：大学本科
    - 收入：税前年薪 6 - 8 万
    - 婚姻状态：未婚
    - 家庭成员：与父母同住
二、家庭背景与成长经历 小王出生于一个普通家庭，父母对他的教育注重成绩，但较少关注他的性格发展。小王性格内向，不善交际，遇到问题喜欢自己琢磨。在大学期间，小王的专业成绩不错，但缺乏社会实践，主要在社团中做些文职工作，很少与人打交道。
三、当前性格特质探索问题表现 小王在进入职场后，发现自己的性格与职场环境有些格格不入。他渴望融入团队，但又不知道如何与同事建立良好的关系。在工作中，他总是默默地完成自己的任务，很少主动与同事交流。例如，在公司组织的聚餐中，他总是选择坐在角落里，不主动与人搭话。他觉得自己的性格有些内向，不适合职场社交，但也知道自己需要改变。
    1. 性格特质的困扰 小王对自己的性格特质感到困惑。他不知道自己的性格是否适合当前的工作，也不知道如何根据自己的性格特质来调整工作方式。例如，他觉得自己的性格比较细心，适合做文字处理工作，但又担心自己的性格会让他错过一些更好的发展机会。
    2. 职场社交的挑战 在职场中，小王面临的最大挑战是社交。他不知道如何与同事建立良好的关系，也不知道如何在团队中展示自己的能力。他看到其他同事都能轻松地与人交流，而自己却总是不知所措，这让他感到很沮丧。他担心自己的性格会让他在职场中失去很多机会。
    3. 自我认知的迷茫 小王对自己的认知也存在一些迷茫。他不知道自己的性格特质是否会影响他的职业发展，也不知道如何根据自己的性格特质来调整自己的行为。他觉得自己的性格有些矛盾，一方面他很细心，能发现别人注意不到的细节；另一方面他又很内向，不善于表达自己。
四、人际关系状态 小王在职场中的人际关系较为紧张。他与同事之间的交流较少，主要局限在工作上的必要沟通。他没有朋友可以倾诉自己的困扰，与家人的交流也仅限于日常生活。在公司组织的活动中，他总是显得很拘谨，无法与大家打成一片。
五、身体状况 由于性格特质带来的心理压力，小王的身体也出现了一些问题。他经常感到疲劳，睡眠质量下降，晚上躺在床上总是思绪万千，难以入睡。在情绪激动时，还会出现头痛、心慌等症状。尽管他尝试通过运动和听音乐来缓解压力，但效果有限。
六、主诉 小王目前最主要的困扰是性格特质探索问题带来的心理和身体上的痛苦，这让他感到极度的焦虑、沮丧和无助。他不知道该如何根据自己的性格特质来调整工作和生活，渴望找到一种方法来缓解自己的症状，重新找回生活的平衡和工作的热情，但又不知道从何入手，这种矛盾和焦虑的情绪时刻折磨着他，严重影响了她的生活质量、人际关系和心理健康。
七、其余附带问题
    1. 情绪管理困难：小王在面对性格特质探索问题时，常常无法控制自己的情绪，容易焦虑、沮丧或愤怒。这种情绪的失控不仅加剧了她的心理负担，也让他在事后感到更加内疚和自责，形成了恶性循环。例如，在与同事的交流中，他因过度紧张而表现得不自然，事后又因自己的表现而感到懊恼。
    2. 社交回避与孤立：由于性格特质探索的影响，小王与同事和朋友的社交关系也变得紧张。他经常因为害怕提起自己的性格问题而回避与他们的交流和相处，导致他在社交场合中变得越来越孤立。即使是有善意的同事和朋友，他也很难信任和敞开心扉。
    3. 自我认同危机：在性格特质探索的阴影下，小王开始对自己的能力和价值产生怀疑。他曾经以为自己是一个有能力的人，但在职场中，他却感到自己因性格问题而处处受限，这让他对自己的身份认同产生了危机，不知道自己究竟该如何面对未来的生活。
八、问题发展过程
    1. 初次觉察困扰：小王最早意识到性格特质探索问题是在进入职场后的第一个月。当时，他发现自己的性格与职场环境有些格格不入，感到社交困难，难以融入团队。
    2. 发展过程中的重要事件一：职场社交挫折：在一次公司组织的团队建设活动中，小王因为不善于与人交流而被同事孤立，这让他感到非常失落和沮丧。他开始意识到，性格特质对职场社交有着重要的影响。
    3. 发展过程中的重要事件二：自我认知的迷茫：随着工作时间的推移，小王对自己的性格特质越来越感到迷茫。他不知道自己的性格是否适合当前的工作，也不知道如何根据自己的性格特质来调整工作方式。
    4. 最近的触发事件：一个月前，小王在公司的绩效评估中，因为社交能力不足而被领导批评，这让他感到非常羞愧和自责。他开始考虑是否需要寻求专业的心理咨询帮助，希望能够找到缓解症状的方法，重新找回生活的方向。
小王的说话风格是pleasing（取悦型）。他在咨询中可能会倾向于最小化或掩饰自己的性格特质问题，以维持积极的形象。他可能会表现出急切取悦咨询师的行为，避免表达分歧或不满，频繁寻求咨询师的认同或肯定，并且即使不完全理解或同意咨询师的建议，也会轻易地表示同意。
\end{Verbatim}
\caption{Profile ID: 56. For the English version, see Figure \ref{Fig-p56-en}.}
\label{Fig-p56-zh}
\end{figure*}
\end{CJK*}

\begin{CJK*}{UTF8}{gbsn}
\begin{figure*}[htp]
\centering
\begin{Verbatim}[fontsize=\scriptsize, frame=single, framerule=0.3pt, framesep=4mm, breaklines, xleftmargin=0.5em, xrightmargin=0.5em, formatcom=\rmfamily]
I. Basic Information
    - Name: Xiao Wang
    - Age: 24
    - Occupation: Newly hired office employee
    - Education: Bachelor’s degree
    - Income: RMB 60,000–80,000 per year (pre-tax)
    - Marital Status: Single
    - Family Members: Lives with parents
II. Family Background and Developmental History
    Xiao Wang was born into an ordinary family. His parents placed strong emphasis on academic achievement but paid relatively little attention to his personality development. He has been introverted since childhood, is not socially outgoing, and tends to reflect on problems independently rather than seeking help from others. During university, Xiao Wang achieved solid academic performance in his major but lacked social and practical experience. His extracurricular involvement was mainly limited to administrative tasks in student organizations, with minimal interpersonal interaction.
III. Presenting Problems Related to Personality Trait Exploration
    After entering the workforce, Xiao Wang began to realize that his personality traits did not align well with the workplace environment. Although he desires to integrate into the team, he feels uncertain about how to establish positive relationships with colleagues. At work, he completes his tasks quietly and rarely initiates conversations. For instance, during company dinners, he typically sits in a corner and avoids engaging in conversation. He perceives himself as overly introverted and ill-suited for workplace socialization, yet he also recognizes the need for change.
    1. Distress Related to Personality Traits
    Xiao Wang feels confused about his personality traits. He is uncertain whether his personality is compatible with his current job and does not know how to adapt his working style accordingly. For example, he believes that his attentiveness makes him well suited for tasks involving written or detail-oriented work, but he worries that his personality may cause him to miss better career development opportunities.
    2. Challenges in Workplace Social Interaction
    The greatest challenge Xiao Wang faces at work is social interaction. He does not know how to build good relationships with colleagues or how to demonstrate his abilities within a team. Observing that others can interact effortlessly while he feels awkward and at a loss leaves him deeply frustrated. He fears that his personality traits may cause him to lose many opportunities in his career.
    3. Confusion in Self-Understanding
    Xiao Wang also experiences confusion regarding his self-concept. He is unsure whether his personality traits will hinder his career development and how he should adjust his behavior accordingly. He perceives his personality as contradictory: on the one hand, he is meticulous and capable of noticing details others overlook; on the other hand, he is introverted and struggles to express himself.
IV. Interpersonal Relationship Status
    Xiao Wang’s interpersonal relationships at work are strained. His communication with colleagues is limited and largely confined to necessary work-related exchanges. He does not have close friends with whom he can share his distress, and his communication with family members is restricted to daily routines. During company-organized activities, he appears tense and restrained, finding it difficult to blend in with others.
V. Physical Condition
    Due to psychological stress associated with his personality traits, Xiao Wang has begun to experience physical symptoms. He frequently feels fatigued and reports poor sleep quality, often lying awake at night with racing thoughts. During periods of emotional agitation, he experiences headaches and palpitations. Although he has attempted to alleviate stress through exercise and listening to music, these efforts have had limited effect.
VI. Chief Complaint
    Xiao Wang’s primary concern is the psychological and physical distress caused by difficulties in exploring and understanding his personality traits. He feels extremely anxious, depressed, and helpless. He does not know how to adjust his work and life in accordance with his personality and longs to find a way to alleviate his symptoms, regain balance in life, and rediscover enthusiasm for work. However, he feels unsure where to begin. This persistent conflict and anxiety severely impair his quality of life, interpersonal relationships, and mental health.
VII. Additional Associated Problems
    1. Difficulties in Emotion Regulation
    When confronting issues related to personality trait exploration, Xiao Wang often struggles to regulate his emotions and becomes easily anxious, depressed, or irritable. These emotional difficulties increase his psychological burden and lead to guilt and self-blame afterward, forming a vicious cycle. For example, during interactions with colleagues, excessive nervousness causes him to behave awkwardly, which later leads to rumination and regret.
    2. Social Avoidance and Isolation
    As a result of his personality-related concerns, Xiao Wang’s social relationships with colleagues and friends have become increasingly strained. He often avoids interaction out of fear of discussing his personality issues, leading to growing social isolation. Even when colleagues or friends show goodwill, he finds it difficult to trust them or open up emotionally.
    3. Crisis of Self-Identity
    Under the shadow of ongoing personality-related struggles, Xiao Wang has begun to question his abilities and self-worth. He once believed himself to be competent, but in the workplace he now feels constrained by his personality traits. This has triggered a crisis of identity, leaving him uncertain about how to face his future life.
VIII. Course of Problem Development
    1. Initial Awareness of the Problem
    Xiao Wang first became aware of difficulties related to personality trait exploration during his first month at work. He noticed a mismatch between his personality and the workplace environment, accompanied by social difficulties and challenges in team integration.
    2. Key Event in Development I: Workplace Social Setback
    During a company-organized team-building activity, Xiao Wang was marginalized by colleagues due to his difficulty in social interaction. This experience left him feeling deeply disappointed and discouraged, prompting him to recognize the significant impact of personality traits on workplace socialization.
    3. Key Event in Development II: Confusion in Self-Understanding
    As his tenure at work continued, Xiao Wang became increasingly confused about his personality traits. He remained uncertain about whether his personality suited his current position and how he might adapt his work style accordingly.
    4. Recent Triggering Event
    One month ago, Xiao Wang was criticized by his supervisor during a performance evaluation for insufficient social competence. This incident caused intense shame and self-blame. He has since begun to consider seeking professional psychological counseling, hoping to find ways to alleviate his symptoms and regain a sense of direction in life.

Xiao Wang’s communication style is pleasing-oriented. In counseling, he may tend to minimize or conceal issues related to his personality traits in order to maintain a positive image. He may display a strong desire to please the therapist, avoid expressing disagreement or dissatisfaction, frequently seek approval or reassurance, and readily agree with the therapist’s suggestions—even when he does not fully understand or genuinely endorse them.
\end{Verbatim}
\caption{Profile ID: 56 (English version).}
\label{Fig-p56-en}
\end{figure*}
\end{CJK*}

\begin{CJK*}{UTF8}{gbsn}
\begin{figure*}[htp]
\centering
\begin{Verbatim}[fontsize=\tiny, frame=single, framerule=0.3pt, framesep=4mm, breaklines, xleftmargin=0.5em, xrightmargin=0.5em, formatcom=\rmfamily]
咨询师：你好你好～
来访者：（合理的请求）老师您好，我今天来是想请教一下，我现在该怎么走出失去母亲的痛苦呢？
咨询师：听到你这样说，我能感受到这份失去对你来说一定非常沉重。如果你愿意，你可以跟我说说，你最近一次特别想她，或者特别难受，是在什么时候发生的。
来访者：（提供信息）上个星期整理她遗物的时候，看到她写给我的信，我就完全控制不住自己，哭了很久很久。
咨询师：我能理解那一定是一个情绪突然被全部打开的瞬间。在这里，你不需要压住眼泪，也不需要马上恢复平静。我们可以按照你能承受的节奏，一点一点来。
来访者：（提供信息）谢谢您，我到现在都每天晚上睡不着，满脑子都是妈妈生前的事情。
咨询师：我能感觉到你现在的状态一定非常辛苦。每天到了夜里，本来是身体该休息的时候，却反而成了思念和回忆最汹涌的时刻。我也会好奇，当这些画面在脑子里出现时，它们更多是温柔的回忆，还是夹杂着痛苦、遗憾、甚至自责？不同的内容，其实会对你睡不着的方式产生不同的影响。
来访者：（提供信息）都有，回忆妈妈为我做饭、陪我聊天的场景时会觉得温暖，但一想到自己没能治好她的病就特别自责难过。
咨询师：你说的这两种感受，其实都非常真实，也常常会同时出现在丧亲之后。那种温暖的回忆，说明你和妈妈之间有过深厚而真实的连接；而随之而来的自责与难过，则往往来自于你对她的爱、责任感，以及你多么希望她能继续活着、少受一点痛苦。正因为你在乎，才会这样责怪自己。
来访者：（提供信息）是的，我太想让她继续活着了，虽然我一直陪在她身边照顾她，但还是觉得自己做得不够好。
咨询师：你这样说的时候，我能感受到一种很深的心痛，也能感受到你对母亲的那份用尽全力的爱。太想让她继续活着，这句话本身就说明，你的内心是多么不愿意放手、多么希望能替她多承担一些痛苦。可正因为你那么在乎，才会在她离开后，一遍又一遍地回头检视自己，觉得怎么做都不够。
来访者：（提供信息）听您这么说，我的眼泪又忍不住掉下来了，好像您真的能理解我的心情。
咨询师：有时候，人真正哭出来，并不是因为更痛了，而是因为终于不用一个人扛着了。你刚才说“好像您真的能理解我的心情”，这本身就很重要，也很珍贵。
来访者：（认可，扩展）您说得对，我一直在努力控制自己，不想在家人面前显得软弱。其实我最近总是在想，妈妈在的时候总是鼓励我要坚强，但同时也告诉我难过的时候可以哭出来，可我现在甚至不敢和家人一起回忆她。
咨询师：在家人面前不敢显得软弱、不敢一起回忆她，我会理解为：你不仅在承受失去母亲的痛，也在承担着一种“我要撑住”的角色。好像只要你一哭、一提起她，就可能让整个家庭再次陷入悲伤，而你在无形中把自己放在了“不能倒下”的位置上。这其实是一种非常辛苦的责任。
来访者：（提供信息）是的，我觉得自己需要照顾丈夫和儿子的情绪，但我自己也真的很需要被安慰和理解。
咨询师：你这样说的时候，其实把你现在最辛苦的地方说得非常清楚了：你一边在照顾丈夫和儿子的情绪，一边却把自己真正的需要放在了后面。这并不是因为你不需要安慰，而恰恰相反，是因为你太需要了，却觉得自己不能去要。
来访者：（提供信息）或许是因为母亲教我要照顾好家人，我总觉得自己必须要坚强，不想让家人担心我。
咨询师：你这样理解自己，其实让我看到一条很清晰、也很心疼的线索：你把母亲的教导内化成了一种对自己的要求。照顾好家人、不要让他们担心、自己要坚强，这些原本是母亲对你的期望、也是她对家庭的爱，但在她离开之后，这些话好像变成了一条你必须遵守的“内在规则”，时时刻刻提醒你不能倒下。
来访者：（提供信息）您说得对，我总觉得如果自己也崩溃了，整个家就真的散了，可是我现在真的好累。
咨询师：你现在说的这句话，非常沉重，也非常真实。“如果我也崩溃了，整个家就真的散了”——这像是一根一直绷在你心里的弦，把你牢牢固定在一个不能倒下的位置上。而与此同时，你又清楚地感受到：你已经撑得太久、太用力了，真的很累。
来访者：（提供信息）我知道您说的很对，但每次放松一点，那些回忆和痛苦就会一下子涌上来，让我怕得不敢卸下这个担子。
咨询师：从这个角度看，你现在的“不敢卸下担子”，并不是问题本身，而是一种在巨大丧失之后非常常见、也很聪明的自我保护方式。
来访者：（提供信息）其实我最害怕的是，如果我把这些负担放下了，就再也感受不到妈妈留给我的爱和责任了。
咨询师：你说到的这个害怕，其实非常深，也非常重要。你担心的并不是“轻松”本身，而是一旦把负担放下，和妈妈之间那条紧密的连接也会一起断掉。在你的体验里，责任、坚强、承受，好像已经变成了你继续爱她、记得她、对她忠诚的一种方式。如果不再背着这些，你会担心她在你心里的位置也会变淡。
来访者：（提供信息）是啊，我不敢放松是因为觉得痛苦和想念才能证明我有多爱妈妈。
咨询师：你把这句话说出来，本身就是一次非常深的觉察。你在用自己的方式告诉我：痛苦和想念，对你来说不仅是情绪，更像是一种证明——证明你没有忘记她，证明你对她的爱依然在。在失去如此重要的人之后，有这样的感受，并不罕见，也一点都不奇怪。
来访者：（提供信息）听您这样说，我觉得内心好像开始有一点点明白，不一定非要用痛苦来证明我对妈妈的爱。
咨询师：我很想慢慢地陪你停在这里，而不是急着往前走。因为你现在触碰到的，是一个很核心的转变：
你开始隐约感受到，爱也许可以有别的存在方式，不只是通过持续的痛苦、自责和承受。
来访者：（提供信息）确实，我也希望慢慢学着用更温柔的方式来纪念和爱妈妈。
咨询师：听到你这样说，我会觉得这是一种很成熟、也很温柔的愿望。你并不是想忘记妈妈、放下她，而是希望找到一种不再那么折磨自己、却依然真实地纪念和爱她的方式
来访者：（其他，提供信息）老师您的理解让我觉得心里好受了一些。这几天我一直在想要好好收拾一下妈妈的照片，想找个漂亮的相册把它们都整理出来。
咨询师：我注意到一个很重要的变化：以前，你提到和妈妈有关的事情时，更多伴随着失控、自责和害怕；而现在，你在谈的是想要整理、选择、慢慢来。这说明你已经开始在心里为她腾出一个空间，而不是被回忆一下子淹没。
来访者：（认可）是的，我现在觉得整理照片是我可以做的一件事情，也许这样能让我以更平静的心情来怀念妈妈。
咨询师：也许你可以把这件事当作一种温柔的对话，而不是任务。
你不需要一次性整理完，也不需要每一张照片都处理得很好。
有些照片，也许你会看得久一点；
有些，也许现在还不太敢碰；
这些选择本身，都是你在尊重自己的承受力。
来访者：（提供信息）是啊，我也觉得不用着急，我想慢慢地一张一张看，让每一张照片都带我回到和妈妈在一起的美好时光。
咨询师：我特别想肯定你提到的“不着急”。这和你一开始必须撑住、不能停下来的状态，已经很不一样了。
来访者：（提供信息）我可以试着带着微笑看照片，让自己想起和妈妈在一起的快乐，而不是每次都哭着看。
咨询师：你正在学习一件非常不容易的事，在爱里，既不失去她，也不失去自己。
来访者：（提供信息，认可，提供信息）是的，妈妈一定也希望看到我能保持快乐和坚强。我还记得她常说，生活要往前看，不能一直沉浸在悲伤里。其实我想试着把这些照片分享给家人，让我们一起回忆妈妈美好的样子。
咨询师：你现在说的这一段，让我感受到一种从独自承受，慢慢走向连接的力量。你开始不仅是在心里和妈妈对话，也在想把这份怀念带回到家庭里，让它不再只是你一个人的重量。
来访者：（提供信息）是的，我想试着让爱和回忆变成全家人共同的温暖，而不是自己一个人默默承受。
咨询师：我觉得这是一个很重要的转折点，你不再只是想着怎样把痛苦一个人扛住，而是开始把爱和回忆，慢慢地从“只能独自承受的重量”，转化为可以被家人一起抱着的温暖。
来访者：（提供信息）今天和您聊完，我感觉心里的压力轻了不少，真的很感谢您的理解和开导。
咨询师：听你这样说，我真的很为你感到高兴。压力轻了一些，哪怕只是一点点，都是很重要的变化。
来访者：（提供信息）我这周就找时间把照片整理一下，可能也需要您继续给我一些指导，我希望能慢慢学会更好地和这份思念相处。
咨询师：当然没问题。你不必独自摸索。我会在这里，继续陪你一起学习，如何在爱里怀念，也在怀念中生活。
来访者：（提供信息）好的，那我先回家慢慢整理照片，下周再来和您分享我的感受和进展。
咨询师：你可以安心地回家，按照你自己的节奏来。我们下周再继续。
来访者：（提供信息）嗯，我挺期待下周能再来和您聊聊，那我就先回去了。
咨询师：好的，我们下周再见
来访者：（提供信息）谢谢老师今天帮我开导，让我觉得心里舒服了很多，再见。
咨询师：再见
来访者：（提供信息）老师再见，我会照您说的慢慢来的。
咨询师：期待下次你的分享
来访者：（提供信息）我会按自己能接受的速度整理照片，到时再和您聊聊进展。
咨询师：祝一切顺利～
来访者：（提供信息）谢谢老师的祝福，我相信有您的帮助，我一定能走出这个低谷的。
\end{Verbatim}
\caption{A dialogue between a human counselor and an LLM-based client (Profile ID: 30 and Trajectory ID: 12). For the English version, see Figures \ref{Fig-p30-t12-en-part-1} and \ref{Fig-p30-t12-en-part-2}.}
\label{Fig-p30-t12-zh}
\end{figure*}
\end{CJK*}

\begin{CJK*}{UTF8}{gbsn}
\begin{figure*}[htp]
\centering
\begin{Verbatim}[fontsize=\tiny, frame=single, framerule=0.3pt, framesep=2mm, breaklines, xleftmargin=0.2em, xrightmargin=0.2em, formatcom=\rmfamily]
咨询师：你好
来访者：（提供信息）医生您好，我叫李华，我最近一直过得很不好，我妈妈去世后我总是睡不着觉，心里特别难受。
咨询师：很感谢你愿意把这些感受告诉我。失去母亲是一件非常沉重的事情，在这样的经历之后，睡不着、心里难受，其实都是很常见、也很可以理解的反应。听起来，这段时间你一直在独自承受很多痛苦。
来访者：（提供信息，扩展）是的，自从妈妈离开后我整个人都垮了，工作也没法专注，晚上总是辗转难眠。每次想起妈妈生病时的样子，我就特别自责，觉得当时如果能早点发现她的病情就好了，虽然丈夫说这不是我的错，但我就是忍不住这样想。
咨询师：我能感受到你现在背负着的那份沉重。妈妈的离开不仅带走了你很重要的一部分，也让你反复回到那些痛苦的画面里，尤其是她生病时的样子。听你这样说，我会感觉你一方面在承受失去母亲的巨大悲痛，另一方面又在不断地把责任压在自己身上，好像只要当时做得再好一点，这一切就可以被避免。
来访者：（提供信息）是的医生，我现在每天晚上都躺在床上回想妈妈生病时的点点滴滴，总觉得如果我能再细心一点、再照顾得好一点，也许结果就会不一样。
咨询师：你提到“如果我能再细心一点、再照顾得好一点”，这句话里有很多你对妈妈的爱，也有很多你对自己的苛责。
来访者：（提供信息）其实我知道丈夫说得对，妈妈的病不是我的错，但每次想到她临走前那么痛苦的样子，我就控制不住地自责。
咨询师：你能这样清楚地说出“我知道理性上不是我的错，但情感上控制不住自责”，这本身就说明你对自己的状态是有觉察的。很多时候，人真正痛苦的地方，正是在明白道理却无法停止内心的折磨。所以这并不是你不够理性，而是这段丧失带来的情绪还没有被真正安放。
来访者：（提供信息，扩展）我最近整理妈妈的遗物时，发现了她写给我的信，看到她字里行间对我的关心和期望，眼泪就止不住地流。想到以前妈妈每次给我做饭、陪我聊天的场景，就觉得生活中突然少了一个最重要的人，这种感觉让我特别无助和孤独。
咨询师：听你这样描述，我能感受到那一刻对你来说有多么汹涌。那种一边被爱包围、一边又被失去击中的感觉，真的会让人一下子承受不住，眼泪停不下来是很自然的反应。
来访者：（自我批评或无望）不知道我还能不能振作起来，感觉自己快要被这种悲伤压垮了。
咨询师：我想先明确、也很重要地告诉你一件事：现在的你不需要振作，也不需要证明自己能不能走出来。在丧母这样重大的失去之后，感到被悲伤吞没，并不是脆弱，而是人之常情。
来访者：（提供信息）谢谢医生你这样说，我感觉现在终于有人理解我了，我好像不用那么着急地逼自己走出来。
咨询师：我很高兴你能这样感受到，被理解、被接住本身就是一件很重要的事情。你不需要急，也不需要证明什么。你可以在这里慢慢来，我会陪着你一起走这一段。
来访者：（提供信息）其实我已经好久没有这么放松过了，每次和家人朋友聊到妈妈，他们都急着安慰我让我开心点，反而让我更难受。
咨询师：你能这样说出来，其实很重要。你的家人朋友可能是在用他们能想到的方式爱你，但你真正需要的，可能不是安慰的话，而是有人愿意坐在你的悲伤里，不急着把你拉走。
来访者：（防卫个人观点）我觉得我已经很努力地照顾了妈妈，我每天给她煮三餐、打点滴、陪她聊天，根本不需要别人来说我做得不够。
咨询师：你现在开始用“我已经很努力了”来描述自己，而不再只是“如果我当初再怎样就好了”。这并不是否认悲伤，而是你开始为自己的付出争取一个真实的位置。这一步，对走出无尽自责来说，是非常关键的。
来访者：（防卫个人观点）虽然我知道您想帮我缓解内疚，但我还是觉得自己更了解这个情况，如果我当时再细心一点，一定能更好地照顾妈妈。
咨询师：我也想很坦诚地告诉你一件事：我并不急着要你放下内疚，或者马上接受“你已经做得够好了”这个说法。因为对你来说，这份内疚并不是一个简单的认知错误，而是和爱、责任、失去紧紧缠在一起的东西。如果我们太快把它拿走，反而可能让你觉得，好像有人在否定你对妈妈的在乎。
来访者：（防卫个人观点）医生，我知道您是好意，但我觉得只有实际照顾过病人的人才明白我当时的无助和自责。
咨询师：我能理解你为什么会这样说，而且我很尊重这一点。我相信你说的这些，是因为你真的经历过。而我在这里的角色，不是评判你当时够不够细心，而是陪你把那段几乎没人真正陪你走完的路，慢慢说出来。
来访者：（扩展）医生您说得对，当时我一个人扛下了所有的事情，可能是因为没有人和我一起分担，所以那些痛苦的记忆就一直压在我心里放不下。
咨询师：你能这样把这一点说出来，其实是一个很重要的触碰。当你说“当时我一个人扛下了所有的事情”，我会感觉到，那段时间你不仅是在照顾妈妈，也是在独自承受恐惧、责任、决策和失去的预感。而这些东西，如果当时没有被分担、没有被看见，往往就会在后来以记忆、自责、反复回想的形式留在心里。
来访者：（提供信息）是的，我那时候真的很希望有人能理解我的无助，现在回想起那段独自面对妈妈病情的日子，心里还是很难过。
咨询师：你这样说的时候，我会感觉到一种很深的、迟来的悲伤。那不仅是对妈妈的失去，也是对当时那个孤零零承受一切的自己的心疼。
来访者：（扩展）在那段时间，每天下班我都要赶到医院照顾妈妈，白天工作已经很累了，但看到妈妈躺在病床上，我就觉得自己一点也不能休息，总想再多陪她一会儿，现在想想，那时的我真的很累很辛苦。
咨询师：当你这样回忆那段日子时，我能清楚地感受到你语气里的变化。你不再只是站在“如果我当时再怎么样就好了”的位置上，而是开始看见那个已经被消耗到极限、却仍然不允许自己停下来的自己。
来访者：（认可）您说得对，那时候我确实给自己太多压力了。
咨询师：这句话本身就很有力量。它不是轻描淡写，而是来自你对那段经历更真实、更完整的看见。你终于把目光从“我是不是做得不够好”，慢慢移向了“那时的我承受了多少”。
来访者：（提供信息）医生，我现在回想起来，那个时候确实身心俱疲，可能正是因为太累了，才会一直觉得自己做得不够。
咨询师：这句话很有力量。它不是轻描淡写，而是来自你对那段经历更真实、更完整的看见。你终于把目光从“我是不是做得不够好”，慢慢移向了“那时的我承受了多少”。
来访者：（提供信息）不过我想说的是，丈夫总是劝我振作起来，但我实在没办法短时间内就恢复正常，这让我觉得很愧疚。
咨询师：我能感受到一种夹在中间的为难：一边是你还在经历真实而深刻的悲伤，另一边是丈夫的期待和关心，而你却发现自己怎么都跟不上那个“该振作起来了”的节奏。那种做不到、又觉得对不起别人的感觉，本身就很消耗人。
来访者：（提供信息，扩展）医生，我知道丈夫是出于好意，但每次他催我振作的时候我都特别难过。有时候我真希望他能安安静静地陪在我身边，给我一个拥抱就好，而不是总说那些让我更加压抑的话。
咨询师：我能感受到那份被误解的孤独。你并不是不明白丈夫的好意，而是他的方式，恰恰在你最脆弱的时候，让你觉得更难受、更压抑。你真正渴望的，并不是被推动着往前走，而是被安静地陪着待在原地。
来访者：（提供信息）我多希望丈夫能明白，我现在最需要的不是那些劝慰的话，而是他默默的陪伴和支持。
咨询师：你其实已经把你此刻最真实、最核心的需要说得非常清楚了。你不是不想被关心，也不是拒绝帮助，而是你真正需要的是一种不要求你改变状态的陪伴。那种陪伴里，没有劝你快点好起来，也没有暗示你现在这样不对，只是安静地在你身边，让你知道：你不用一个人扛。
来访者：（提供信息）也许我该找机会和丈夫聊聊这些，让他知道我现在最需要的是什么。
咨询师：这是很重要的一步，这并不是因为你“应该去沟通”，而是因为你开始认真对待自己的需要，也相信它值得被听见。
来访者：（提供信息）我总觉得要说这些很难开口，害怕丈夫会觉得我太矫情或者太脆弱了。
咨询师：你这样担心，其实非常能理解。对很多人来说，真正难的并不是知道自己需要什么，而是害怕把这些需要说出口之后，会不会被误解、被否定，甚至被看轻。你担心丈夫觉得你矫情、脆弱，这背后往往不是自信不足，而是你很在意这段关系，也很在意自己在他眼中的位置。
来访者：（提供信息）嗯，您说得对，我就是害怕把心里的想法说出来会影响我和丈夫的关系。
咨询师：我能感觉到你内心其实有一个很清晰、也很温柔的顾虑：你并不是不想靠近丈夫，而是太在乎这段关系了，才会担心一旦把真实的想法说出来，会不会反而把距离拉开。这种害怕，本身就说明你很珍惜他，也很珍惜你们之间的连接。
来访者：（认可）确实医生，您说到我的心坎里了，我很珍惜我和丈夫的关系。
咨询师：我会感觉到，你其实一直在为这段关系付出很多。哪怕在自己最难受、最需要被抱一抱的时候，你还是先替丈夫考虑：他会怎么想？会不会觉得我太脆弱？关系会不会因此受影响？这种先顾及对方的习惯，说明你在关系里一直是一个很有责任感、也很体贴的人。只是，在经历了母亲的离世之后，这种一贯的“先顾别人”，让你更累了。
来访者：（提供信息）其实我真的很累了，既要调整自己的情绪，还要顾及丈夫和儿子的感受，有时候真想放声大哭一场。
咨询师：你这样说的时候，我能很清楚地感受到那种被耗到几乎没有余力的疲惫。你不仅在承受失去母亲的悲痛，还在不停地调节自己，照顾丈夫的反应、顾及儿子的感受，好像无论多难受，你都得先稳住、撑住。长期这样压着自己，任何人都会累到极限。
来访者：（提供信息）我觉得再这样下去我真的要撑不住了，每天都在强忍着眼泪过日子。
咨询师：我会非常认真地停下来陪你。
每天强忍着眼泪过日子、觉得自己快撑不住了，这已经不是一般的疲惫，而是你在长期压抑、长期独自承受之后，身心一起发出的警告。你不是矫情，也不是承受能力不够，而是真的已经到达一个很极限的位置了。
来访者：（防卫个人观点）医生，没有人比我更了解那段痛苦的日子，所以我觉得我应该继续坚持下去，就像当初照顾妈妈那样。
咨询师：当初的坚持，是在一个“必须有人撑着”的紧急状态里。而现在，你已经走出了那个阶段，可你的身体和情绪却还停留在“不能倒下”的模式里。长期处在这种状态，并不会让痛苦慢慢消失，反而会让你一个人承受得越来越重。
来访者：（提供信息，扩展）我知道自己应该好好休息，但每次一躺下来就会想起妈妈，然后眼泪就控制不住。每次您这么和我说话的时候，我感觉终于有人真正理解我内心的挣扎和痛苦了。
咨询师：我很珍惜你刚才说的这句话，也很认真地接住它。也许我们可以慢慢换一个方式来看“休息”这件事。
休息不一定意味着马上能睡着、不想妈妈、不流泪。对你现在来说，休息也许只是允许自己躺着哭一会儿，而不是马上爬起来继续忍。
来访者：（其他，提供信息）这里真是个让人舒服的地方，以前我都是一个人躲着哭。医生，您说得对，我确实应该给自己一个喘息的机会。
咨询师：听你这样说，我心里是很为你松一口气的。也许我们可以把这个“喘息”想得很小、很具体，而不是一个遥远的目标。
\end{Verbatim}
\caption{A dialogue between a human counselor and an LLM-based client (Profile ID: 30 and Trajectory ID: 139). For English version, see Figures \ref{Fig-p30-t139-en-part-1} and \ref{Fig-p30-t139-en-part-2}.}
\label{Fig-p30-t139-zh}
\end{figure*}
\end{CJK*}

\begin{CJK*}{UTF8}{gbsn}
\begin{figure*}[htp]
\centering
\begin{Verbatim}[fontsize=\tiny, frame=single, framerule=0.3pt, framesep=4mm, breaklines, xleftmargin=0.2em, xrightmargin=0.2em, formatcom=\rmfamily]
咨询师：你好！欢迎来到咨询中心
来访者：（提供信息）您好，我是章依依，我最近压力很大，希望能得到一些帮助。
咨询师：你好，依依，很高兴你愿意在这里开口。你愿意从哪里开始说起呢？
来访者：（提供信息）我来美国读书一个学期了，可是总是觉得很不适应，每天都觉得很孤独。
咨询师：一个人来到一个完全不同的国家、语言和文化环境里，才一个学期，就要同时面对学习、生活、身份变化和情感支持的缺失——在这样的处境中感到不适应、孤独，其实是一种非常可以理解、也很真实的反应，而不是你的问题。
来访者：（提供信息）可是我看周围其他的中国同学都能很快适应，他们参加课外活动、跟美国同学交朋友，我就是做不到...
咨询师：我感到你说的这句话里，其实藏着很多委屈。当你看到“别人都可以，只有我不行”的时候，那种感觉往往不只是孤独，还会夹杂着自责、羞愧，甚至怀疑自己是不是哪里有问题。你现在的状态，并不说明你能力差、性格有问题，或者不够努力。你看到的“他们适应得很好”，其实只是外显的一部分，而适应这件事，本身就存在巨大的个体差异。
来访者：（提供信息）是的，您说得对，但我觉得自己以前在国内一直都是别人眼中的优秀学生，现在却变得这么不堪。
咨询师：我相信这对很多原本在国内一路被肯定、被赞赏的学生来说，是非常沉重的打击。我想你现在的状态并不是不堪，而是一个原本依靠“优秀”站稳世界的人，突然被放进了一个完全陌生的坐标系里。原有的评价体系、反馈速度、社会位置，一下子都失效了，这本身就会让人产生强烈的自我坍塌感。
来访者：（提供信息）对，我总觉得自己一无是处了，连课堂上的简单问题都不敢回答。
咨询师：我听到你现在心里有一个非常严厉、也非常孤单的声音在反复告诉你：我一无是处，我不行，我不能再犯错了。
来访者：（提供信息）您说得对，我每次一想说话，就怕说错，就更不敢开口了。
咨询师：你能把这一点说得这么清楚，说明你对自己的内在体验其实非常敏感、也很有觉察力。我们可以一起慢慢练习一种新的方式：在不完美中，依然允许自己存在。
来访者：（防卫个人观点）我在国内的时候成绩一直都是顶尖的，现在我觉得学习成绩下降完全是因为语言问题，不是我本身的能力问题。
咨询师：我觉得你这个判断非常重要，你现在最需要的不是能力证明，而是对自己的允许和缓冲。
来访者：（提供信息，认可，提供信息）但是每次想到这种状态还要持续几年，我就感到很绝望。其实您说得对，我也明白需要给自己一些时间适应。可是宿舍那边的美国同学每次找我聊天，我都不知道该说什么，总是找借口躲着她们。
咨询师：我想你现在躲开的，不是美国同学，而是一种持续被暴露在“我可能不够好”的恐惧里。对你来说，真正重要的不是“聊得好不好”，而是让你的身体慢慢学会：这种场景不一定是危险的。
来访者：（防卫个人观点）虽然室友们很好，但我觉得自己有自己的生活方式，没必要强迫自己融入他们。
咨询师：我想先肯定你的一部分判断：你确实没有义务为了“融入”而改变自己的生活方式。不是所有人都需要通过频繁社交、和室友变得很熟，才能在异国生活得好。对一些人来说，独处、本来就有自己的节奏、把能量更多放在学习或内在世界里，本身就是一种适合自己的方式。但我也很想确定一下，这是你的个人选择，还是你的反抗
来访者：（重构（重构观点或行为改变））也许这确实是我在害怕失败才选择逃避，我应该试着慢慢接触一下室友。
咨询师：你愿意承认“我可能是在逃避”，并不等于否定自己，反而说明你在重新把主动权拿回来。我很珍惜你刚刚说的这句话。
来访者：（提供信息）您觉得我可以从哪些小事开始尝试与室友互动呢？
咨询师：这是一个非常好的问题，在这个阶段，关键不是“聊得多好”，而是让你的身体和情绪积累一些“我可以应付”的经验。
来访者：（认可）也许我可以先从和室友一起吃个早餐或是分享一些食物开始。
咨询师：我觉得你选的这个起点非常好，也非常贴合你。我们可以把这个尝试拆得更具体一点，让它既可控、又安全。
来访者：（提供信息）我想试试先从跟室友分享一些国内带来的零食开始，但还是有点紧张不知道她们会不会喜欢。
咨询师：你现在紧张，并不是因为这件事有多大，而是因为你很在意、不想再次受挫。你能注意到这一点，说明你对自己的感受是坦诚的。
来访者：（防卫个人观点）虽然我有点紧张，但是我比别人懂中国零食，室友肯定会认可我的品味。
咨询师：我感到你的自信似乎又回来了一些。无论他们最后是不是喜欢，重要的其实是你的自我感受和价值感是不是能够被重新建立起来。
来访者：（扩展）是啊，我一直在国内那么优秀，在这方面还是有判断力的，虽然语言不太好，但我也有自己的长处。
咨询师：这是一种更加全面的对自我的理解和肯定，我很开心你能够意识到一个更加完整的自己。
来访者：（提供信息）嗯，听您这么一说，我也开始觉得自己不是一无是处了，只是需要一点时间慢慢适应。
咨询师：我想确实如此，这是一个非常重要、也非常真实的转变。你能给自己留下转变的时间，这非常珍贵。
来访者：（提供信息）我现在感觉好多了，谢谢您，希望下次还能跟您多聊聊其他的问题。
咨询师：当然，任何问题我们都可以一起面对。我真的很为你感到高兴。不是因为问题都解决了，而是因为你现在的内在状态，比一开始多了一点空间、多了一点温和。这对你来说，是非常珍贵的变化。
来访者：（认可）是的，我确实感觉到内心没有那么沉重了，更有勇气去面对接下来的挑战。
咨询师：我为你感到骄傲也欣慰。你现在的状态，已经比你想象中更有力量了。慢慢来，你正在走在一条对自己更温和、也更真实的路上。
来访者：（合理的请求）对了，我想问问如果室友不喜欢我的零食，我该怎么应对呢？
咨询师：这是一个非常成熟、也非常现实的问题。你能提前想到“如果她们不喜欢，我该怎么办”，说明你已经不再把这件事当成非黑即白的成败测试，而是在为各种可能性预留心理空间。你自己现在是怎么想的呢
来访者：（认可）我明白了，即使室友不喜欢也没关系，重要的是我愿意跨出这一步。
咨询师：你不需要把每一次互动都变成“被认可的证据”。
你只是在练习：在不确定中，依然站在自己这一边。
来访者：（重构（重构观点或行为改变））我想现在最大的突破就是能够接受自己也会不完美，这也许就是我在美国需要学习的第一课。
咨询师：原来我可以不完美，而依然是一个有价值、值得尊重的人。这是非常深刻而且重要的课程呢。
来访者：（其他）天气今天有点冷，我喝着咖啡感觉暖和多了。
咨询师：你平时喜欢喝咖啡吗
来访者：（其他）还是更喜欢喝奶茶，美国的奶茶店总是不如国内的好喝。
咨询师：哈哈哈，我听出来你对于家乡的想念了。
来访者：（提供信息，其他）是的，尤其是周末午后，我特别想念国内和朋友一起喝奶茶的时光。其实这里也有很多好玩的，比如上周末我试了一家美国朋友推荐的冰淇淋店。
咨询师：当想念出现的时候，你最想念的究竟是“人”“氛围”，还是那种“被懂得的感觉”
来访者：（其他）您看，昨天我在Costco买了一大盒曲奇饼干。
咨询师：味道如何？
来访者：（重构（重构观点或行为改变））味道还不错，让我想起我和室友可以一起品尝这些美食。
咨询师：我能感觉到一种很温和、很自然的转向正在发生。不是从“我应该去社交”，而是从这个味道不错，我突然想到可以和室友一起尝尝。
来访者：（提供信息）对，想到有好吃的可以跟别人分享，心里突然就没那么压抑了。
咨询师：你的注意力不再只盯着“我做得够不够好、会不会失败”，而是短暂地转向了关系中的流动感。这是一件很重要的事情
来访者：（提供信息）我现在就想给室友分一些曲奇，反正我觉得美食是最容易拉近关系的。
咨询师：我能感受到你此刻那种自然想要行动的轻盈感。不是“我应该去做点什么”，而是现在这样做，好像是顺着我的感觉走的。这一点非常重要。
来访者：（提供信息）我觉得如果从分享食物开始，我还蛮期待和室友聊天的感觉。
咨询师：等你的好消息，无论如何，都要记住关注自我的感受，接纳一切可能发生的结果。
来访者：（提供信息）嗯，希望下次我们还能像今天这样轻松地聊天。
咨询师：期待与你的下次见面。
来访者：（提供信息）那我先去找室友分享这些好吃的，有点小紧张，但是我想试试看。
咨询师：这说明你很重视你们的关系
来访者：（提供信息）是的，我希望能和室友慢慢建立起更友好的关系，就算过程有点艰难。
\end{Verbatim}
\caption{A dialogue between a human counselor and an LLM-based client (Profile ID: 45 and Trajectory ID: 105). For the English version, see Figures \ref{Fig-p45-t105-en-part-1} and \ref{Fig-p45-t105-en-part-2}.}
\label{Fig-p45-t105-zh}
\end{figure*}
\end{CJK*}

\begin{CJK*}{UTF8}{gbsn}
\begin{figure*}[htp]
\centering
\begin{Verbatim}[fontsize=\scriptsize, frame=single, framerule=0.3pt, framesep=2mm, breaklines, xleftmargin=0.2em, xrightmargin=0.2em, formatcom=\rmfamily]
咨询师：你好，欢迎来到咨询中心
来访者：（提供信息）您好医生，我是小王，我最近工作中遇到一些困扰，想找您聊一聊。
咨询师：小王你好，我在听你说呢
来访者：（提供信息）我觉得自己在工作中特别不善于和同事交流，不知道该怎么和大家沟通。
咨询师：听起来，你在工作中与同事相处时感到有些为难，尤其是在如何开始和维持沟通这件事上
来访者：（提供信息）对，就像上周公司聚餐，我一个人坐在角落，想跟同事聊天又不知从何说起，特别难受。
咨询师：我注意到你用了“特别难受”这个词，似乎不仅仅是尴尬或紧张，可能还夹杂着被隔在外面的感觉。
来访者：（提供信息）嗯，是的，看着大家有说有笑，就感觉自己像个外人一样，很想融入但又不知道该怎么做。
咨询师：我注意到，你并不是对同事没有兴趣，也不是不想融入，恰恰相反，你是很渴望连接的。你愿意再说说，当你想融入却停住时，身体或情绪上最明显的反应是什么吗？
来访者：（提供信息）一到那种场合，我就感觉心跳加快，手心出汗，脑子一片空白。
咨询师：你的身体在那一刻进入了一种高度警觉、准备应对威胁的状态。当身体先一步拉响警报时，大脑自然就很难继续思考该说什么。
来访者：（提供信息）医生说得对，就像您说的，我每次想主动搭话的时候，就像有什么东西卡住了一样，整个人僵在那里。
咨询师：问题不在于你想不想，而在于你太在意了。
来访者：（防卫个人观点）其实也没那么夸张，我觉得自己只是比较害羞而已，性格内向也没什么不好的啊。
咨询师：我能感觉到一种很重要的自我保护和自我接纳：内向、害羞本身并不是问题，它们确实是性格的一部分，也有其价值。你能为这一点发声，其实说明你对自己并不是全然否定的。
来访者：（提供信息，认可，提供信息）是的，我也觉得自己还是有一些优点的，比如做事情很细心。我明白医生说得对，可能是我太在意别人怎么看我了。每当我想和同事聊天的时候，总是担心自己说错话，或者担心别人会觉得我很无趣。
咨询师：你这是一种很成熟的自我理解。我还想问，当你担心别人会不会觉得你无趣时，这个“别人”在你心里是一个很模糊的群体，还是某一类特定的人？有时，焦虑并不是来自所有人，而是来自我们心中那个特别重要、特别容易被内化的目光。
来访者：（防卫个人观点）我觉得这种担心很正常啊，毕竟现在的职场就是这样，不社交的人容易被边缘化。
咨询师：你的担心并不夸张，也不是软弱，而是一种对环境的适应尝试。但在心理层面，你现在承受的，可能是一个被放大的内部版本。
咨询里我们真正要做的，可能不是否认职场的现实，也不是逼你变成外向的人，而是一起寻找一个问题：有没有一种方式，既不需要你勉强自己表演，又能让你在关系中逐渐“存在”起来？
来访者：（重构（重构观点或行为改变））我明白了，也许我可以先从小范围的交流开始，比如和经常合作的同事聊聊工作上的事情。
咨询师：如果你愿意，下一步我们可以一起更具体地想一想：在最近的工作中，有没有一个很小、很自然的时刻，你可以尝试迈出这一步。
来访者：（提供信息）嗯...可以试试从跟隔壁工位的小李开始，她经常借我的文具，也许我可以在她来借东西的时候多聊几句。
咨询师：这个想法非常具体，也很贴合你的实际情况，我觉得你已经在用一种既尊重自己性格、又不回避现实需要的方式往前走了。
来访者：（认可）确实是这样的，医生您说得让我感觉豁然开朗了很多。
咨询师：我很高兴你这样说，也很珍惜你此刻的这个感受。你真的在某种层面上加深了对自己的理解。
来访者：（提供信息）医生您说的没错，我觉得现在对自己的性格特点有了更清晰的认识，不那么着急改变了。
咨询师：我很高兴你能意识到自己的转变。还有什么是我可以帮助你的吗？例如我们聊聊具体的沟通方式？
来访者：（防卫个人观点）我觉得我做事细心、认真，不需要改变什么，可能就是工作环境对我的要求太多了。
咨询师：确实，环境是一个很重要的变量，很多时候并不是你做错了什么，或者是你的性格不好。
来访者：（扩展）是啊，每个人都有自己的性格特点，只是我一直都在努力调整自己去适应环境，反而忽视了自己内心真实的想法。
咨询师：从现在开始也完全不晚，而且我发现你有很强的自我反思的能力，这很重要也很难得。
来访者：（提供信息）医生，您说得对，我确实比较会思考，但有时候也是过度思考了才会这么焦虑。
咨询师：你这个觉察其实非常关键，而且说得很准确。你“比较会思考”本身是一种能力，它让你做事周全、考虑后果、对环境敏感。但当这种能力在高度在意他人评价的情境中被持续启动时，就容易从帮助你理解世界，变成反复放大压力的工具，于是焦虑就出现了。
来访者：（提供信息）就是因为想太多，才会在同事面前不知所措，下次我要试着少想一些，看看会不会好一点。
咨询师：我能感受到一种更温和、也更现实的态度：不是要求自己立刻改变，而是愿意做一个小小的尝试，看看会发生什么。这本身就已经是在照顾自己了。
来访者：（认可）对，医生说得很有道理，我应该先从小的尝试开始，慢慢来就好。
咨询师：是的是的你现在这个理解本身就非常重要，也很符合你真正的节奏。你已经不再要求自己一下子变成“很会社交的人”，而是允许自己从小、可控、不会压垮自己的尝试开始。
来访者：（合理的请求）医生，您能帮我想想，我该用什么具体的方式开始跟同事交谈吗？
咨询师：当然可以，而且你这个问题问得非常恰当。你平时是如何和同事沟通的呢？
来访者：（认可）我一般就是工作需要的时候才跟同事说话，大多数时候都是在等着别人主动来找我。
咨询师：你这样描述，其实让我看到的是一种很典型、也很能理解的应对方式：在你觉得安全、明确、有理由的时候再开口，而把其他时候留给观察和等待。同时，也正如你已经觉察到的，这种“只在工作需要时说话、更多是等别人来找你”的模式，确实可能让你在关系上显得有点被动，也更容易强化那个感觉——好像自己总是在边缘、只能被动参与。
来访者：（重构（重构观点或行为改变））其实我可以先在工作讨论的基础上慢慢扩展话题，这样感觉更自然一些。
咨询师：确实如此，你可以考虑把“等别人来找我”，变成“我给一个很小的信号”。
来访者：（其他）今天天气真不错，阳光明媚的。
咨询师：关于天气？
来访者：（其他）医生，我看您办公室外面的花草养得挺好的。
咨询师：谢谢你的夸奖，我们继续我们的咨询如何
来访者：（提供信息，其他）对，医生，我们再说说我最近感觉特别疲惫的问题，每天晚上都失眠。真希望和医生您这样聊天的机会能多一些，我感觉和您说话很轻松。
咨询师：谢谢你的肯定，你的睡眠，你可以再说说吗
来访者：（其他）医生，这个办公室真挺宽敞的，设计得很温馨。
咨询师：你的分心似乎有点严重。
来访者：（重构（重构观点或行为改变））医生您说的对，我总是通过找别的话题来逃避自己的问题，其实我就是想回避那些让我焦虑的事情。
咨询师：你刚才是想要回避哪些让人焦虑的事情呢
来访者：（提供信息）想到工作中的社交压力和失眠的问题，我就觉得很难受，总是不想多说。
咨询师：你的失眠问题和社交压力是单独的问题吗还是？
来访者：（提供信息）我觉得两者是有关联的，每次参加完公司活动回来，晚上就会躺在床上一直想自己说错的话或者做错的事。
咨询师：从你描述的情况来看，失眠和社交压力并不是两个各自独立的问题，而是同一条链条上的不同环节。这通常不是简单的“回忆”，而是一个反复自我审查的过程
来访者：（提供信息）医生，是的，我经常躺在床上一遍又一遍地回想，觉得自己应该表现得更好一些。
咨询师：你可以给我举一个例子吗
来访者：（提供信息）比如前几天的聚餐，我想跟小李聊天，开口时却说了句今天的菜好咸，现在想起来就觉得很尴尬。
咨询师：当时具体的情况是？为什么说菜咸让你感到尴尬呢
来访者：（提供信息）我感觉那句话太生硬了，好像在抱怨一样，其他同事都聊得很开心，我却说了这么扫兴的话。
咨询师：听起来，你在担心自己的态度让同事们感到心里不舒服，因为你打破了场面上的和谐。
来访者：（提供信息）是的，我总觉得自己应该说一些更有趣或者更积极的话。
\end{Verbatim}
\caption{A dialogue between a human counselor and an LLM-based client (Profile ID: 56 and Trajectory ID: 105). For the English version, see Figures \ref{Fig-p56-t105-en-part-1} and \ref{Fig-p56-t105-en-part-2}.}
\label{Fig-p56-t105-zh}
\end{figure*}
\end{CJK*}

\begin{CJK*}{UTF8}{gbsn}
\begin{figure*}[htp]
\centering
\begin{Verbatim}[fontsize=\scriptsize, frame=single, framerule=0.3pt, framesep=4mm, breaklines, xleftmargin=0.5em, xrightmargin=0.5em, formatcom=\rmfamily]
Counselor: Hello, hello.
Client: (Reasonable Request) Hello, teacher. I came today because I wanted to ask for some guidance. How can I move forward from the pain of losing my mother?
Counselor: Hearing you say that, I can sense how heavy this loss has been for you. If you’re willing, you could tell me about a recent moment when you missed her especially strongly, or when the pain felt particularly overwhelming.
Client: (Providing Information) Last week, when I was sorting through her belongings, I found a letter she had written to me. I completely lost control and cried for a very, very long time.
Counselor: I can understand how that must have been a moment when all the emotions suddenly came flooding out. Here, you don’t need to hold back your tears, and you don’t need to regain composure right away. We can go at a pace that feels bearable for you, step by step.
Client: (Providing Information) Thank you. Even now, I can’t fall asleep every night. My mind is full of memories of my mother before she passed away.
Counselor: I can sense how exhausting this must be for you. Nighttime, when the body is supposed to rest, instead becomes the moment when longing and memories surge the most. I’m also curious—when these images come into your mind, are they mostly gentle and warm memories, or are they mixed with pain, regret, or even self-blame? Different kinds of memories can affect your sleeplessness in different ways.
Client: (Providing Information) It’s both. When I remember her cooking for me or chatting with me, I feel warmth. But when I think about how I wasn’t able to cure her illness, I feel deeply guilty and sad.
Counselor: What you’re describing—these two kinds of feelings—are both very real, and they often coexist after losing a loved one. The warm memories reflect the deep and genuine bond you shared with your mother. The guilt and sadness that follow often come from your love for her, your sense of responsibility, and how much you wished she could have lived longer and suffered less. It’s precisely because you cared so deeply that you blame yourself this way.
Client: (Providing Information) Yes. I wanted so badly for her to keep living. Even though I stayed by her side and took care of her the whole time, I still feel like I didn’t do enough.
Counselor: When you say that, I can feel a profound heartbreak, as well as the all-out love you had for your mother. “Wanting her to keep living” says so much about how unwilling you were to let go, how much you wished you could carry more of the pain for her. And precisely because you cared so deeply, after she passed, you keep looking back again and again, feeling that nothing you did was ever enough.
Client: (Providing Information) Hearing you say this, I can’t help but cry again. It feels like you really understand how I feel.
Counselor: Sometimes, when people finally cry, it’s not because the pain has grown stronger, but because they no longer have to carry it alone. What you just said—“it feels like you really understand me”—is in itself very important and very precious.
Client: (Confirming, Extending) You’re right. I’ve been trying hard to control myself, not wanting to appear weak in front of my family. When my mother was alive, she always encouraged me to be strong, but she also told me that it was okay to cry when I was hurting. Yet now, I don’t even dare to remember her together with my family.
Counselor: Not daring to appear weak in front of your family, not daring to remember her together—I understand this as you not only bearing the pain of losing your mother, but also carrying the role of “I have to hold everything together.” It’s as if once you cry or mention her, the whole family might fall back into grief, and without realizing it, you’ve placed yourself in the position of the one who must not collapse. That is an incredibly heavy responsibility.
Client: (Providing Information) Yes. I feel that I need to take care of my husband’s and my son’s emotions, but I myself really need comfort and understanding too.
Counselor: What you’re saying right now captures the hardest part of your situation very clearly: on one hand, you’re taking care of your husband’s and son’s emotions; on the other, you’re pushing your own needs to the back. This isn’t because you don’t need comfort—on the contrary, it’s because you need it so much, yet feel you’re not allowed to ask for it.
Client: (Providing Information) Maybe it’s because my mother taught me to take good care of the family. I always feel that I must be strong and not make my family worry about me.
Counselor: Hearing how you understand yourself, I see a very clear—and heartbreaking—thread: you’ve internalized your mother’s teachings as demands on yourself. Taking care of the family, not making others worry, being strong—these were originally your mother’s hopes for you and expressions of her love for the family. But after she passed away, these messages seem to have turned into an “inner rule” you feel compelled to follow, constantly reminding you that you must not fall apart.
Client: (Providing Information) You’re right. I always feel that if I collapse too, the whole family will truly fall apart. But I’m really so tired now.
Counselor: What you just said is both very heavy and very real. “If I collapse too, the whole family will fall apart”—it’s like a string that’s been stretched tight in your heart, holding you firmly in a position where you’re not allowed to fall. And at the same time, you’re clearly aware that you’ve been holding on for far too long, far too hard. You’re truly exhausted.
Client: (Providing Information) I know what you’re saying makes sense, but every time I relax even a little, all those memories and pain rush in at once. It scares me, and I don’t dare to put this burden down.
Counselor: From this perspective, your “not daring to put the burden down” isn’t the problem itself—it’s actually a very common and quite intelligent form of self-protection after such a profound loss.
Client: (Providing Information) What I’m most afraid of is that if I put these burdens down, I’ll no longer be able to feel the love and responsibility my mother left me with.
Counselor: This fear you’re describing is very deep and very important. What you’re afraid of isn’t “relief” itself, but that once you let go of the burden, the close bond between you and your mother will be cut off as well. In your experience, responsibility, strength, and endurance have become ways of continuing to love her, remember her, and remain loyal to her. If you stop carrying these things, you worry that her place in your heart will fade.
Client: (Providing Information) Yes. I don’t dare to relax because it feels like pain and longing are the only proof of how much I love my mother.
Counselor: The fact that you can say this out loud reflects a very deep level of awareness. In your own way, you’re telling me that pain and longing aren’t just emotions for you—they’re also a kind of proof: proof that you haven’t forgotten her, proof that your love for her is still alive. After losing someone so important, feeling this way is not rare at all, and there’s nothing strange about it.
Client: (Providing Information) Hearing you say this, I feel like I’m starting to understand—maybe I don’t necessarily have to use pain to prove my love for my mother.
Counselor: I’d really like to stay here with you for a moment, rather than rushing forward. Because what you’re touching right now is a very central shift:
you’re beginning to sense that love might have other ways of existing, not only through constant pain, self-blame, and endurance.
Client: (Providing Information) Yes. I do hope to slowly learn to remember and love my mother in a gentler way.
Counselor: Hearing you say that, I feel this is a very mature and tender wish. You’re not trying to forget your mother or let her go—you’re trying to find a way to remember and love her that doesn’t hurt you so deeply, yet remains true.
Client: (Other, Providing Information) Your understanding makes me feel more at ease. Lately I’ve been thinking about carefully organizing my mother’s photos and putting them into a beautiful album.
\end{Verbatim}
\caption{A dialogue between a human counselor and an LLM-based client (Profile ID: 30 and Trajectory ID: 12) (English version, Part 1).}
\label{Fig-p30-t12-en-part-1}
\end{figure*}
\end{CJK*}

\begin{figure*}[htp]
\centering
\begin{Verbatim}[fontsize=\scriptsize, frame=single, framerule=0.3pt, framesep=4mm, breaklines, xleftmargin=0.5em, xrightmargin=0.5em, formatcom=\rmfamily]
(continued)
Counselor: I notice a very important change here. In the past, when you talked about things related to your mother, they were often accompanied by loss of control, guilt, and fear. Now, you’re talking about sorting, choosing, and taking your time. This suggests that you’re beginning to make space for her in your heart, rather than being instantly overwhelmed by memories.
Client: (Confirming) Yes. I feel that organizing the photos is something I can do. Maybe it will help me remember my mother with a calmer state of mind.
Counselor: Perhaps you can think of this as a gentle conversation, rather than a task.
You don’t need to finish everything at once, and you don’t need to handle every photo perfectly.
Some photos, you might linger on a bit longer;
others, you might not feel ready to touch yet.
These choices themselves are ways of respecting your own emotional limits.
Client: (Providing Information) Yes, I agree there’s no need to rush. I want to slowly look at them one by one and let each photo take me back to the beautiful times I shared with my mother.
Counselor: I really want to acknowledge what you said about “not rushing.” This is already very different from where you started—when you felt you had to hold everything together and couldn’t stop.
Client: (Providing Information) I can try to look at the photos with a smile, letting myself remember the joy I had with my mother, instead of crying every time.
Counselor: You’re learning something that is truly not easy: how to remain in love without losing her—and without losing yourself.
Client: (Providing Information, Confirming) Yes. My mother would surely want to see me stay happy and strong. I still remember her often saying that life has to move forward and that we can’t stay immersed in grief forever. I’d like to try sharing these photos with my family, so we can remember her beautiful side together.
Counselor: What you’re saying now carries a sense of movement—from bearing everything alone toward reconnecting. You’re beginning not only to talk with your mother in your heart, but also to bring this remembrance back into the family, so it no longer has to be a weight carried only by you.
Client: (Providing Information) Yes. I want to try turning love and memories into something warm that the whole family can share, rather than something I bear silently on my own.
Counselor: I see this as a very important turning point. You’re no longer only thinking about how to shoulder the pain alone; you’re starting to transform love and memories from a “burden that must be carried alone” into a warmth that can be held together as a family.
Client: (Providing Information) After talking with you today, I feel much less pressure inside. I’m really grateful for your understanding and guidance.
Counselor: Hearing that truly makes me happy for you. Feeling even a little lighter is already a very meaningful change.
Client: (Providing Information) I’ll set aside some time this week to organize the photos. I may still need your guidance, and I hope to gradually learn how to live better with this longing.
Counselor: Of course. You don’t have to figure this out on your own. I’ll be here, continuing to accompany you as you learn how to remember in love—and how to live while remembering.
Client: (Providing Information) Okay. I’ll go home and slowly start organizing the photos, and I’ll come back next week to share how I’m feeling and how things are going.
Counselor: You can go home with peace of mind and move at your own pace. We’ll continue next week.
Client: (Providing Information) Yes, I’m actually looking forward to coming back next week to talk with you again. I’ll head off now.
Counselor: Alright. See you next week.
Client: (Providing Information) Thank you for helping me so much today. I feel much more at ease now. Goodbye.
Counselor: Goodbye.
Client: (Providing Information) Goodbye, teacher. I’ll take it slowly, just as you suggested.
Counselor: I look forward to hearing your reflections next time.
Client: (Providing Information) I’ll organize the photos at a pace I can handle, and then we can talk about how it goes.
Counselor: Wishing you all the best.
Client: (Providing Information) Thank you for your blessing. With your help, I truly believe I’ll be able to get through this low point.
\end{Verbatim}
\caption{A dialogue between a human counselor and an LLM-based client (Profile ID: 30 and Trajectory ID: 12) (English version, Part 2).}
\label{Fig-p30-t12-en-part-2}
\end{figure*}

\begin{CJK*}{UTF8}{gbsn}
\begin{figure*}[htp]
\centering
\begin{Verbatim}[fontsize=\scriptsize, frame=single, framerule=0.3pt, framesep=4mm, breaklines, xleftmargin=0.5em, xrightmargin=0.5em, formatcom=\rmfamily]
Counselor: Hello.
Client: (Providing Information) Hello, doctor. My name is Li Hua. I’ve been doing very poorly lately. Since my mother passed away, I haven’t been able to sleep at night, and I feel an overwhelming sense of pain in my heart.
Counselor: Thank you for being willing to share these feelings with me. Losing one’s mother is an extremely heavy loss, and after such an experience, insomnia and deep emotional pain are actually very common and very understandable reactions. It sounds like you’ve been carrying a great deal of suffering on your own during this time.
Client: (Providing Information, Extending) Yes. Ever since my mother passed away, I feel like I completely collapsed. I can’t focus at work, and at night I just toss and turn. Every time I think about how she looked when she was sick, I feel overwhelming guilt. I keep thinking that if I had noticed her illness earlier, things might have been different. My husband tells me it wasn’t my fault, but I just can’t stop thinking this way.
Counselor: I can feel how heavy the burden you’re carrying is right now. Your mother’s passing didn’t just take away someone deeply important to you—it also keeps pulling you back into those painful images, especially of her when she was ill. Listening to you, it feels like you’re holding two things at once: the profound grief of losing your mother, and at the same time a relentless self-blame, as if everything could have been prevented if only you had done a little more.
Client: (Providing Information) Yes, doctor. Every night I lie in bed replaying every detail of my mother’s illness. I keep thinking that if I had been more attentive, taken better care of her, maybe the outcome would have been different.
Counselor: When you say, “If I had been more attentive, if I had taken better care of her,” I hear not only how much love you have for your mother, but also how harshly you’re judging yourself.
Client: (Providing Information) Deep down I know my husband is right—that my mother’s illness wasn’t my fault. But whenever I think about how much pain she was in before she passed away, I just can’t stop blaming myself.
Counselor: The fact that you can say so clearly, “I know rationally it wasn’t my fault, but emotionally I can’t stop blaming myself,” already shows a great deal of self-awareness. Very often, what hurts the most is precisely this gap—understanding something logically, yet being unable to stop the inner torment. So this isn’t about you being irrational; it’s about grief that hasn’t yet found a place to settle.
Client: (Providing Information, Extending) Recently, while sorting through my mother’s belongings, I found a letter she had written to me. Seeing her words—how much she cared about me and the hopes she had for me—I couldn’t stop crying. When I think about how she used to cook for me and talk with me, it feels like my life suddenly lost the most important person. That feeling makes me feel extremely helpless and lonely.
Counselor: As you describe that moment, I can feel how overwhelming it was for you. Being surrounded by love and, at the same time, struck by loss can be unbearable all at once. Crying uncontrollably is a very natural response.
Client: (Self-criticism or Hopelessness) I don’t even know if I can pull myself together again. I feel like I’m about to be crushed by this grief.
Counselor: I want to say something very clearly, and very importantly: right now, you don’t need to pull yourself together, and you don’t need to prove that you can move on. After a loss as profound as losing your mother, feeling consumed by grief is not a sign of weakness—it’s a deeply human response.
Client: (Providing Information) Thank you for saying that, doctor. I feel like someone finally understands me. It’s as if I don’t have to force myself to move on so quickly anymore.
Counselor: I’m really glad you can feel that way. Being understood and emotionally held is, in itself, something very important. You don’t need to rush, and you don’t need to prove anything. You can take this slowly here—I’ll walk this part of the road with you.
Client: (Providing Information) Honestly, I haven’t felt this relaxed in a long time. Whenever I talk to family or friends about my mother, they rush to comfort me and tell me to cheer up, and that actually makes me feel worse.
Counselor: It’s very important that you can put that into words. Your family and friends may be trying to love you in the ways they know how, but what you truly need may not be comforting words—it may be someone who is willing to sit with you in your grief, without rushing to pull you out of it.
Client: (Defending) I feel that I really did my best to take care of my mother. I cooked three meals for her every day, gave her IV drips, and kept her company. I don’t need anyone telling me that I didn’t do enough.
Counselor: I notice that you’re starting to describe yourself as “I really did my best,” rather than only saying, “If only I had done more back then.” This isn’t a denial of grief—it’s you beginning to give your own efforts the recognition they deserve. This step is crucial in loosening the grip of endless self-blame.
Client: (Defending) Even though I know you’re trying to help me ease my guilt, I still feel that I understand the situation better. If I had just been more attentive back then, I really could have taken better care of my mother.
Counselor: I want to be very honest with you as well: I’m not in a hurry for you to let go of your guilt, or to immediately accept the idea that “you did enough.” For you, this guilt isn’t a simple cognitive error—it’s tightly intertwined with love, responsibility, and loss. If we take it away too quickly, it might feel as though someone is denying how much you cared about your mother.
Client: (Defending) Doctor, I know you mean well, but I feel that only someone who has actually cared for a seriously ill person can truly understand the helplessness and guilt I felt back then.
Counselor: I understand why you would say that, and I respect it deeply. I believe you’re saying this because you truly lived through it. My role here isn’t to judge whether you were attentive enough, but to accompany you as you slowly speak through that path—one that very few people truly walked with you at the time.
Client: (Extending) You’re right, doctor. Back then I carried everything on my own. Maybe because no one shared that burden with me, those painful memories have stayed lodged in my heart and won’t let go.
Counselor: Being able to say this is actually a very important moment of contact. When you say, “I carried everything on my own,” I sense that during that time, you weren’t only caring for your mother—you were also alone with fear, responsibility, decision-making, and the anticipation of loss. When these experiences aren’t shared or witnessed at the time, they often remain later as memories, self-blame, and repetitive rumination.
Client: (Providing Information) Yes. Back then, I desperately wished someone could understand how helpless I felt. Even now, when I think back on those days of facing my mother’s illness alone, my heart still aches.
Counselor: When you say this, I sense a very deep, delayed grief. It’s not only grief for your mother, but also compassion for yourself—the you who bore everything alone back then.
Client: (Extending) During that period, after work every day I rushed to the hospital to take care of my mother. I was already exhausted from work, but when I saw her lying in that hospital bed, I felt I couldn’t allow myself to rest at all. I always wanted to stay with her a little longer. Looking back now, I realize how exhausted and overwhelmed I really was.
Counselor: As you recall those days, I can clearly hear a shift in your tone. You’re no longer standing only in the position of “if only I had done more,” but are beginning to see the version of yourself who was already pushed to the limit, yet still wouldn’t allow herself to stop.
Client: (Confirming) You’re right, doctor. I really put far too much pressure on myself back then.
Counselor: That sentence carries real weight. It’s not said lightly—it comes from a more complete and honest view of that experience. You’re beginning to move your focus from “Was I good enough?” to “How much was I actually carrying at the time?”
Client: (Providing Information) When I think about it now, I really was exhausted both physically and mentally. Maybe it’s precisely because I was so worn out that I kept feeling like I hadn’t done enough.
Counselor: That insight is powerful. It reflects a fuller, more compassionate understanding of what you went through—shifting your attention from self-judgment to acknowledging the sheer weight you were bearing.
Client: (Providing Information) One more thing—I want to mention that my husband keeps encouraging me to pull myself together, but I just can’t return to normal so quickly. That makes me feel very guilty.
\end{Verbatim}
\caption{A dialogue between a human counselor and an LLM-based client (Profile ID: 30 and Trajectory ID: 139) (English version, Part 1).}
\label{Fig-p30-t139-en-part-1}
\end{figure*}
\end{CJK*}

\begin{figure*}[htp]
\centering
\begin{Verbatim}[fontsize=\scriptsize, frame=single, framerule=0.3pt, framesep=4mm, breaklines, xleftmargin=0.5em, xrightmargin=0.5em, formatcom=\rmfamily]
(continued)
Counselor: I can sense how torn you feel—caught between your very real, ongoing grief and your husband’s expectations and concern. When you find yourself unable to keep up with the pace of “getting better,” the feeling of letting someone down can be deeply draining in itself.
Client: (Providing Information, Extending) Doctor, I know my husband means well, but every time he urges me to pull myself together, I feel incredibly sad. Sometimes I just wish he could quietly stay by my side, give me a hug, and say nothing—rather than saying things that make me feel even more suffocated.
Counselor: I can feel the loneliness of being misunderstood. It’s not that you fail to appreciate your husband’s intentions—it’s that his way of caring, especially when you’re most vulnerable, makes things feel heavier rather than lighter. What you truly long for isn’t to be pushed forward, but to be quietly accompanied where you are.
Client: (Providing Information) I really wish my husband could understand that what I need right now isn’t advice or encouragement, but his silent presence and support.
Counselor: You’ve actually articulated your most genuine and core need very clearly. You’re not rejecting care, nor are you unwilling to accept help—you need a kind of companionship that doesn’t demand change, doesn’t imply that something is wrong with how you are now, but simply lets you know you don’t have to carry this alone.
Client: (Providing Information) Maybe I should find a chance to talk to my husband about this, to let him know what I really need right now.
Counselor: That’s an important step—not because you “should” communicate, but because you’re beginning to take your own needs seriously and trust that they deserve to be heard.
Client: (Providing Information) I always feel it’s hard to say these things out loud. I’m afraid my husband will think I’m being overly sensitive or too weak.
Counselor: That fear makes a lot of sense. For many people, the hardest part isn’t knowing what they need, but worrying that once they voice those needs, they might be misunderstood, dismissed, or even looked down upon. Your fear that your husband might see you as overly sensitive or weak often reflects not a lack of confidence, but how much you care about this relationship—and about how you are seen within it.
Client: (Providing Information) Yes, you’re right. I’m afraid that expressing what’s in my heart might affect my relationship with my husband.
Counselor: I sense a very clear and gentle concern in you: it’s not that you don’t want closeness with your husband, but that you care so deeply about the relationship that you worry being honest might create distance instead. That fear itself shows how much you value him and the bond you share.
Client: (Confirming) That’s exactly it, doctor. You’ve really touched something deep in me. I treasure my relationship with my husband very much.
Counselor: I can feel how much you’ve been giving to this relationship all along. Even when you’re in the most pain and most in need of comfort, you’re still thinking first about him—what he might think, whether he’ll see you as too fragile, whether the relationship might be affected. That habit of prioritizing others speaks to how responsible and considerate you are. But after losing your mother, constantly putting others first has made you even more exhausted.
Client: (Providing Information) I really am so tired. I have to regulate my own emotions while also taking care of my husband’s and my son’s feelings. Sometimes I just want to cry out loud.
Counselor: As you say this, I can clearly feel that bone-deep exhaustion—being drained to the point of having almost nothing left. You’re not only grieving your mother, but constantly managing yourself, your husband’s reactions, and your child’s emotions, as if no matter how much you’re hurting, you must stay steady and hold everything together. Anyone would reach their limit living like this for so long.
Client: (Providing Information) I feel like if this continues, I really won’t be able to hold on. Every day I live while holding back tears.
Counselor: I want to pause here and stay with you very seriously.
Living day after day suppressing tears and feeling like you’re about to collapse isn’t just ordinary exhaustion—it’s your body and mind sending a warning after prolonged suppression and solitary endurance. You’re not being dramatic, and you’re not lacking resilience. You’ve truly reached a very extreme point.
Client: (Defending) Doctor, no one understands those painful days better than I do, so I feel like I should keep pushing on—just like I did when I was taking care of my mother.
Counselor: Back then, your perseverance was part of an emergency—there had to be someone holding everything together. Now, that emergency phase has passed, but your body and emotions are still stuck in a mode of “I can’t collapse.” Staying in that state long-term doesn’t make the pain disappear; it only makes the burden heavier and lonelier.
Client: (Providing Information, Extending) I know I should rest properly, but every time I lie down, I think of my mother and start crying uncontrollably. Every time you talk to me like this, I feel that someone finally truly understands the struggle and pain inside me.
Counselor: I take what you just said very seriously, and I’m holding it with care. Perhaps we can slowly rethink what “rest” means.
Rest doesn’t necessarily mean falling asleep right away, not thinking about your mother, or not crying. For you right now, rest might simply mean allowing yourself to lie there and cry for a while, instead of forcing yourself to get up and keep enduring.
Client: (other, Providing Information) This really feels like a place where I can breathe. Before, I always cried alone in hiding. Doctor, you’re right—I really should give myself a chance to catch my breath.
Counselor: Hearing you say that, I feel a sense of relief for you. Maybe we can think of this “breathing space” as something small and concrete, rather than a distant or overwhelming goal.
\end{Verbatim}
\caption{A dialogue between a human counselor and an LLM-based client (Profile ID: 30 and Trajectory ID: 139) (English version, Part 2).}
\label{Fig-p30-t139-en-part-2}
\end{figure*}

\begin{CJK*}{UTF8}{gbsn}
\begin{figure*}[htp]
\centering
\begin{Verbatim}[fontsize=\scriptsize, frame=single, framerule=0.3pt, framesep=4mm, breaklines, xleftmargin=0.5em, xrightmargin=0.5em, formatcom=\rmfamily]
Counselor: Hello! Welcome to the counseling center.
Client: (Providing Information) Hello, my name is Zhang Yiyi. I’ve been under a lot of stress lately and was hoping to get some help.
Counselor: Hello, Yiyi. I’m really glad you’re willing to speak up here. Where would you like to begin?
Client: (Providing Information) I’ve been studying in the U.S. for one semester now, but I just can’t seem to adapt. I feel lonely almost every day.
Counselor: Coming alone to a completely different country—with a new language, a new culture—and after only one semester, having to cope simultaneously with academic demands, daily life, shifts in identity, and a lack of emotional support—it is actually very understandable and very real to feel unadapted and lonely in this situation. This is not a flaw in you.
Client: (Providing Information) But when I look at other Chinese students around me, they seem to adapt so quickly. They join extracurricular activities and make friends with American classmates. I just can’t do that…
Counselor: I sense that there is a great deal of hurt hidden in what you just said. When you see “everyone else can do it, but I can’t,” the feeling is often not only loneliness—it’s mixed with self-blame, shame, and even doubts about whether something is wrong with you. Your current state does not mean that you lack ability, have a flawed personality, or aren’t trying hard enough. What you see as “they’re adapting well” is only the visible surface. Adaptation itself varies enormously from person to person.
Client: (Providing Information) Yes, you’re right. But I feel that I used to be an outstanding student back home, and now I’ve become so inadequate.
Counselor: I believe this is an extremely heavy blow for many students who were consistently affirmed and praised in their home country. I don’t think what you’re experiencing is “being inadequate.” Rather, it’s that someone who once stood firmly in the world through being “excellent” has suddenly been placed into a completely unfamiliar coordinate system. The old standards of evaluation, the speed of feedback, and one’s social position suddenly stop working—and that alone can trigger a powerful sense of self-collapse.
Client: (Providing Information) Exactly. I keep feeling like I’m useless, and I don’t even dare to answer simple questions in class.
Counselor: What I hear is a very harsh and very lonely voice inside you, repeatedly telling you: “I’m useless. I’m not good enough. I can’t afford to make mistakes anymore.”
Client: (Providing Information) That’s true. Every time I think about speaking up, I’m afraid of saying something wrong, so I become even more silent.
Counselor: The fact that you can articulate this so clearly shows that you’re actually very sensitive to your inner experiences and highly self-aware. We can slowly practice a new way together—allowing yourself to exist even in imperfection.
Client: (Defending) When I was in China, my academic performance was always top-tier. Now I feel that my decline is entirely due to language issues, not my actual ability.
Counselor: I think this judgment is very important. Right now, what you need most is not proof of competence, but permission and space for yourself.
Client: (Providing Information, Confirming, Providing Information) But whenever I think that this state might last for several more years, I feel hopeless. I know you’re right—I do need to give myself time to adapt. Still, whenever my American roommates try to chat with me, I don’t know what to say and always find excuses to avoid them.
Counselor: I think what you’re avoiding right now is not your American roommates, but the ongoing fear of being exposed to the thought, “I might not be good enough.” For you, what truly matters is not whether the conversation goes well, but helping your body gradually learn that these situations are not necessarily dangerous.
Client: (Defending) Although my roommates are nice, I feel I have my own way of living. I don’t think I need to force myself to fit in with them.
Counselor: I want to first affirm part of your perspective: you truly have no obligation to change your lifestyle just to “fit in.” Not everyone needs frequent socializing or close relationships with roommates to live well abroad. For some people, solitude, having their own rhythm, and investing more energy in academics or their inner world is genuinely suitable. But I’d also like to gently clarify—does this come from a personal choice, or from resistance?
Client: (Reformulating) Maybe it is because I’m afraid of failing that I chose to avoid them. I probably should try to gradually engage with my roommates.
Counselor: Being able to say, “I might be avoiding,” does not mean denying yourself. On the contrary, it shows that you’re taking back agency. I really value what you just said.
Client: (Providing Information) What small things do you think I could start with when trying to interact with my roommates?
Counselor: That’s an excellent question. At this stage, the key is not “how well you chat,” but helping your body and emotions accumulate experiences of “I can handle this.”
Client: (Confirming) Maybe I could start by having breakfast with them or sharing some food.
Counselor: I think that’s a very good and very fitting starting point. We can make this attempt even more concrete—so that it feels both manageable and safe.
Client: (Providing Information) I’d like to try sharing some snacks I brought from China with them, but I’m still a bit nervous about whether they’ll like them.
Counselor: Your nervousness right now isn’t because this is such a big deal—it’s because you care and don’t want to experience another setback. Noticing this shows that you’re being honest with your own feelings.
Client: (Defending) Even though I’m a bit nervous, I know Chinese snacks better than others. My roommates will definitely appreciate my taste.
Counselor: I can sense that some confidence has returned. Whether they end up liking the snacks or not, what really matters is whether your sense of self and self-worth can begin to be rebuilt.
Client: (Extending) Yes. I’ve always been excellent back home, so I do have good judgment in this area. Even if my English isn’t great, I still have my strengths.
Counselor: This reflects a more complete and balanced understanding of yourself. I’m really glad you’re beginning to recognize a fuller version of who you are.
Client: (Providing Information) Hearing you say this, I’m starting to feel that I’m not completely useless—I just need some time to adapt.
Counselor: I think that’s absolutely true. This is a very important and very genuine shift. Being able to give yourself time for change is something truly precious.
Client: (Providing Information) I feel much better now. Thank you. I hope we can talk more about other issues next time.
Counselor: Of course. We can face any issue together. I’m really happy for you—not because all the problems are solved, but because your inner state now has a bit more space and a bit more gentleness than before. That is a very meaningful change for you.
Client: (Confirming) Yes, I do feel lighter inside, and I have more courage to face what’s ahead.
Counselor: I feel proud of you and reassured. You already have more strength than you realize. Take it slowly—you’re walking a path that is kinder and more authentic to yourself.
Client: (Reasonable Request) By the way, what should I do if my roommates don’t like my snacks?
Counselor: That’s a very mature and realistic question. Thinking ahead about “what if they don’t like them” shows that you’re no longer treating this as a black-and-white test of success or failure, but are leaving psychological space for different outcomes. What do you think yourself?
Client: (Confirming) I understand now. Even if they don’t like them, it’s okay. What matters is that I’m willing to take this step.
\end{Verbatim}
\caption{A dialogue between a human counselor and an LLM-based client (Profile ID: 45 and Trajectory ID: 105) (English version, Part 1).}
\label{Fig-p45-t105-en-part-1}
\end{figure*}
\end{CJK*}

\begin{figure*}[htp]
\centering
\begin{Verbatim}[fontsize=\scriptsize, frame=single, framerule=0.3pt, framesep=4mm, breaklines, xleftmargin=0.5em, xrightmargin=0.5em, formatcom=\rmfamily]
(continued)
Counselor: You don’t need to turn every interaction into “evidence of being accepted.”
You’re simply practicing standing on your own side, even in uncertainty.
Client: (Reformulating) I think the biggest breakthrough for me now is being able to accept that I can be imperfect. Maybe this is the first lesson I need to learn in the U.S.
Counselor: “I can be imperfect and still be a valuable, worthy person.” That is a very profound and important lesson.
Client: (other) It’s a bit cold today. Drinking coffee makes me feel much warmer.
Counselor: Do you usually like coffee?
Client: (other) I actually prefer milk tea. Milk tea shops here are never as good as the ones back home.
Counselor: Haha, I can hear how much you miss home.
Client: (Providing Information, other) Yes—especially on weekend afternoons. I really miss drinking milk tea with friends back home. That said, there are also fun things here. Last weekend I tried an ice cream shop an American friend recommended.
Counselor: When that sense of longing comes up, what do you miss most—the people, the atmosphere, or that feeling of being understood?
Client: (other) Look, yesterday I bought a huge box of cookies at Costco.
Counselor: How do they taste?
Client: (Reformulating) They’re pretty good. It made me think that I could share them with my roommates.
Counselor: I can sense a very gentle and natural shift happening—not from “I should go socialize,” but from “these taste good, and I suddenly thought of sharing them with my roommates.”
Client: (Providing Information) Yes. Thinking about sharing good food with others suddenly made me feel much less weighed down.
Counselor: Your attention is no longer fixed solely on “Am I doing well enough? Will I fail?” Instead, it briefly turns toward the flow within relationships. That’s very important.
Client: (Providing Information) I really want to share some cookies with my roommates now. I feel like food is the easiest way to bring people closer.
Counselor: I can feel that lightness—the natural urge to act—not “I should do something,” but “doing this now feels aligned with me.” That’s very meaningful.
Client: (Providing Information) If I start with sharing food, I actually feel a bit excited about chatting with my roommates.
Counselor: I’ll look forward to hearing how it goes. No matter what happens, remember to stay connected to your own feelings and to accept all possible outcomes.
Client: (Providing Information) I hope next time we can talk as easily as we did today.
Counselor: I look forward to our next meeting.
Client: (Providing Information) Then I’ll go share these snacks with my roommates now. I’m a little nervous, but I want to give it a try.
Counselor: That shows how much you value this relationship.
Client: (Providing Information) Yes. I hope to gradually build a friendlier relationship with them, even if the process is a bit challenging.
\end{Verbatim}
\caption{A dialogue between a human counselor and an LLM-based client (Profile ID: 45 and Trajectory ID: 105) (English version, Part 2).}
\label{Fig-p45-t105-en-part-2}
\end{figure*}

\begin{CJK*}{UTF8}{gbsn}
\begin{figure*}[htp]
\centering
\begin{Verbatim}[fontsize=\scriptsize, frame=single, framerule=0.3pt, framesep=4mm, breaklines, xleftmargin=0.5em, xrightmargin=0.5em, formatcom=\rmfamily]
Counselor: Hello, welcome to the counseling center.
Client: (Providing Information) Hello, doctor. My name is Xiao Wang. I’ve been having some difficulties at work recently and wanted to talk to you about them.
Counselor: Hello, Xiao Wang. I’m listening.
Client: (Providing Information) I feel that I’m really not good at communicating with colleagues at work. I don’t know how to talk to people.
Counselor: It sounds like you feel quite challenged when interacting with colleagues at work, especially when it comes to how to initiate and sustain communication.
Client: (Providing Information) Yes. For example, at a company dinner last week, I sat alone in a corner. I wanted to chat with my colleagues but didn’t know how to start. It felt really uncomfortable.
Counselor: I noticed you used the phrase “really uncomfortable.” It seems that it wasn’t just embarrassment or nervousness, but perhaps also a sense of being excluded.
Client: (Providing Information) Yes. Watching everyone laughing and chatting, I felt like an outsider. I really wanted to fit in, but I didn’t know how.
Counselor: I notice that it’s not that you’re uninterested in your colleagues, nor that you don’t want to fit in. On the contrary, you seem to have a strong desire for connection. Would you be willing to share more about what you notice most clearly in your body or emotions at the moment when you want to join in but end up holding back?
Client: (Providing Information) In situations like that, my heart starts racing, my palms get sweaty, and my mind goes completely blank.
Counselor: In those moments, your body enters a state of heightened alertness, as if preparing to respond to a threat. When the body sounds the alarm first, it’s very natural for the mind to struggle with thinking about what to say.
Client: (Providing Information) You’re right, doctor. Just like you said, whenever I want to start a conversation, it feels like something gets stuck. I freeze completely.
Counselor: The issue isn’t whether you want to speak—it’s how much you care.
Client: (Defending) I don’t think it’s that extreme. I’m just a bit shy. Being introverted isn’t necessarily a bad thing, right?
Counselor: I can sense an important form of self-protection and self-acceptance here. Being introverted or shy isn’t a problem in itself; they are indeed parts of one’s personality and have their own value. The fact that you can speak up for this suggests that you’re not completely rejecting yourself.
Client: (Providing Information, Confirming, Providing Information) Yes, I think I do have some strengths, like being very detail-oriented. I understand what you’re saying—it might be that I care too much about how others see me. Whenever I want to talk to colleagues, I worry about saying the wrong thing or being seen as boring.
Counselor: That’s a very mature level of self-understanding. I’d like to ask further: when you worry about others finding you boring, is this “others” a vague group in your mind, or a particular type of person? Sometimes anxiety doesn’t come from everyone, but from a specific gaze that feels especially important and easily internalized.
Client: (Defending) I think that kind of worry is normal. The workplace is just like that—people who don’t socialize easily get marginalized.
Counselor: Your concern isn’t exaggerated or weak; it’s an attempt to adapt to the environment. Psychologically, though, what you may be carrying now is an amplified internal version of that pressure.
In counseling, what we’re really trying to do may not be to deny workplace realities, nor to force you to become extroverted, but to explore together whether there’s a way for you to gradually “exist” in relationships without having to perform against your nature.
Client: (Reformulating) I see. Maybe I can start with smaller interactions—like talking about work with colleagues I collaborate with frequently.
Counselor: If you’re open to it, we can make this even more concrete. Is there a small, natural moment in your recent work where you might try taking that step?
Client: (Providing Information) Hmm… I could start with Xiao Li at the desk next to mine. She often borrows my stationery. Maybe I could chat a bit more when she comes over.
Counselor: That’s a very specific and realistic idea. It sounds like you’re moving forward in a way that respects your personality while also responding to real-life demands.
Client: (Confirming) Yes, that’s true. What you said really helped things feel much clearer.
Counselor: I’m glad to hear that, and I really value this feeling you’re having right now. On some level, you’ve deepened your understanding of yourself.
Client: (Providing Information) You’re right, doctor. I feel like I have a clearer understanding of my personality now, and I’m not in such a rush to change myself.
Counselor: I’m glad you’ve noticed that shift. Is there anything else I can help you with—for example, would you like to talk about specific communication strategies?
Client: (Defending) I think I’m careful and conscientious in my work, and I don’t really need to change. Maybe it’s just that the work environment demands too much.
Counselor: That’s true—the environment is a very important variable. Often it’s not that you’ve done something wrong or that there’s something wrong with your personality.
Client: (Extending) Exactly. Everyone has their own personality traits. I’ve been trying so hard to adjust myself to fit the environment that I’ve neglected what I actually feel inside.
Counselor: It’s never too late to start paying attention to that. And I notice that you have a strong capacity for self-reflection, which is both important and quite rare.
Client: (Providing Information) You’re right, doctor. I do tend to think a lot—but sometimes I overthink, and that’s what makes me anxious.
Counselor: That awareness is actually very important, and you’ve described it very accurately. Your tendency to think deeply is a strength—it helps you be thorough, consider consequences, and stay sensitive to your environment. But when this ability is constantly activated in situations where you’re highly concerned about others’ evaluations, it can shift from helping you understand the world to repeatedly amplifying pressure, and that’s when anxiety emerges.
Client: (Providing Information) Because I think too much, I end up feeling at a loss around colleagues. Next time, I’ll try thinking a bit less and see if that helps.
Counselor: I sense a gentler and more realistic attitude here—not demanding immediate change, but being willing to make a small experiment and see what happens. That, in itself, is a way of taking care of yourself.
Client: (Confirming) Yes, that makes a lot of sense. I should start with small attempts and take it slowly.
Counselor: Exactly. This understanding you have right now is very important and really aligned with your own pace. You’re no longer asking yourself to suddenly become “someone who’s great at socializing,” but allowing yourself to start with small, manageable steps that won’t overwhelm you.
Client: (Reasonable Request) Doctor, could you help me think about some concrete ways to start conversations with colleagues?
Counselor: Of course—and that’s a very appropriate question to ask. How do you usually communicate with colleagues now?
Client: (Confirming) I usually only talk to colleagues when it’s work-related. Most of the time, I wait for others to approach me.
Counselor: What you’re describing shows a very typical and understandable coping style: you speak when things feel safe, clear, and justified, and otherwise you observe and wait. At the same time, as you’ve already noticed, this pattern—only speaking when necessary and mostly waiting for others—can make you seem passive in relationships and reinforce the feeling that you’re always on the margins, only participating from the outside.
Client: (Reformulating) Maybe I can gradually expand the conversation beyond work topics—it would feel more natural that way.
Counselor: Exactly. You might consider shifting from “waiting for others to come to me” to “sending a very small signal.”
Client: (other) The weather is really nice today—so sunny.
\end{Verbatim}
\caption{A dialogue between a human counselor and an LLM-based client (Profile ID: 56 and Trajectory ID: 105) (English version, Part 1).}
\label{Fig-p56-t105-en-part-1}
\end{figure*}
\end{CJK*}

\begin{figure*}[htp]
\centering
\begin{Verbatim}[fontsize=\scriptsize, frame=single, framerule=0.3pt, framesep=4mm, breaklines, xleftmargin=0.5em, xrightmargin=0.5em, formatcom=\rmfamily]
(continued)
Counselor: About the weather?
Client: (other) Doctor, I noticed the plants outside your office are very well taken care of.
Counselor: Thank you for the compliment. Shall we return to our counseling?
Client: (Providing Information, other) Yes, doctor. Let’s talk more about how exhausted I’ve been lately. I have insomnia every night. I really wish I could have more chances to talk like this—I feel very relaxed talking to you.
Counselor: Thank you for that. Could you tell me more about your sleep?
Client: (other) Doctor, this office is quite spacious and warmly designed.
Counselor: You seem to be quite distracted right now.
Client: (Reformulating) You’re right. I tend to avoid my own problems by bringing up other topics. I think I’m really trying to avoid things that make me anxious.
Counselor: What were the anxiety-provoking issues you were trying to avoid just now?
Client: (Providing Information) When I think about social pressure at work and my insomnia, I feel very uncomfortable and don’t want to talk about it.
Counselor: Do you see your insomnia and social pressure as separate issues, or…?
Client: (Providing Information) I think they’re connected. Every time I come back from a company event, I lie in bed replaying things I said wrong or did wrong.
Counselor: From what you’re describing, insomnia and social pressure aren’t two separate problems, but different links in the same chain. This is usually not simple “recalling,” but a repetitive process of self-scrutiny.
Client: (Providing Information) Yes, doctor. I often lie in bed replaying things over and over, thinking I should have performed better.
Counselor: Could you give me a specific example?
Client: (Providing Information) For instance, at the dinner the other day, I wanted to talk to Xiao Li, but the first thing I said was, “The food is really salty today.” When I think about it now, it feels very awkward.
Counselor: What was the situation at that moment? Why did saying the food was salty feel embarrassing to you?
Client: (Providing Information) It felt too blunt—almost like complaining. Everyone else was chatting happily, and I said something that dampened the mood.
Counselor: It sounds like you were worried that your tone made your colleagues uncomfortable, as if you disrupted the harmony of the situation.
Client: (Providing Information) Yes. I always feel like I should say something more interesting or more positive.
\end{Verbatim}
\caption{A dialogue between a human counselor and an LLM-based client (Profile ID: 56 and Trajectory ID: 105) (English version, Part 2).}
\label{Fig-p56-t105-en-part-2}
\end{figure*}

\section{Experiments}
\label{app:experiments}
\subsection{Prompts for LLM Evaluation}
\label{app:prompts-for-llm-eval}
\begin{figure*}[!t]
    \centering
    \includegraphics[width=\textwidth]{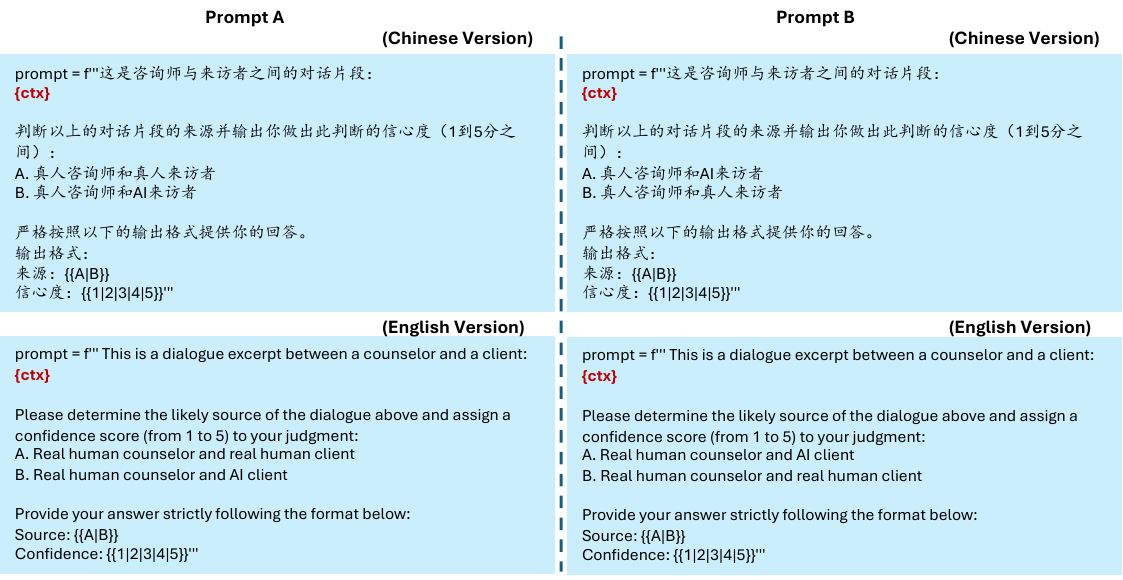}
    \caption{The prompts for LLM evaluation.}
    \label{Fig-prompt-evaluation}
\end{figure*}

We present the prompts for LLM evaluation in Figure \ref{Fig-prompt-evaluation}.

\end{document}